\documentclass[10pt,journal,compsoc]{IEEEtran}

\usepackage{epsfig}
\usepackage{graphicx}
\usepackage{amssymb}
\usepackage{float}
\usepackage{mathrsfs}
\usepackage{epstopdf}
\usepackage{multirow}
\usepackage{color}
\usepackage{algorithmic}
\usepackage{array}
\usepackage{dblfloatfix}    
\usepackage{amsmath}
\usepackage{booktabs}

\pdfoutput=1

\newcommand\T{{\mathpalette\raiseT\intercal}}
\newcommand\raiseT[2]{\raisebox{0.3ex}{$#1#2$}}

\makeatletter
\g@addto@macro\normalsize{%
  \setlength\abovedisplayskip{3pt}
  \setlength\belowdisplayskip{3pt}
  \setlength\abovedisplayshortskip{3pt}
  \setlength\belowdisplayshortskip{4pt}
}
\makeatother

\def\eg{{\em e.g.}}
\def\ie{{\em i.e.}}

\mathchardef\mhyphen="2D 

\ifCLASSOPTIONcompsoc
  \usepackage[nocompress]{cite}
\else
  \usepackage{cite}
\fi

\ifCLASSOPTIONcompsoc
 \usepackage[caption=false,font=footnotesize,labelfont=sf,textfont=sf,subrefformat=parens,labelformat=parens]{subfig}
\else
 \usepackage[caption=false,font=footnotesize]{subfig}
\fi

\usepackage{url}

\begin{document}
\bstctlcite{IEEEexample:BSTcontrol}

\graphicspath{{figures/}}

\title{Harmonized Multimodal Learning with Gaussian Process Latent Variable Models}


\author{Guoli~Song, Shuhui~Wang,~\IEEEmembership{Member,~IEEE,}
        Qingming~Huang,~\IEEEmembership{Fellow,~IEEE,}
        and~Qi~Tian,~\IEEEmembership{Fellow,~IEEE}
\IEEEcompsocitemizethanks{
\IEEEcompsocthanksitem  G. Song and Q. Huang are with the School of Computer Science and Technology, University  of  Chinese  Academy  of  Sciences, Beijing 100049, China, and with the Key Laboratory of Intelligent Information Processing, Institute of Computing Technology, Chinese Academy of Sciences, Beijing 100190, China. \protect\\
E-mail:  guoli.song@vipl.ict.ac.cn, qmhuang@ucas.ac.cn.
\IEEEcompsocthanksitem  S. Wang is with the Key Laboratory  of Intelligent  Information  Processing, Institute  of  Computing  Technology,  Chinese  Academy  of  Sciences, Beijing 100190, China. \protect\\
E-mail: wangshuhui@ict.ac.cn.
\IEEEcompsocthanksitem  Q. Tian is with Noah's Ark Lab, Shenzhen 518129, China, and also with the Department of Computer Science, The University of Texas at San Antonio, San Antonio, TX 78249 USA. \protect\\
E-mail: tian.qi1@huawei.com.}
}

\IEEEtitleabstractindextext{%
\begin{abstract}
Multimodal learning aims to discover the relationship between multiple modalities. It has become an important research topic due to extensive multimodal applications such as cross-modal retrieval.
This paper attempts to address the modality heterogeneity problem based on Gaussian process latent variable models (GPLVMs) to represent multimodal data in a common space. Previous multimodal GPLVM extensions generally adopt individual learning schemes on latent representations and kernel hyperparameters, which ignore their intrinsic relationship.
To exploit strong complementarity among different modalities and GPLVM components, we develop a novel learning scheme called \emph{Harmonization}, where latent model parameters are jointly learned from each other.
Beyond the correlation fitting or intra-modal structure preservation paradigms widely used in existing studies, the harmonization is derived in a model-driven manner to encourage the agreement between modality-specific GP kernels and the similarity of latent representations.
We present a range of multimodal learning models by incorporating the harmonization mechanism into several representative GPLVM-based approaches.
Experimental results on four benchmark datasets show that the proposed models outperform the strong baselines for cross-modal retrieval tasks, and
that the harmonized multimodal learning method is superior in discovering semantically consistent latent representation.

\end{abstract}

}

\maketitle

\IEEEdisplaynontitleabstractindextext

%
\IEEEpeerreviewmaketitle

\IEEEraisesectionheading{\section{Introduction}\label{sec:introduction}}
\IEEEPARstart{M}{ultimodal} learning aims to discover the relationship between data of multiple modalities such as images, texts, graphics and audios ~\cite{cvpr/YanM15,icml/NgiamKKNLN11,baltruvsaitis2017multimodal}. It has been extensively studied in recent years owning to the demands evoked by a wide range of applications, \eg, audio-visual speech recognition~\cite{apin/NodaYNOO15}, image captioning~\cite{cvpr/VinyalsTBE15,pami/FuJCSZ17} and cross-modal retrieval~\cite{pami/PereiraCDRLLV14,journals/pami/WangHWWT16,tmm/Peng18}. In this work, we focus on cross-modal retrieval task, which aims to retrieve semantic relevant data objects from one modality in response to a query from another modality. Users can retrieve images that are matched to the content of a text, or search relevant textual descriptions of an interesting picture.

\begin{figure}[!t]
\centering
\includegraphics[width=0.49\textwidth]{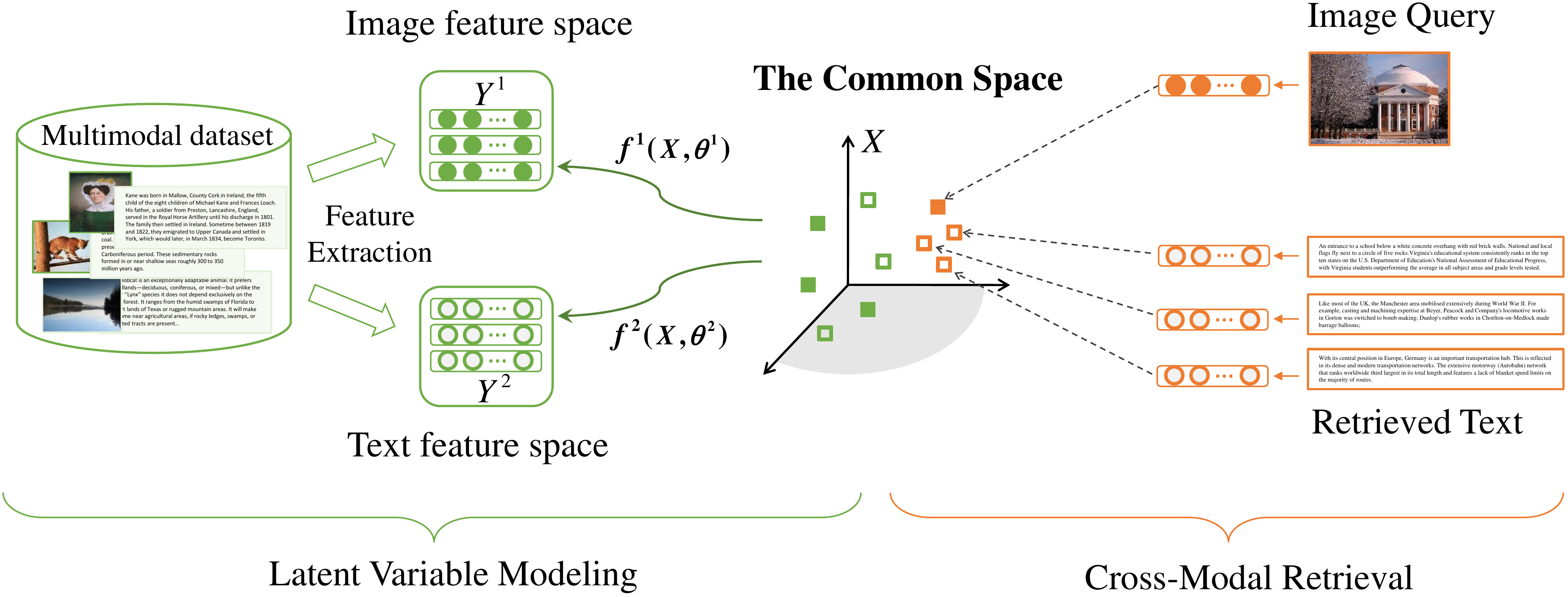}
\caption{Overview of latent variable modeling for multimodal data (left) and an example of cross-modal retrieval (right). In the common space, the solid square indicates the image modality and the hollow square indicates the text modality.}\label{fig:intro-framework}
\end{figure}

The \emph{heterogeneity} of different modalities makes multimodal learning a challenging problem. Instances across different modalities cannot be compared directly since they are represented in different feature spaces. To solve this problem, latent variable modeling techniques are often employed to represent multimodal data in a latent space that captures the correlation information shared by all the modalities.
As shown in Fig.~\ref{fig:intro-framework}, data of different modalities (\eg, image and text) are represented as feature vectors through feature extraction, and then the features are embedded into a common latent space $X$ through modality-specific mapping functions. At last, latent representations can be used to measure cross-modal similarity and enable cross-modal retrieval via similarity ranking.

As a probabilistic non-linear non-parametric latent variable method, Gaussian process latent variable model (GPLVM)~\cite{journals/jmlr/Lawrence05} relies on mappings that constitute samples from Gaussian process (GP) determined by its covariance or kernel function.
Being extended to multimodal learning setting~\cite{nips/ShonGHR05}, the non-parametric nature of modality-specific GPs allows for more flexibility than parametric mappings in learning the low dimensional embedding for heterogeneous data.
Following the correlation observation fitting principle, existing studies on GPLVM extensions focus on the design of the latent space structure, but let hyperparameters of the mapping function be learned independently without any informative prior.
Such schemes are less adaptive to the multimodal information complementarity, and the independently learned GP functions are faced with risk to be mutually incompatible on real-world problems.

In fact, as a key component of GPLVM, the GP kernels, parameterized by the latent representation and the hyperparameters, are recognized as a means of pattern discovery for data extrapolation~\cite{icml/DamianouETL12,icml/WilsonA13}.
Gaussian process mappings with more expressive kernels lead to more explicit representations for heterogeneous data modalities.
The GP kernel can also be seen as a special kind of metric constructed on the latent representation.
Similar as the latent space, the kernel hyperparameters are also regularized with some individual priors in fully Bayesian GPLVM \cite{damianou2015variational,nips/TitsiasL13}. Despite of more smoothness endowed by the priors, the GP kernels are learned towards maximal modality-specific data likelihoods without any explicit structural or semantic guidance. This may lead to limited expressiveness on GP functions, and constrain the multimodal data extrapolation quality in generative learning process.
Besides, the topology or semantic information in observation space is only transferred directly onto the latent space~\cite{tip/EleftheriadisRP15,song2017multimodal} for multimodal complementarity sharing~\cite{apin/NodaYNOO15,nips/MaoXJY16}.
On account of the lack of sufficient information sharing among  the observation, latent and functional spaces of multimodal GPLVM, it may lead to insufficient multimodal information sharing when the structure or semantic guidance tends to be sparse.

Based on the non-parametric multimodal GPLVM, we aim to pave a new way of multimodal information sharing among different GPLVM components beyond the correlation observation fitting paradigm. Specifically, we study a joint learning scheme for latent model parameters in a way that exploits the complementary information across multiple modalities and multiple GPLVM components.
A harmonized learning strategy is proposed for multimodal GPLVMs to obtain a more expressive latent space shared by multimodal data.
In this paper, \emph{harmonized learning} or \emph{harmonization} is a learning mechanism which can enforce the structural agreement between different parameter spaces.
We define a harmonization loss function on GP kernels and the similarity matrix of latent points, and formulate a joint model-driven prior to learn the latent positions of observed data and the modality-specific kernel hyperparameters. Minimizing the harmonization function encourages the structural agreement between similarity in the observed data space, functional space and latent space, and leads to better structural and semantic consistency on the learned subspace.

In the context of multimodal GPLVMs, we propose three harmonized formulations, {\it i.e.}, the F-norm, the $l_{2,1}$-norm and the trace-based harmonization, to provide a full generalization on the harmonization prior with different statistical interpretations. Accordingly, the harmonization constraint is combined with existing multimodal GPLVM learner. Harmonized Multimodal GPLVM (\emph{hmGPLVM}) is the standard multimodal GPLVM~\cite{nips/ShonGHR05} with the harmonization prior.  Harmonized Similarity GPLVM (\emph{hm-SimGP}) enforces harmonized learning on the similarity-based GPLVM~\cite{iccv/SongWHT15} to ensure consistency between different similarity structures. Harmonized m-RSimGP (\emph{hm-RSimGP}) combines the harmonization prior with the inter-modal similarity/dissimilarity prior in the supervised m-RSimGP model~\cite{iccv/SongWHT15} to better preserve semantic correlation between data in the latent space.
The proposed harmonized multimodal GPLVM methods show remarkable multimodal representation learning ability and state-of-the-art performance over existing competitors.

The main contributions of this work are as follows.
\begin{itemize}
  \item We propose multimodal harmonized learning which exploits the complementary information across multiple modalities. Beyond data level fitting mechanism, the {\it Harmonization} enforces a functional level information sharing and alignment among modality-specific transformations via latent representations.
  \item The harmonization, expressed by joint model priors with different divergence measurements on kernel/ similarity matrices, enforces mutual influence between the GP covariance and the similarity in latent space, thus the structural and semantic consistency on the latent data points can be enhanced more comprehensively with good theoretic guarantee.
  \item We extend several multimodal GPLVMs to harmonized versions, and show that harmonization works well with other learning mechanisms to produce a better joint representation for multimodal data.
\end{itemize}

This paper is an extension of our previous conference paper~\cite{iccv/SongWHT17}. Besides the F-norm harmonized multimodal GPLVM proposed in \cite{iccv/SongWHT17}, we extend the harmonization function to more general formulations with theoretic guarantees towards a wider range of applications. Also we apply the harmonization with a range of multimodal GPLVM variants, {\it e.g.}, the m-RSimGP model~\cite{iccv/SongWHT15}. For large scale learning, we leverage GPflow~\cite{GPflow2017}, a Gaussian process library that uses tensorflow for faster/bigger computation. For cross-modal retrieval tasks, we compare the latent representations learned by our harmonized models to strong competitors. On four widely used benchmark datasets, {\it i.e.}, PASCAL, Wiki, TVGraz and MSCOCO, experimental results in terms of mAP score and precision-recall curve demonstrate that the proposed method outperforms the state-of-the-art multimodal learning methods.

\section{Related Work}
\label{sec:relatedwork}
The work of learning latent representations from multimodal data consists of both deterministic (\emph{e.g.}, CCA) and probabilistic (\emph{e.g.}, GPLVM) methods.

\textbf{Deterministic methods} have been popular in multimodal learning field, and many different kinds of methods have been proposed to learn consistent latent representations. Canonical correlation analysis (CCA)~\cite{hotelling1936relations,neco/HardoonSS04} is one of the most representative methods, which uses linear projections to learn latent representations of different modalities that maximize their correlation. Due to its effectiveness in latent variable modeling, CCA has also been widely used in recent works such as \cite{mm/RasiwasiaPCDLLV10,sharma2012generalized,pami/PereiraCDRLLV14}. However, linear CCA-based methods are prone to producing trivial solutions because of the correlation noise among multimodal data~\cite{jmlr/SalzmannEUD10}. Non-linear latent variable models are proposed, among which kernel-based methods~\cite{neco/HardoonSS04,cvpr/Socher010,ijcv/HwangG12} have been applied in many multimodal scenarios. Recently deep learning methods~\cite{icml/NgiamKKNLN11,icml/WangALB15,cvpr/YangRCMBL17,DBLP:conf/mm/WangCZH018} have drawn considerable attention due to their promising representation learning ability. For example, deep extensions of CCA~\cite{icml/AndrewABL13,cvpr/YanM15} are proposed which learn stacked non-linear mappings of two data modalities.
Besides these aforementioned methods, there are many other ways to deal with multimodal learning problem, such as partial least squares (PLS) methods~\cite{slsfs/RosipalK05,mm/HeMWLH16}, hashing methods~\cite{kdd/ZhenY12,sigir/LongCWY16,cvpr/JiangL17}, metric learning methods~\cite{icml/QuadriantoL11,aaai/ZhaiPX13,DBLP:conf/cvpr/WangJHT12}, dictionary learning methods~\cite{aaai/ZhuangWWZL13,tip/MandalB16}, and so on.
These aforementioned approaches just project heterogeneous data into a latent manifold with specific structure, while they lack a general intrinsic interpretation on the modeling mechanism for relations between data modalities.

\textbf{Probabilistic models} are used to construct common representations for different modalities via latent random variables.
For example, multimodal topic models~\cite{sigir/BleiJ03,iccv/JiaSD11} learn topics as the shared latent variables to represent the underlying  correlations in multimodal data. Deep Boltzmann machines~\cite{jmlr/SrivastavaS14} are used to learn multimodal representations where each successive layer is expected to represent the data at a higher level of abstraction. GPLVM-based approaches~\cite{mlmi/EkRTRL08,icml/DamianouETL12} relate data from different modalities in a shared latent space through the use of GP mappings. To enhance the robustness of the latent representation, additive priors~\cite{iccv/SongWHT15,tip/EleftheriadisRP15} or regularization techniques~\cite{mlmi/EkRTRL08,icml/DamianouETL12} are used in the GPLVMs to preserve intra-modal topology and inter-modal semantic correlation.
Recently deep GP extensions~\cite{aistats/DamianouL13,Dai:VAEDGP16} are also proposed by stacking GPLVM modules as building blocks.
Considering the nice properties of GPLVM as discussed in the subsequent sections, we conduct study on such probabilistic non-linear and non-parametric multimodal latent space learning model.

\section{Background}
\label{sec:pre}
This section reviews GPLVM for multimodal learning and its widely used regularization techniques.

\subsection{Multimodal GPLVMs}
\label{sec:mgplvm}
\begin{figure}[!t]
\centering
    \subfloat[mGPLVM]{\includegraphics[height=2.9cm]{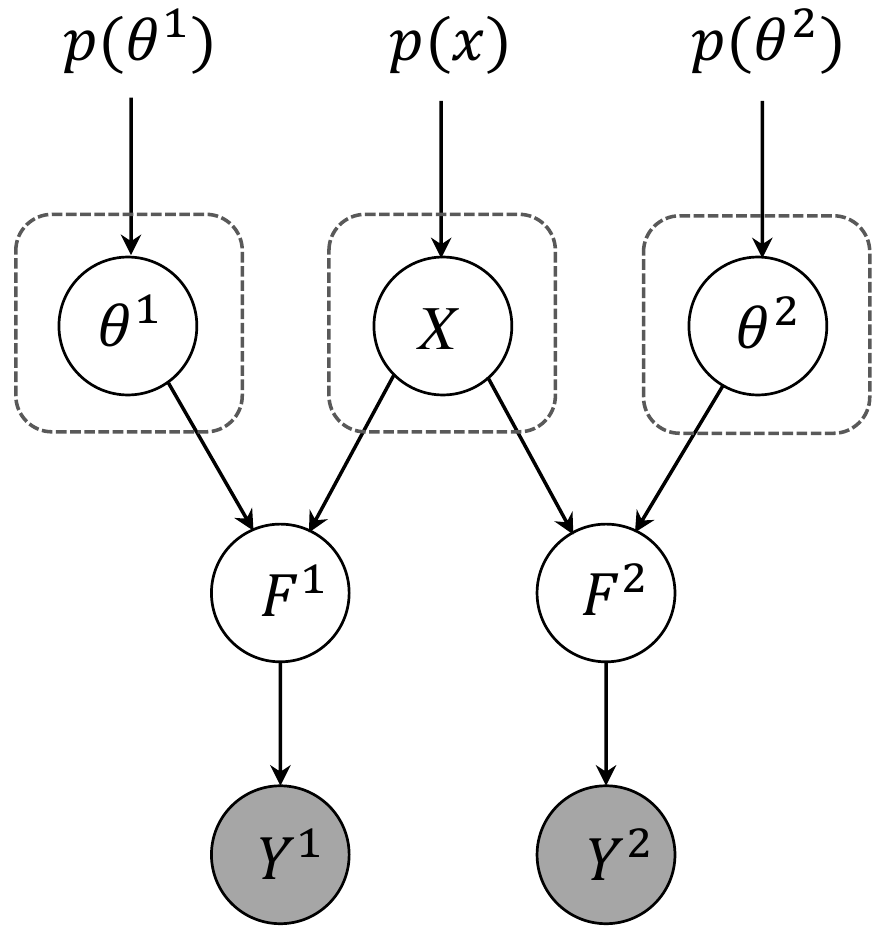}\label{fig:independent}}
    \hspace{0.03\textwidth}
    \subfloat[hmGPLVM]{\includegraphics[height=2.9cm]{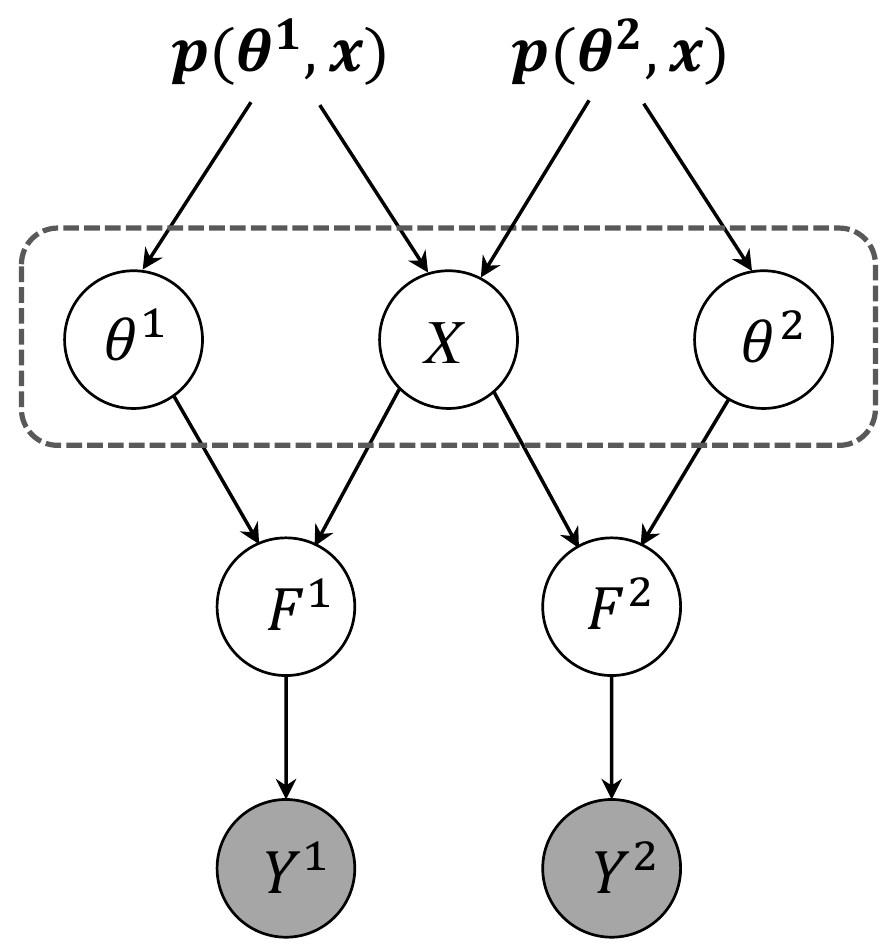}\label{fig:joint}}
    \hspace{0.02\textwidth}
    \subfloat{\includegraphics[height=2.95cm]{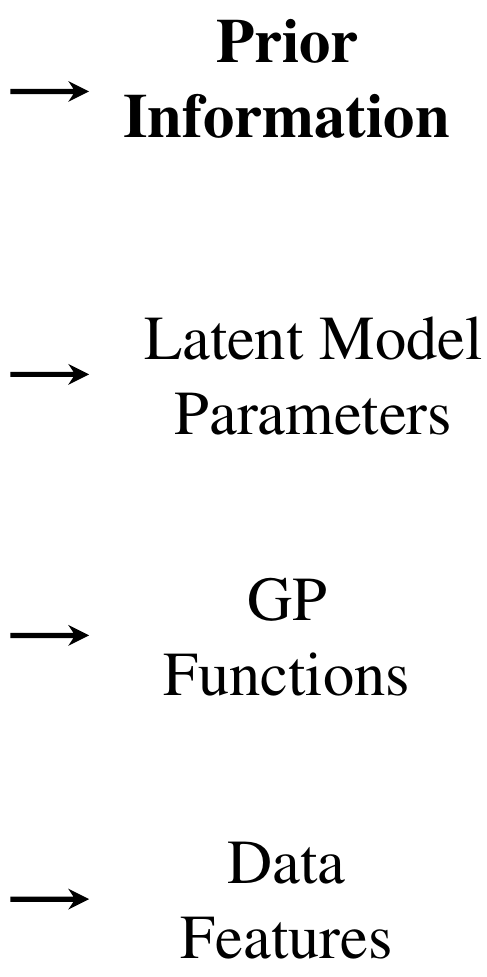}}
\caption{(a) Multimodal GPLVM (mGPLVM): Independent priors are imposed over the latent model parameters ($X, \theta ^1$, $\theta ^2$). (b) Harmonized multimodal GPLVM (hmGPLVM): A joint prior in factored form is imposed over the parameters, which is used to harmonize the learning of $X$ and kernel hyperparameters $\theta ^1$ and $\theta ^2$.} \label{fig:hmgplvms}
\end{figure}

Different data modalities are assumed to be aligned in a latent shared manifold for the techniques based on multimodal Gaussian process latent variable model (m-GPLVM)~\cite{icml/DamianouETL12,nips/ShonGHR05,mlmi/EkRTRL08}. Without loss of generality, we consider two data modalities \( Y^1 = [y_1^1, \ldots, y_N^1]^\T \in {\mathbb{R}^{N \times {d_1}}}\) and \(Y^2 = [y_1^2, \ldots, y_N^2]^\T \in {\mathbb{R}^{N \times {d_2}}}\).
As shown in Fig.~\subref*{fig:independent}, the aim of standard multimodal GPLVM is to learn a single latent variable \(X \in {\mathbb{R}^{N \times q}}\) (where \(q \ll \min \left( {{d_1},{d_2}} \right)\)), which is used to explain the implicit alignment between each pair
of instances from $Y^1$ and $Y^2$.

It is assumed that two data modalities for multimodal GPLVMs are independently generated from the same latent variable $X$ via two sets of GP mappings. We can obtain the latent variable $X$ by learning the joint distribution over both data modalities, {\it i.e.}, $p\left( {Y^1, Y^2 \left| X \right.}\right)$.
With appropriate factorizations, the joint marginal likelihood of the full generative model can be written in closed form as
\begin{equation}\label{eq:1}
\resizebox{0.75\hsize}{!}{$
\begin{split}
p\left( {Y^1, Y^2 \left| {X,{\theta}} \right.} \right) & = p\left( {Y^1\left| {X,{\theta ^1}} \right.} \right)p\left( {Y^2\left| {X,{\theta ^2}} \right.} \right) \\
 & = \int {p\left( {Y^1\left| F^1 \right.} \right)p\left( {F^1\left| X, {\theta ^1} \right.} \right)dF^1}\\
& \cdot \int {p\left( {Y^2\left| F^2 \right.} \right)p\left( {F^2\left| X, {\theta ^2} \right.} \right)dF^2},
\end{split}
$}
\end{equation}
where $F^1$ and $F^2$ are GP mapping functions defined by separate kernel or covariance functions with hyperparameters $\theta = \{\theta ^1, \theta ^2\}$.
For multimodal GPLVMs, the inputs to the GP covariance functions, {\it i.e.}, $X$, are latent representations. In this paper, we treat both $X$ and $\theta$ as latent model parameters. Let $c \in \{1,2\}$ denote the reference to data modalities in subsequent sections.

As a typical way to train multimodal GPLVMs, maximum a posteriori (MAP) has been applied to learn the latent inputs $X$ and the kernel hyperparameters $\theta$. Traditionally, separate prior distributions, {\it i.e.}, $p\left( X \right)$, $p\left( \theta ^1\right)$ and $p\left( \theta ^2\right)$, are imposed on the parameters. Thus, the parameters of the standard mGPLVM are learned together by minimizing the following negative log-posterior,
\begin{equation} \label{eq:2-2}
{\cal L} = \sum\limits_c {{{\cal L}_c} - \log p\left( \theta ^c \right)} - \log p\left( X \right),
\end{equation}
where $\mathcal {L}_c$ is the corresponding negative log-likelihood of $p\left( {Y^c\left| {X,{\theta ^c}} \right.} \right)$, $c \in \{1,2\}$, and can be computed according to
\begin{equation} \label{eq:3}
 \mathcal {L}_c = \frac{d_c}{2}\ln \left| K_c(X, \theta ^c) \right| + \frac{1}{2}\text{tr}\left( {{(K_c(X, \theta ^c))^{ - 1}}Y^c{(Y^c)^\T}} \right),
\end{equation}
where $K_c(X, \theta ^c) = k_c(X, X)$ is the {\it covariance matrix} defined by a kernel function $k_c$ with hyperparameters $\theta ^c$. For notational simplicity, we use $K_c$ to denote $K_c(X, \theta ^c)$.

The standard mGPLVM is incapable of preserving the topological structure of high dimensional multimodal data in the function embedding process. To maximize the consistency to the modality-specific topologies, similarity GPLVM (m-SimGP)~\cite{iccv/SongWHT15} is proposed for multimodal learning, which assumes that the intra-modal similarities are generated from a shared latent space through multimodal GP mappings. For simplicity, Gaussian kernel is used in \cite{iccv/SongWHT15} to compute intra-modal similarities \({S^1} \in {\mathbb{R}^{N \times N}}\) and  \({S^2} \in {\mathbb{R}^{N \times N}}\), \ie, $S^{1}( y_i^1,y_j^1) = \exp \left( {- \|y_i^1 - y_j^1 \|^2} / {2{\gamma _1}}\right)$, $S^{2}( y_i^2,y_j^2) = \exp \left( {- \| y_i^2 - y_j^2 \|^2} / {2{\gamma _2}}\right)$, where $\gamma _1, \gamma _2 >0$.

Similar to multimodal GPLVM, the joint marginal likelihood of m-SimGP over $X$ and $\theta$ is given by,
\begin{align}
&p\left( {{S^1},{S^2}\left| {X,{\theta}} \right.} \right) = p\left( {{S^1}\left| {X,{\theta ^1}} \right.} \right)p\left( {{S^2}\left| {X,{\theta ^2}} \right.} \right),  \\
&p(S^c\left| {X,{\theta ^c} } \right.) =
 \frac{1}{\mathcal {A}^c}\exp \left( { - \frac{1}{2}{\text{tr}}\left( {K_c^{ - 1}{S^c}{{(S^c)}^\T}} \right)} \right), \label{eq:2s}
\end{align}
where $\mathcal {A}^c = {\sqrt {{{\left( {2\pi } \right)}^{N^2}}{{\left| K_c \right|}^N}} }$, $c \in \{1,2\}$. Likewise, the point estimation over
the latent representation $X$ and the kernel hyperparameters $\theta $ can be learned by optimizing an MAP objective function with priors.

\subsection{Regularized learning for GPLVM}
\label{sec:DR}

To enhance model flexibility in dealing with real-world problems, regularized GPLVMs have been proposed, which provide smoother estimation of the latent model parameters. Typically, regularization in Bayesian GPLVM is achieved by specifying a prior over $X$ and $\theta$, and then averaging over the posterior.

Current GPLVM approaches generally treat latent variable $X$ and kernel hyperparameters $\theta$ independently, combined with individual priors.
For example, the kernel hyperparameters are assigned with a prior distribution $p\left( \theta \right)$ for GP regression in \cite{nips/TitsiasL13}.
A prior $p\left( X \right)$ is added to the latent space for GPLVM , which can be instantiated as back constraint~\cite{icml/LawrenceC06,li2017shared}, class prior~\cite{Urtasun07discriminativegaussian}, locally linear embedding prior~\cite{icml/UrtasunFGPDL08}, discriminative shared-space prior~\cite{tip/EleftheriadisRP15}, distance-preserved constraint~\cite{song2017multimodal}, \emph{etc}.

In the m-RSimGP work~\cite{iccv/SongWHT15}, cross-modal semantic constraints are imposed over the latent representation $X$, and the constraints are derived from the following problem:
\begin{equation}\label{eq:4-1}
\begin{split}
&\mathop {\min }\limits_X \ \sum\nolimits_{\left( {{o_i}, {o_j}} \right) \in \mathscr {S}} {{{\left\| {{x_i} - {x_j}} \right\|}^2}} \\
&{\text{s.t.}} \quad {\left\| {{x_i} - {x_j}} \right\|^2} \geqslant 1,  \forall \left( {{o_i}, {o_j}} \right) \in \mathscr {D},
\end{split}
\end{equation}
where ${x_i}$ is the representation of the data point \({o_i} = \left\{ {{y_i^1},{y_i^2}} \right\}\) in the latent space.
$\mathscr {S} = \{\left({o_i},{o_j}\right)\}$ denotes the set of object pairs with similar semantics, and $\mathscr {D} = \{\left({o_i},{o_j}\right)\}$ denotes the set of object pairs with dissimilar semantics.
We can derive the learning objective of m-RSimGP by incorporating Eq.~\eqref{eq:4-1} into the likelihood maximization model in m-SimGP.

Without exception, all the above-mentioned regularization techniques are either imposed on $X$ independently, or derived from the observation space to the latent space $X$.

\section{Harmonized Multimodal Learning}
\label{harmonized}
Beyond existing individual learning for the latent representations $X$ and the kernel hyperparameters $\theta$ in multimodal GPLVM, we propose a joint learning strategy which can guarantee a more structural consistent solution by encouraging sufficient information sharing among different GPLVM components.

As shown in Fig.~\subref*{fig:joint}, we assume that the joint prior distribution $p\left( \theta ^1, \theta ^2, X \right)$ factorizes as a product of factors $p\left(\theta ^c, X \right)$,
\begin{equation}\label{eq:factor}
  p\left( {\theta ^1},{\theta ^2},X \right) = {1 \over Z}p\left( {\theta ^1},X \right)p\left( {\theta ^2},X \right),
\end{equation}
where $Z$ is a normalization constant to guarantee that the joint distribution $p\left( {\theta ^1},{\theta ^2},X \right)$ is properly normalized.
The joint learning of the shared latent representation and kernel hyperparameter is accomplished by minimizing the negative log-posterior,
\begin{equation}\label{eq:3-1}
\begin{split}
  & \arg \mathop {\min }\limits_{X,\theta } \sum\limits_c {{{\cal L}_c} - \log p\left( {{\theta ^1},{\theta ^2},X} \right)}  \\
  &  = \arg \mathop {\min }\limits_{X,\theta } \sum\limits_c {{{\cal L}_c} - \log p\left( {{\theta ^c},X} \right)}.
\end{split}
\end{equation}
In this paper, we call the factorable joint prior $p\left( \theta ^1, \theta ^2, X \right)$ as the harmonization prior over latent model parameters.
Different from Eq.~\eqref{eq:2-2} where the three kinds of latent model parameters are assumed to be mutually independent, direct connections are built among $X$ and $\theta ^c$ by the factor $p\left(\theta ^c, X \right)$ of the harmonization prior.
Next, we define the factors using different types of harmonization constraints over the modality-specific kernels and the similarity matrix in the latent space.

\subsection{The general formulation of harmonization}
\label{ssec:hmconstraint}
Multimodal learning aims to reduce the divergence between similarity in the observed space and the latent space.
In the context of GPLVM, the similarity structure among observed data can be characterized by the GP covariance kernel. Accordingly, we propose to encourage the agreement between the modality-specific kernels and the similarity of latent positions for multimodal data structure preservation. This results in the following \emph{harmonization constraints}:
\begin{equation}\label{eq:3-2}
\mathcal {H}_c({K_c}, {S^x})\le {\rho _c}, \quad c \in \{1, 2\},
\end{equation}
where $\mathcal {H}_c(\cdot, \cdot)$ is called the harmonization function.
The covariance matrices \(K_1\) and \(K_2\) are determined by the shared latent inputs and their respective kernel hyperparameters $\theta^1$ and $\theta^2$.
Since we have little prior knowledge on selecting the optimal similarity formulation for the harmonization function, we adopt the widely used similarity measure from exponential family to define \(S^x \in {\mathbb{R}^{N \times N}}\), which is computed on pairwise distance between latent positions, \ie, $S^{x}( x_i,x_j) = \exp \left( {- d^2(x_i, x_j) } / {2{\gamma _x}}\right)$, where $\gamma _x > 0$.
Parameters ${\rho _1, \rho _2 > 0}$ control the divergence between modality-specific GP kernels and the similarity of shared latent points.

The proposed constraints in Eq.~\eqref{eq:3-2} make the modality-specific kernels harmonize with the similarity in the shared latent space,
and thus enforce the agreement between GP kernels ($K_1$ and $K_2$) for different modalities. Therefore, it can preserve the inter-modal consistency among the similarity structures in the observed data space.

We formulate the harmonization constraint as a factor over the kernel hyperparameter $\theta ^c$ and the latent representation $X$, written as
\begin{equation}\label{eq:303}
  p\left( {{\theta ^c},X} \right) = \exp \left( { - \frac{1}{{\sigma _c^2}}{\mathcal {H}_c}\left( {{K_c},{S^x}} \right)} \right),
\end{equation}
where $\sigma _c^2$ is a global scaling of the factor $p(\theta ^c, X)$. By combining the harmonization prior in Eq.~\eqref{eq:factor} and the multimodal likelihoods, latent model parameters can be obtained by minimizing the following problem:
\begin{equation}\label{eq:304}
\sum\limits_c {\mathcal {L}\left( {{F^c}, \mathcal{D}} \right)}  + {\mu _c}{\mathcal {H}_c}\left( {{K_c},{S^x}} \right),
\end{equation}
where ${\mu _c} = \frac{1}{{\sigma _c^2}}$ controls the tradeoff between data fitting and the penalty on the similarity divergence. The penalty term is minimized to enforce the constraint in Eq.~\eqref{eq:3-2} to be satisfied.
$\mathcal {L}\left( F^c, \mathcal{D} \right)$ is the negative log-likelihood of the respective data modality. In particular, $\mathcal {L}\left( F^c, \mathcal{D} \right)$ is given by Eq.~\eqref{eq:3} in the standard mGPLVM.
From the probabilistic view point, the GP mappings $F^1$ and $F^2$ are no longer conditionally independent given the latent inputs $X$. Consequently, data from different modalities are more closely related in the harmonized multimodal learning framework.

It can be noted that we give a general formulation of the objective function in Eq.~\eqref{eq:304}, where $c$ does not have to be limited to  $\{1, 2\}$.
Therefore, based on computational resources available for processing, we can easily extend the proposed approach to more than two modalities by adding new negative log-likelihoods and new harmonization regularizers among $S^x$ and $K^c, c>2$.

\begin{figure*}[!t]
\centering
    \subfloat[hmGPLVM]{\includegraphics[height=2.4cm]{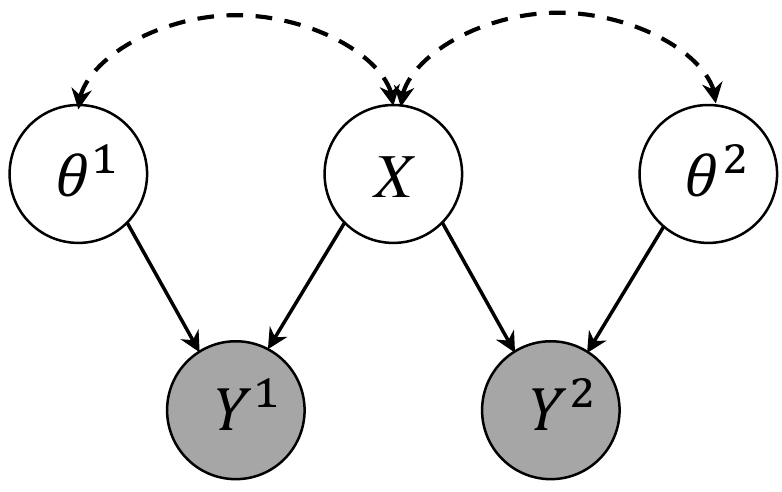}\label{fig:hmgplvm}}
    \hspace{0.04\textwidth}
    \subfloat[hm-SimGP]{\includegraphics[height=2.5cm]{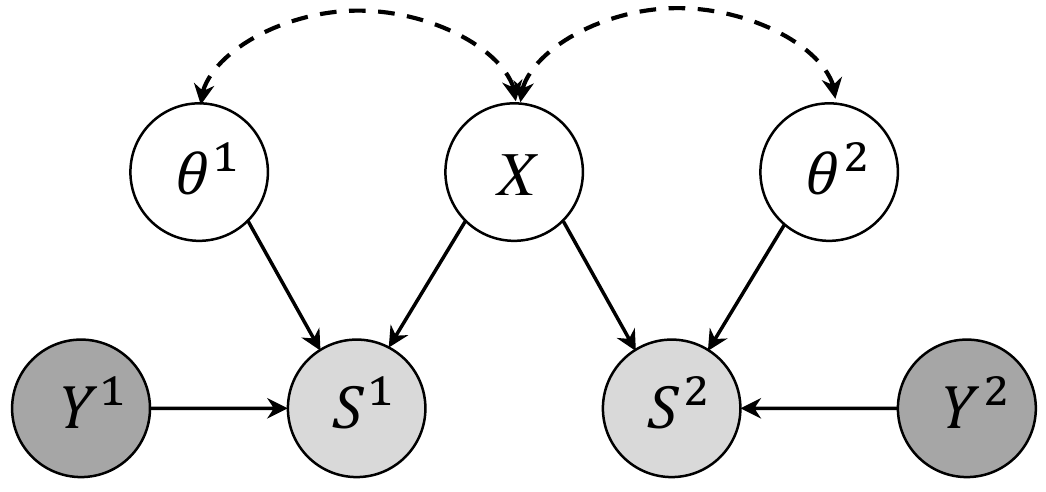}\label{fig:hm-simgp}}
    \hspace{0.04\textwidth}
    \subfloat[hm-RSimGP]{\includegraphics[height=2.9cm]{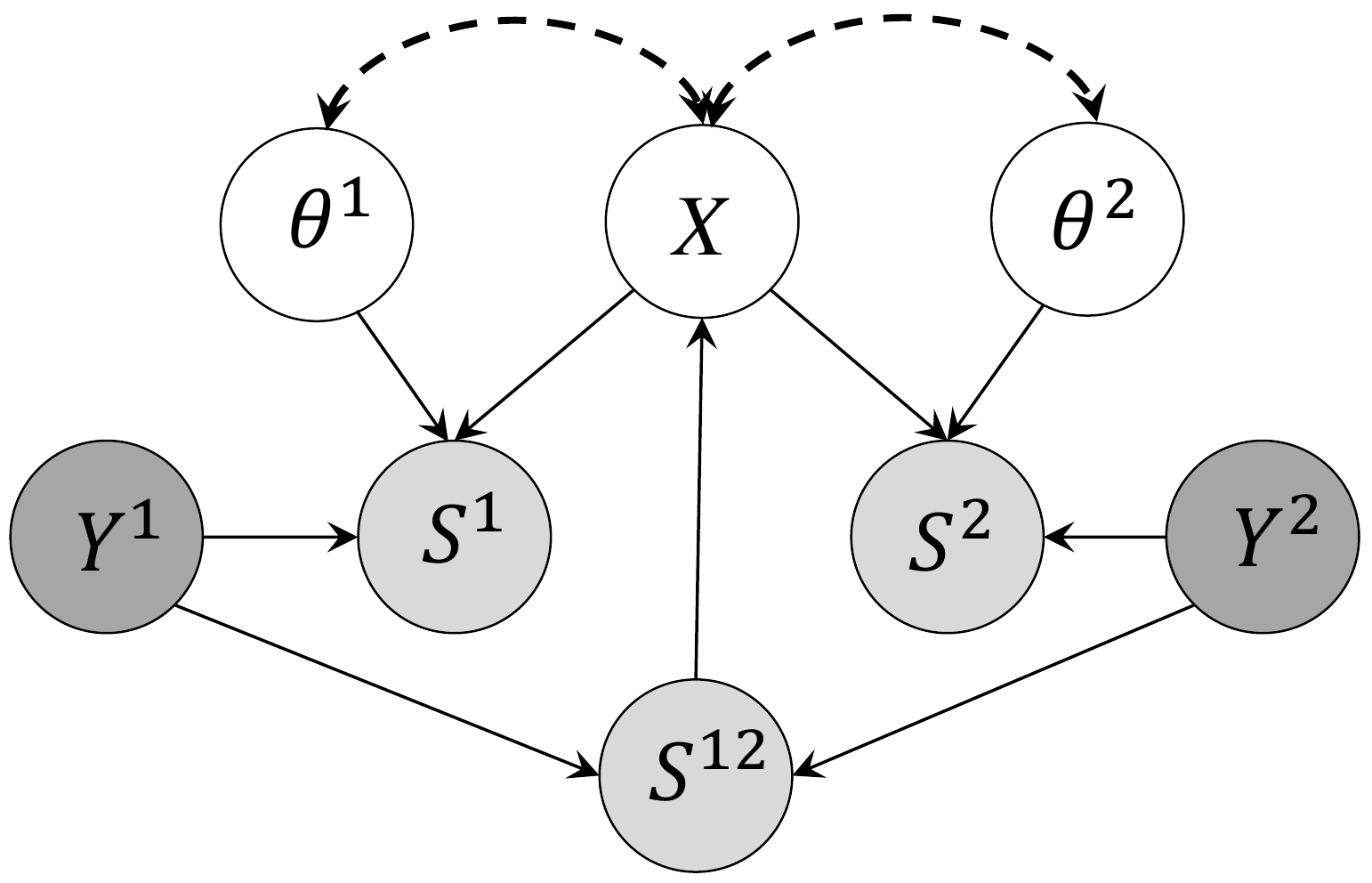}\label{fig:hm-rsimgp}}
\caption{The proposed multimodal GPLVMs with harmonization.} \label{fig:hm-models}
\end{figure*}

\subsection{Examples of the harmonization function}
\label{ssec:examples}
In the GPLVM framework, the optimization of latent representations and hyperparameters is a challenging problem. Based on this, choosing a well-behaved harmonization function does not make the optimization problem more difficult.
Here we provide a brief discussion on the choice of the harmonization function $\mathcal {H}_c$.
For the gradient optimization method, any convex and sub-differentiable function can be used to define harmonization functions $\mathcal {H}_c$.

This work presents two kinds of convex harmonization functions, one is distance-based and the other is ratio-based.
The distance-based harmonization function is defined on the difference between the kernel matrix and the latent similarity matrix, {\it i.e.}, $K_c - S^x$. In this case, two popular matrix norms, the Frobenius norm (F-norm) $\|\cdot\|_F$ and the $l_{2,1}$-norm $\|\cdot\|_{2,1}$ are applied.
\begin{itemize}
  \item The F-norm harmonization function:
    \begin{equation}\label{eq:3-F}
    \mathcal{H}_c({K_c}, {S^x}) = \left\| {{K_c} - {S^x}} \right\|_F^2,
    \end{equation}
    where the F-norm $\|\cdot\|_F$ is the element-wise Euclidean norm on the matrix ${K_c} - {S^x}$.
    The F-norm harmonization constraint ensures that the solution to the kernels $K_c$ should be in the vicinity of $S^x$ measured by the Frobenius norm. Besides, the kernels can also be seen as noisy observations of $S^x$.

  \item The $l_{2,1}$-norm harmonization function:
    \begin{equation}\label{eq:3-L21}
    \mathcal{H}_c({K_c}, {S^x}) = \left\| {{K_c} - {S^x}} \right\|_{2,1},
    \end{equation}
    where the $l_{2,1}$-norm $\|\cdot\|_{2,1}$ is the sum of the Euclidean norms of the columns of the matrix ${K_c} - {S^x}$. Compared to F-norm, the $l_{2,1}$-norm, as a loss/error function, is more robust to outliers or large variations distributed in the input data.
\end{itemize}

The ratio-based harmonization function is defined on the ``division'' between the kernel matrix and the latent similarity matrix, {\it i.e.}, ${K_c}^{-1}{S^x}$. The trace function $\textrm{tr}(\cdot)$ is chosen to operate on the ratio of the two matrices.
\begin{itemize}
  \item The trace harmonization function:
    \begin{equation}\label{eq:3-R}
    \mathcal{H}_c({K_c}, {S^x}) = \frac{1}{2}\textrm{tr}\left({K_c}^{-1}{S^x}\right),
    \end{equation}
    where the trace function $\textrm{tr}(\cdot)$ returns the sum of diagonal entries of the square matrix ${K_c}^{-1}{S^x}$. From an information theoretic perspective, the trace harmonization function is related to the KL-divergence between Gaussian distributions $\mathcal{N}(0, K_c)$ and $\mathcal{N}(0, S^x)$, and the trace constraint forces the two distributions to be close.
\end{itemize}

With a sub-differentiable convex harmonization function, the factor part in the above problem~\eqref{eq:304} can be solved using convex optimization algorithms.
We will show by experiments that each of the three harmonization formulations shows distinguished model learning behaviors on different datasets. In practice, the harmonization function can also be designed for specific application needs.
It is important to clarify that there still are some types of sub-differentiable and convex functions which can turn the joint harmonization prior into a trivial prior. For example, if the harmonization function is a constant function, \ie, $\mathcal {H}_c (\cdot, \cdot) = const$, there is no informative prior over the parameters of the model. If the output of the harmonization function can be formulated as a function of the latent representation or the hyperparameters, \ie, $ \mathcal {H}_c (\cdot, \cdot) = f(X)$ or $\mathcal {H}_c (\cdot, \cdot) = g(\theta ^c)$, the harmonization prior becomes an exclusive prior over $X$ or $\theta ^c$, and the model in Eq.~\eqref{eq:304} equals to the standard multimodal GPLVM.

\section{Multimodal GPLVMs with Harmonization}
\label{proposed}
We discuss a range of specific formulations of the problem~\eqref{eq:304} with the proposed harmonization constraints. To be specific, we consider three representative multimodal extensions of the GPLVM, \ie, mGPLVM~\cite{nips/ShonGHR05}, m-SimGP~\cite{iccv/SongWHT15}, and m-RSimGP~\cite{iccv/SongWHT15}, where the first is constructed on the original high dimensional representations and the latter two are constructed on the similarities in the observation space.
We formulate the harmonization constraints as priors over latent model parameters and incorporate them into the three multimodal GPLVM extensions respectively. We obtain three kinds of harmonized multimodal GPLVM methods, \ie, hmGPLVM, hm-SimGP, and hm-RSimGP, as shown in Fig.~\ref{fig:hm-models}.

\subsection{Harmonized multimodal GPLVM (hmGPLVM)}
As described in Section~\ref{sec:mgplvm}, the mGPLVM algorithm generalizes GPLVM to the multimodal case by assuming that the heterogeneous observed data $Y^1$ and $Y^2$ are generated from a shared latent space $X$. For harmonized multimodal GPLVM, the latent representation $X$ is learned by minimizing the negative log-posterior $\mathcal {L}$ given by Eq.~\eqref{eq:3-1}, where the joint prior over $X$ and $\theta$ is derived from the minimization of the harmonization function $\mathcal {H}_c$ in \eqref{eq:3-2}.
Specifically, we regularize the latent model parameters of the previous mGPLVM with the harmonization prior, and derive the harmonized multimodal GPLVM (hmGPLVM):
\begin{equation} \label{eq:3-21}
\arg \mathop {\min }\limits_{X,{\theta}} \sum\limits_c {{{\cal L}_c}{\rm{ + }}{\mu _c}{{\cal H}_c}({K_c}, {S^x})},
\end{equation}
where $c \in \{1, 2\}$. The negative log-likelihood function $\mathcal {L}_c$ is computed according to Eq.~\eqref{eq:3}. ${\mu _c}$ is the tradeoff parameter. The harmonization regularizer $\mathcal {H}_c$ provides an informative kernel-based way to align both $K_1$ and $K_2$ with $S^x$, and thus ensures consistency of the correlation structure among multimodal data.

\subsection{Harmonized similarity GPLVM (hm-SimGP)}
The similarity-based GPLVM extension m-SimGP is proposed in \cite{iccv/SongWHT15} which assumes that the latent space shared by heterogeneous data modalities is learned from the intra-modal similarities through GP functions.
Similar as hmGPLVM, we remove the square $l_2$-norm imposed on $X$ from the original m-SimGP~\cite{iccv/SongWHT15}, and incorporate the constraint in Eq.~\eqref{eq:3-2} to define the harmonization prior over both the latent and the kernel hyperparameter spaces:
\begin{equation} \label{eq:3-22}
\arg \mathop {\min }\limits_{X,{\theta}} \sum\limits_c {{{\cal L}_c^s}{\rm{ + }}{\mu _c}{{\cal H}_c}({K_c}, {S^x})},
\end{equation}
where $c \in \{1, 2\}$. The negative log-likelihood ${\mathcal {L}_c^s}$ can be computed from Eq.~\eqref{eq:2s}.
${\mu _1}$ and ${\mu _2}$ are tradeoff parameters.

The proposed hm-SimGP models the interaction among similarity structures in three different manifolds, {\it i.e.}, the latent similarity $S^x$ in the shared subspace, the intra-modal similarity $S^c$ in the observed data space, and the covariance $K_c$ for the GP mapping functions.
With the harmonization learning mechanism, the divergence between these similarities is encouraged to be small, and thus a more structural consistent representation can be learned for multimodal data.

\subsection{Harmonized m-RSimGP (hm-RSimGP)}
Compared to the previous m-SimGP model, the m-RSimGP algorithm~\cite{iccv/SongWHT15} incorporates an additional regularizer in Eq.~\eqref{eq:4-1} with semantic information of multimodal data.
The inter-modal semantic relation is used as a smooth prior over the latent positions to maximize cross-modal semantic consistency. In this work, we further add a harmonization regularizer as in Eq.~\eqref{eq:304} into the m-RSimGP, and obtain a new model hm-RSimGP as follows:
\begin{equation} \label{eq:3-23}
\resizebox{0.88\hsize}{!}{$
\begin{split}
\arg \mathop {\min }\limits_{X,{\theta}} & \sum\limits_c {{{\cal L}_c^s}{\rm{ + }}{\mu _c}{{\cal H}_c}({K_c}, {S^x})} \\
& + {\lambda_1}\sum\nolimits_{\left( {{o_i}, {o_j}} \right) \in \mathscr {S}} {{{\left\| {{x_i} - {x_j}} \right\|}^2}} \\
& + {\lambda_2}\sum\nolimits_{\left( {{o_i}, {o_j}} \right) \in \mathscr {D}} \mathop {\max }{{\left(0, 1-{\left\| {{x_i} - {x_j}} \right\|}^2\right)}},
\end{split}
$}
\end{equation}
where ${\mathcal {L}_c^s}, c \in \{1, 2\}$, is the negative log-likelihood of hm-RSimGP, which is also derived from Eq.~\eqref{eq:2s}, similar as hm-SimGP.
As stated in \cite{iccv/SongWHT15,ijcai/XieX13}, the similarity prior in the third term is used to make sure that semantically similar observations are close to each other in the latent space, by minimizing the distance between their latent space positions. The dissimilarity prior in the fourth term is used to make sure that semantically dissimilar observations are far from each other in the latent space, by keeping their latent space positions separated by a margin of 1.
The tradeoff parameters ${\mu _1}$ and ${\mu _2}$ are used to control the influence of the harmonization terms. ${\lambda_1}$ and ${\lambda_2}$ are used to control the influence of cross-modal semantic constraints.


\subsection{Optimization and inference}
\label{ssec:optim}
It is obvious that there are no closed-form solutions for the above objective functions, since the corresponding harmonized multimodal GPLVMs involve highly non-linear functions on the parameters ${\theta}$ and $X$.
In this work, convex subdifferentiable functions are chosen to define the regularization term $\mathcal {H}_c({K_c}, {S^x})$ over the latent model parameters.
Besides, the log-likelihood functions, {\it e.g.}, $\mathcal {L}_c$ in Eq.~\eqref{eq:3-21}, are differentiable on condition that the gradients of GP kernel functions can be derived with respect to the latent model parameters.
Therefore, we turn to a gradient-based optimizer such as scaled conjugate gradients (SCG)~\cite{nn/Moller93} to learn low dimensional representations for multimodal data. For all the above problems, we leverage sparsification technique and GPflow library~\cite{GPflow2017} for fast/big computation.
As in fast GPLVM~\cite{journals/jmlr/Lawrence05}, sparsification is performed by repeatedly selecting an active subset from all data objects to optimize the active points.
Specifically, for the harmonization regularization terms, we also employ the optimization scheme based on active matrix selection strategy for the kernel or similarity matrices to reduce the gradient computation complexity.
As a consequence, the model training is performed in the selected active set (with $M \ll N$ number of active points). The dominant complexity is $O(NM^2)$ for learning and inference.

Existing works provide different ways to reconstruct a given test data in the latent space, such as kernel regression based back-constraint~\cite{tip/EleftheriadisRP15} and autoencoder back-projection~\cite{li2017shared} using a GP prior.
Different from these methods based on back-mappings from the observation space to the latent space, we adopt a simple and straightforward MAP inference procedure for our models only through the GP mappings learned on the training multimodal data.
To be specific, given a test image $y^1_t$, its corresponding latent representation $x_t$ can be obtained by maximizing the posterior probability $p({x_t}|{y^1_t})$.

With the obtained latent representations, we can perform multimodal learning tasks to discover non-linear correlations among multimodal observations.
Specifically, for image-to-text cross-modal retrieval task, we retrieve from a text database for a given image query, by ranking the retrieved text data objects according to the distance among their positions in the common latent space.

\section{Experiments}
\label{sec:experiments}
In this section, we evaluate the proposed multimodal harmonization learning methods on four datasets for the task of cross-modal retrieval.
Our matlab code is available at \url{https://github.com/songguoli87/HMGP}.

\subsection{Datasets}
\label{ssec:data}
\textbf{PASCAL}~\cite{rashtchian2010collecting} contains a total of 1000 images collected from 20 categories of PASCAL 2008. Each image is annotated with 5 caption sentences by Amazon Mechanical Turk. We use a random 70/30 split of the dataset for training/testing. Images and associated sentences are represented in the same way as \cite{cvpr/PereiraV12}\footnote{\url{http://www.svcl.ucsd.edu/~josecp/files/ris_cvpr12.zip}\label{ptv}.}. For each image, the SIFT feature is first extracted, and then the bag-of-visual-words (BoVW) model is used to obtain a 1024-dim feature representation. For each text, the latent Dirichlet allocation (LDA) model is used to obtain a 100-dim feature representation.

\textbf{Wikipedia\footnote{\url{http://www.svcl.ucsd.edu/projects/crossmodal/}.}}~\cite{mm/RasiwasiaPCDLLV10} consists of 2,866 image-text pairs collected from Wikipedia articles. All the pairs are from 10 semantic categories, and each pair is categorized as one of them. We adopt the publicly available 128-dim BoVW image feature based on the SIFT descriptor and 10-dim LDA text feature. We randomly select 2,173 pairs for training and use the remaining 693 pairs for testing.

\textbf{TVGraz}~\cite{khan2009tvgraz} has totally 2,058 image-text pairs, which are collected from webpages retrieved by Google image search with keywords of the 10 categories from the Caltech-256 dataset. We still use the same feature representation as in \cite{cvpr/PereiraV12}\textsuperscript{\ref{ptv}}. Each image is represented by a 1024-dim BoVW vector based on the SIFT feature, and each text is represented by a 100-dim LDA feature vector. 1,558 document pairs are randomly chosen to produce a training set, and the rest pairs are used for testing.

\textbf{MSCOCO\footnote{\url{http://cocodataset.org/}.}}~\cite{lin2014microsoft} contains 82,783 images with 80 labels. Each image is also annotated by 5 independent sentences via Amazon Mechanical Turk.
After removing images without labels, we randomly select 10,000 image-text pairs for testing, and the remaining 72,081 pairs are used for training.
For image representation, we adopt 1,001-dim CNN feature extracted from logits layer in Inception-v4~\cite{aaai/SzegedyIVA17}. For representing text instances, we use 1,000-dim BoW vector with the TF-IDF weighting scheme.

For the first three datasets, each image-text pair is annotated by a single class label. For MSCOCO, each pair is associated to multiple class labels.
In the following experiments, we compare our proposed methods with various state-of-the-art approaches to verify the effectiveness.

\subsection{Experimental setting}
\label{ssec:baselines}
In the experiments, we have used all three different harmonization constraints defined in section~\ref{ssec:examples} with the proposed harmonized GP models.
As shown in Table~\ref{tab:pascal_results}, in total there are nine resulting models, \ie, hmGPLVM (F), hmGPLVM ($l_{2,1}$), hmGPLVM (tr), hm-SimGP (F), hm-SimGP ($l_{2,1}$), hm-SimGP (tr), hm-RSimGP (F), hm-RSimGP ($l_{2,1}$), and hm-RSimGP (tr), where ``F'', ``$l_{2,1}$'' and ``tr'' denote the F-norm constraint, the $l_{2,1}$-norm constraint, and the trace constraint, respectively.
Our methods work for any differentiable kernel functions. For simplicity, a radial basis function (RBF) kernel perturbed by a bias term and an additive white noise is used to define the non-linear covariance matrices \(K_1\) and \(K_2\). For the similarity $S^x$, we use Euclidean distance to compute it, and set $\gamma _x$ to be 1.

For mGPLVM and m-SimGP, their log-likelihood functions are penalized by a simple Gaussian prior over the latent space $X$, {\it i.e.}, \(p\left( X \right) = \prod\nolimits_{n = 1}^N {\mathcal N\left( {{x_n}\left| {0,{\rm{I}}} \right.} \right)} \).
For the baseline m-RSimGP, inter-modal relations ({\it i.e.}, semantic similarity and dissimilarity) are imposed as a smooth prior over the latent space $X$. There is no informative prior over the kernel hyperparameters $\theta$ for all these baselines.
In our experiments, we set the number of inducing points $M$ to be $100$. On the Wiki dataset, hmGPLVM(tr) takes 291.68s for model learning on an ordinary desktop PC, hm-SimGP(tr) takes 558.41s, and hm-RSimGP(tr) takes 950.70s, while mGPLVM takes 235.61s, m-SimGP takes 500.36s, and m-RSimGP takes 932.85s.

In all experiments, we use the optimal settings of the parameters tuned by a parameter validation process. The tradeoff parameters ${\mu _1}$ and ${\mu _2}$ are assigned with the same value, indicating equal importance of different data modalities.
For each dataset, we employ CCA \cite{neco/HardoonSS04} to initialize the shared space with the nearly optimal latent feature dimension.

\subsection{Perfomance on Cross-Modal Retrieval}
\label{ssec:retrieval}
\begin{table}[t]
  \caption{Cross-modal retrieval comparison in terms of mAP on PASCAL dataset.}
  \label{tab:pascal_results}
  \centering
  \begin{tabular}{l c c c}
    \toprule
    Methods & I$\rightarrow$T & T$\rightarrow$I & Average \\
    \midrule

    JFSSL & 0.2263 & 0.1857 & 0.2060 \\
    LGCFL & 0.2570 & 0.2379 & 0.2475 \\
    RL-PLS & 0.2140 & 0.1659 & 0.1900 \\
    CCQ & 0.1663 & 0.3625 & 0.2644 \\
    MLBE & 0.2543 & 0.2215 & 0.2379 \\
    DCCAE & 0.1988 & 0.1670 & 0.1829 \\
    DS-GPLVM & 0.2079 & 0.1797 & 0.1938 \\
    m-DSimGP & 0.2903 & 0.2848 & 0.2876 \\

    \midrule
    mGPLVM & 0.1507 & 0.1318 & 0.1413 \\
    hmGPLVM (F) & 0.1755 & 0.1471 & 0.1613 \\
    hmGPLVM ($l_{2,1}$) & 0.1846 & 0.1503 & 0.1675 \\
    hmGPLVM (tr) & \textbf{0.1950} & \textbf{0.1514} & \textbf{0.1732} \\

    \midrule
    m-SimGP & 0.2860 & 0.2818 & 0.2839 \\
    hm-SimGP (F) & 0.2993 & 0.3074 & 0.3034  \\
    hm-SimGP ($l_{2,1}$) & 0.3168 & 0.3202 & 0.3185 \\
    hm-SimGP (tr) & \textbf{0.3279} & \textbf{0.3283} & \textbf{0.3281} \\

    \midrule
    m-RSimGP & 0.3301 & 0.3275 & 0.3288 \\
    hm-RSimGP (F) & 0.3538 & 0.3514 & 0.3526 \\
    hm-RSimGP ($l_{2,1}$) & 0.3626 & 0.3594 & 0.3610 \\
    hm-RSimGP (tr) & \textbf{0.3823} & \textbf{0.3837} & \textbf{0.3830} \\
    \bottomrule
  \end{tabular}
\end{table}

We evaluate the performance of our methods for two cross-modal retrieval tasks: (1) I$\rightarrow$T: image query vs. text database; and (2) T$\rightarrow$I: text query vs. image database.
The experimental results indicate that the proposed harmonization prior performs very well in enhancing the consistency of cross-modal representations learned by the resulting GP models.

The performance of cross-modal retrieval is measured with mean average precision (mAP)~\cite{Manning:2008:IIR:1394399}, \ie, the mean of the average precision
of all the queries. AP of a set of $R$ retrieved data can be calculated by $\textrm{AP} = \frac{1}{T}\sum\nolimits_{r = 1}^R {p(r)\cdot rel(r)}$, where $T$ is the number of relevant items in the retrieved set, $p(r)$ is the precision of the top $r$ retrieved items. If the $r$-th retrieved result has the same label or shares at least one label with the query, $rel(r)$ is set to be 1, otherwise 0. In this paper, we present the results of mAP score on all retrieved data. The larger the mAP, the better the performance.
Besides the mAP, we also use precision-recall curve~\cite{pami/PereiraCDRLLV14} for comprehensive evaluation.

\subsubsection{Compared methods}
The proposed models are compared with their respective baselines, {\it i.e.}, mGPLVM~\cite{nips/ShonGHR05}, m-SimGP~\cite{iccv/SongWHT15} and m-RSimGP~\cite{iccv/SongWHT15}. Besides, we also compare with a number of state-of-the-art multimodal learning methods.
JFSSL~\cite{journals/pami/WangHWWT16} is a linear subspace learning method which imposes the $l_{2,1}$-norm penalties on the projection matrices.
LGCFL~\cite{tmm/KangXLXP15} and RL-PLS~\cite{mm/HeMWLH16} are supervised cross-modal matching approaches which utilize labels to learn consistent feature representations from heterogeneous modalities.
CCQ~\cite{sigir/LongCWY16} and MLBE~\cite{kdd/ZhenY12} are parametric methods for discrete latent representation learning. CCQ employs a direct encoding for multimodal input by correlation-maximal mappings, and MLBE builds a probabilistic model to learn binary latent factors from intra-modal and inter-modal similarities.
DCCAE~\cite{icml/WangALB15} is a deep extension of the popular CCA for deep multimodal representation learning.
DS-GPLVM~\cite{tip/EleftheriadisRP15} and m-DSimGP~\cite{song2017multimodal} are two non-parametric probabilistic GPLVM-based models with different priors over the latent space. DS-GPLVM defines a discriminative shared-space prior using the data labels, and m-DSimGP defines a prior over the global similarity structure between latent points.

\subsubsection{Experimental results}
\label{ssec:results}

\begin{table}[t]
\renewcommand\arraystretch{1.3} 
  \caption{Cross-modal retrieval comparison in terms of mAP on Wiki and TVGraz datasets.}
  \label{tab:wiki_tvgraz_results}
  \centering
  \resizebox{0.48\textwidth}{!}{
  \begin{tabular}{l ccc l ccc}
    \toprule
   \multirow{2}{*}{Methods} & \multicolumn{3}{c}{Wiki} && \multicolumn{3}{c}{TVGraz} \\
    \cmidrule{2-4}
    \cmidrule{6-8}

     & I$\rightarrow$T & T$\rightarrow$I & Average && I$\rightarrow$T & T$\rightarrow$I & Average \\
     \midrule

    JFSSL & 0.3063 & 0.2275 & 0.2669 && 0.4076 & 0.4202 & 0.4139 \\
    LGCFL & 0.2736 & 0.2241 & 0.2489 && 0.4366 & 0.4140 & 0.4253 \\
    RL-PLS & 0.3087 & 0.2435 & 0.2761 && {\bf 0.5737} & 0.5478 & {\bf 0.5608} \\
    CCQ & 0.2470 & 0.3944 & 0.3207 && 0.3773 & 0.5100 & 0.4436 \\
    MLBE & 0.3787 & 0.4109 & 0.3948 && 0.3468 & 0.3849 & 0.3659 \\
    DCCAE & 0.2542 & 0.1916 & 0.2229 && 0.3879 & 0.3736 & 0.3808 \\
    DS-GPLVM & 0.2822 & 0.2147 & 0.2485 && 0.4817 & 0.4554 & 0.4686 \\
    m-DSimGP & 0.4470 & 0.4242 & 0.4356 && 0.4659 & 0.4667 & 0.4663 \\

    \midrule
    mGPLVM & 0.2054 & 0.1628 & 0.1841 && 0.2645 & 0.2784 & 0.2715 \\
    hmGPLVM (F) & 0.2392 & 0.1826 & 0.2109 && 0.3572 & 0.3227 & 0.3400 \\
    hmGPLVM ($l_{2,1}$) & 0.2449 & 0.1804 & 0.2127 && 0.3688 & 0.3330 & 0.3509 \\
    hmGPLVM (tr) & \textbf{0.2546} & \textbf{0.1911} & \textbf{0.2229} && \textbf{0.3702} & \textbf{0.3387} & \textbf{0.3545} \\

    \midrule
    m-SimGP & 0.4336 & 0.4188 & 0.4262 && 0.4467 & 0.4453 & 0.4460 \\
    hm-SimGP (F) & 0.4557 & 0.4391 & 0.4474 && 0.4647 & 0.4633 & 0.4640 \\
    hm-SimGP ($l_{2,1}$) & \textbf{0.4738} & 0.4494 & \textbf{0.4616} && 0.4680 & 0.4650 & 0.4665 \\
    hm-SimGP (tr) & 0.4594 & \textbf{0.4582} & 0.4588 && \textbf{0.4767} & \textbf{0.4727} & \textbf{0.4747} \\

    \midrule
    m-RSimGP & 0.4697 & 0.4418 & 0.4558 && 0.5102 & 0.5079 & 0.5091 \\
    hm-RSimGP (F) & 0.4861 & 0.4791 & 0.4826 && 0.5435 & 0.5351 & 0.5393 \\
    hm-RSimGP ($l_{2,1}$) & 0.4837 & 0.4870 & 0.4854 && \textbf{0.5531} & 0.5499 & 0.5515 \\
    hm-RSimGP (tr) & \textbf{0.5108} & \textbf{0.5095} & \textbf{0.5102} && 0.5484 & \textbf{0.5628} & \textbf{0.5556} \\
    \bottomrule
  \end{tabular}}
\end{table}

\begin{table}[b]
  \renewcommand\arraystretch{1.1} 
  \caption{Cross-modal retrieval comparison in terms of mAP on MSCOCO dataset.}
  \label{tab:coco_results}
  \centering
  \begin{tabular}{l c c c}
    \toprule
    Methods & I$\rightarrow$T & T$\rightarrow$I & Average \\
    \midrule

    CCA & 0.5801 & 0.5784 & 0.5793 \\
    FSH & 0.6262 & 0.6259 & 0.6261 \\
    3V-CCA & 0.6248 & 0.6196 & 0.6222  \\
    ml-CCA & 0.6288 & 0.6336 & 0.6312  \\
    DCCAE & 0.6206 & 0.6233 & 0.6220 \\

    \midrule
    mGPLVM & 0.5637 & 0.5624 & 0.5631 \\
    hmGPLVM (F) & 0.6071 & 0.6315 & 0.6193 \\
    hmGPLVM ($l_{2,1}$) & 0.6086 & \textbf{0.6372} & 0.6229 \\
    hmGPLVM (tr) & \textbf{0.6179} & 0.6364 & \textbf{0.6272} \\

    \midrule
    m-SimGP & 0.6074 & 0.6169 & 0.6122 \\
    hm-SimGP (F)  & 0.6150 & 0.6383 & 0.6267 \\
    hm-SimGP ($l_{2,1}$) & 0.6199 & 0.6433 & 0.6316 \\
    hm-SimGP (tr) & \textbf{0.6208} & \textbf{0.6461} & \textbf{0.6335} \\

    \midrule
    m-RSimGP & 0.6485 & 0.6624 & 0.6555 \\
    hm-RSimGP (F) & 0.6506 & 0.6848 & 0.6677 \\
    hm-RSimGP ($l_{2,1}$) & 0.6503 & \textbf{0.6859} & 0.6681 \\
    hm-RSimGP (tr) & \textbf{0.6519} & 0.6844 & \textbf{0.6682} \\
    \bottomrule
  \end{tabular}
\end{table}

\begin{figure*}[!t]
\centering
    \subfloat[PASCAL: Image query]{\includegraphics[width=0.22\textwidth]{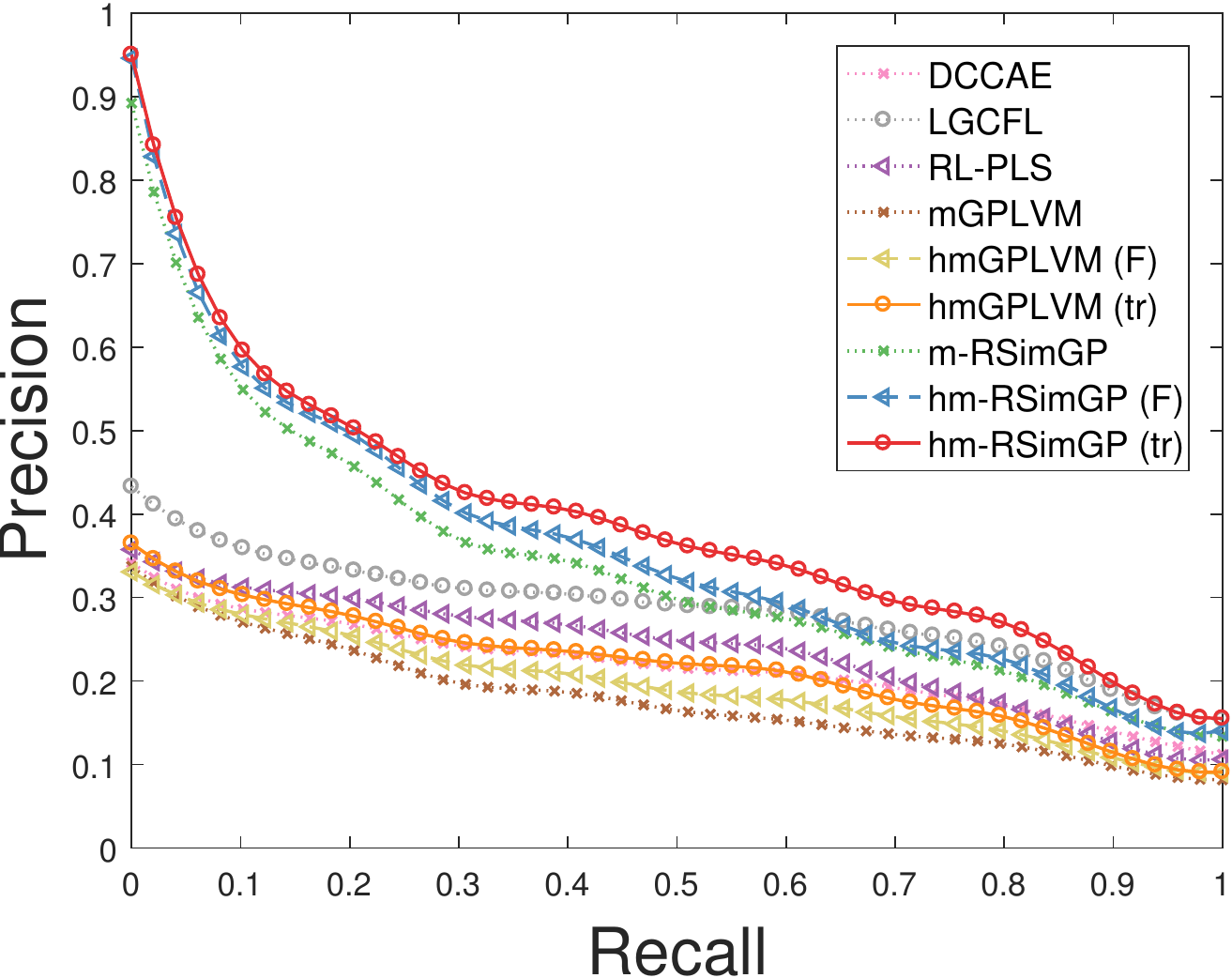}\label{fig:prcurve_im2txt_pascal}}
    \hfill
    \subfloat[PASCAL: Text query]{\includegraphics[width=0.22\textwidth]{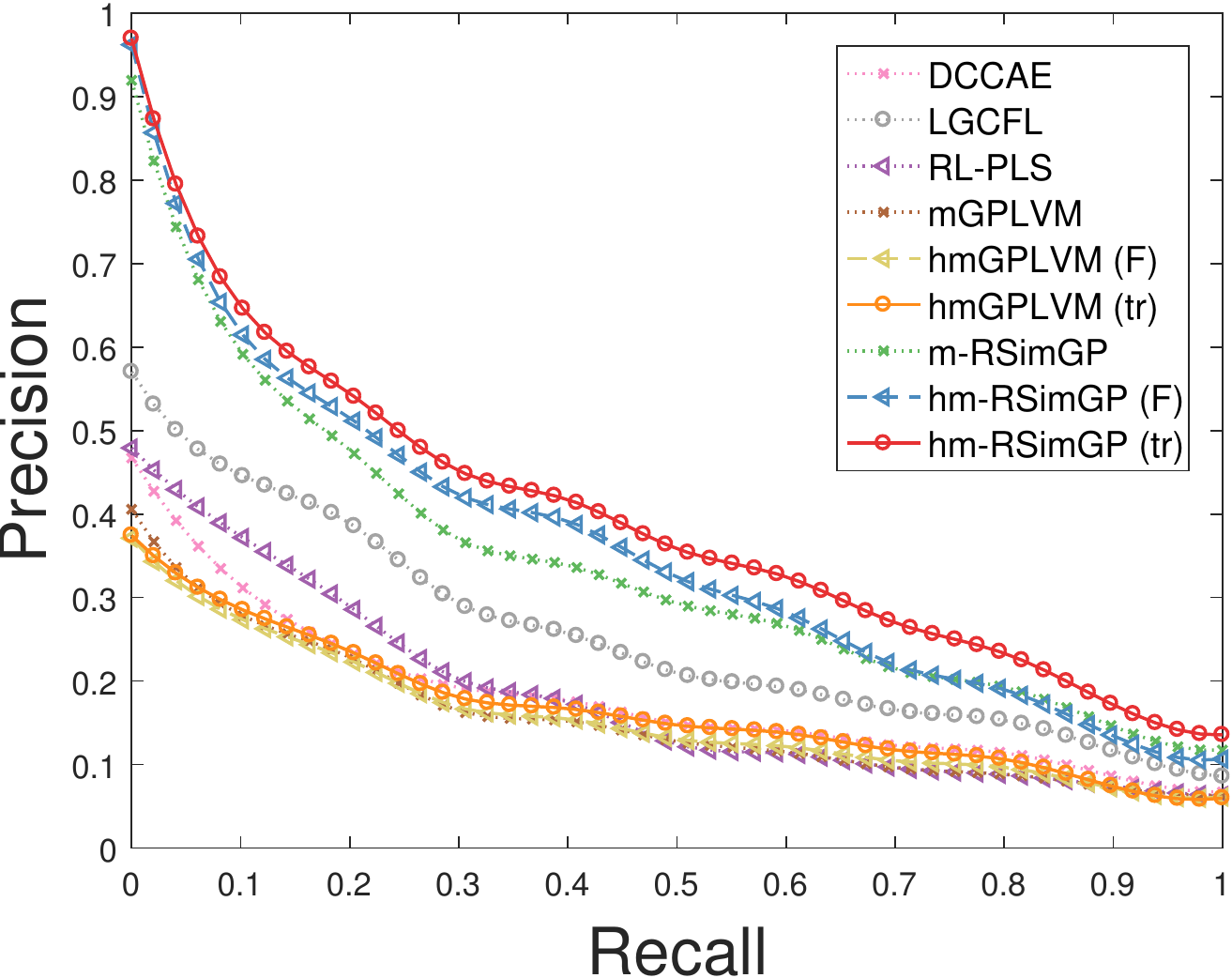}\label{fig:prcurve_txt2im_pascal}}
    \hfill
    \subfloat[Wiki: Image query]{\includegraphics[width=0.22\textwidth]{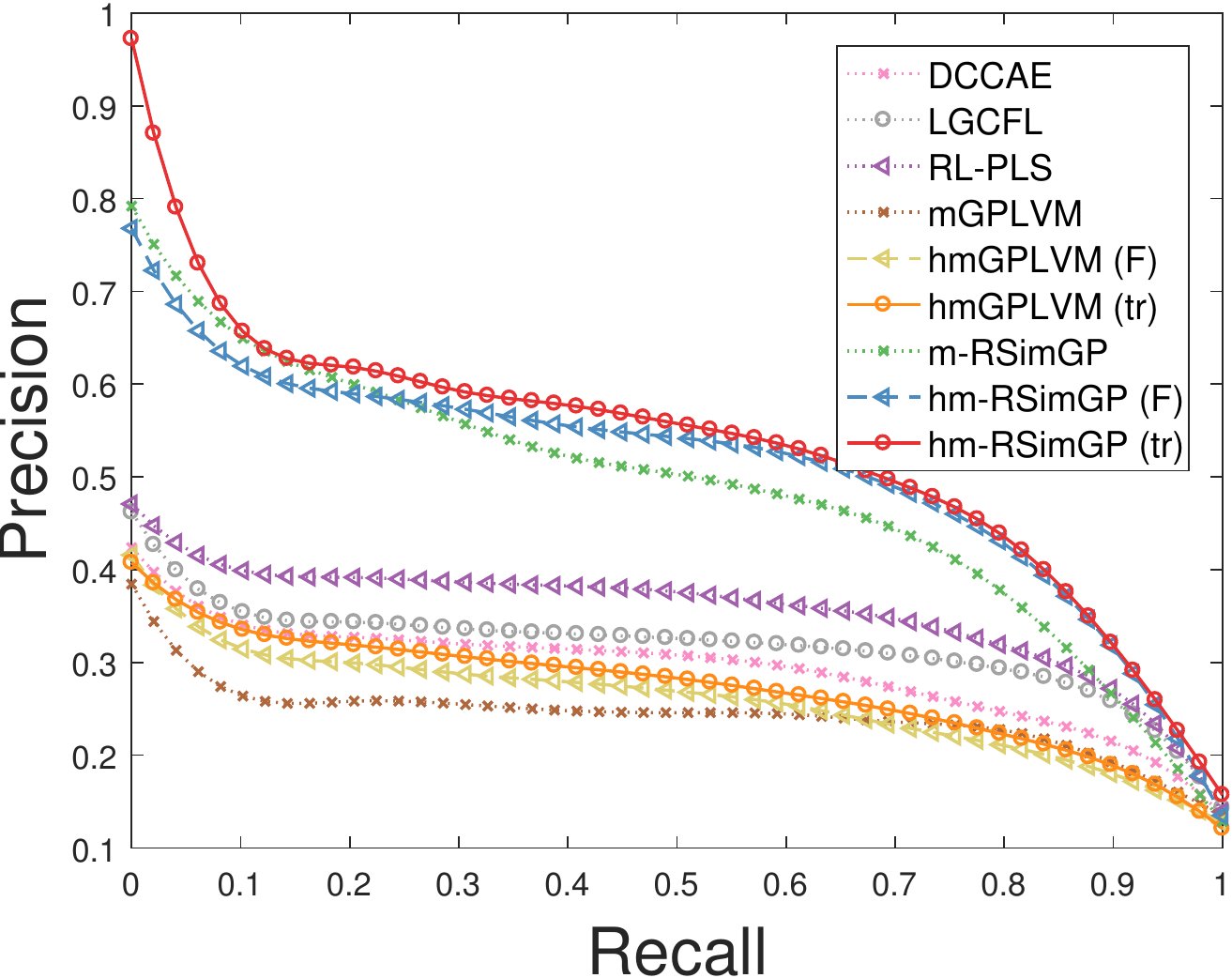}\label{fig:prcurve_im2txt_wiki}}
    \hfill
    \subfloat[Wiki: Text query]{\includegraphics[width=0.22\textwidth]{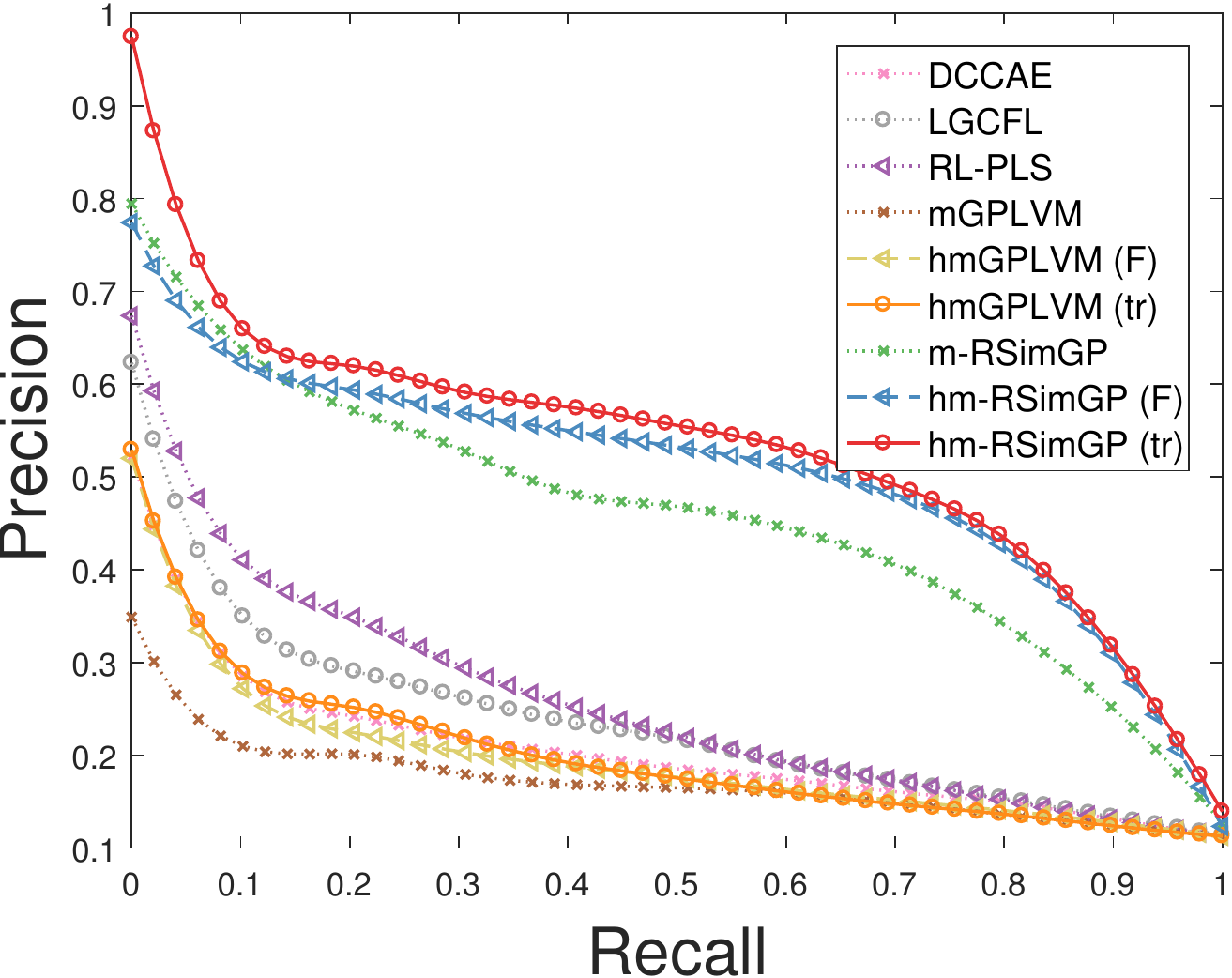}\label{fig:prcurve_txt2im_wiki}}
    \\
    \subfloat[TVGraz: Image query]{\includegraphics[width=0.22\textwidth]{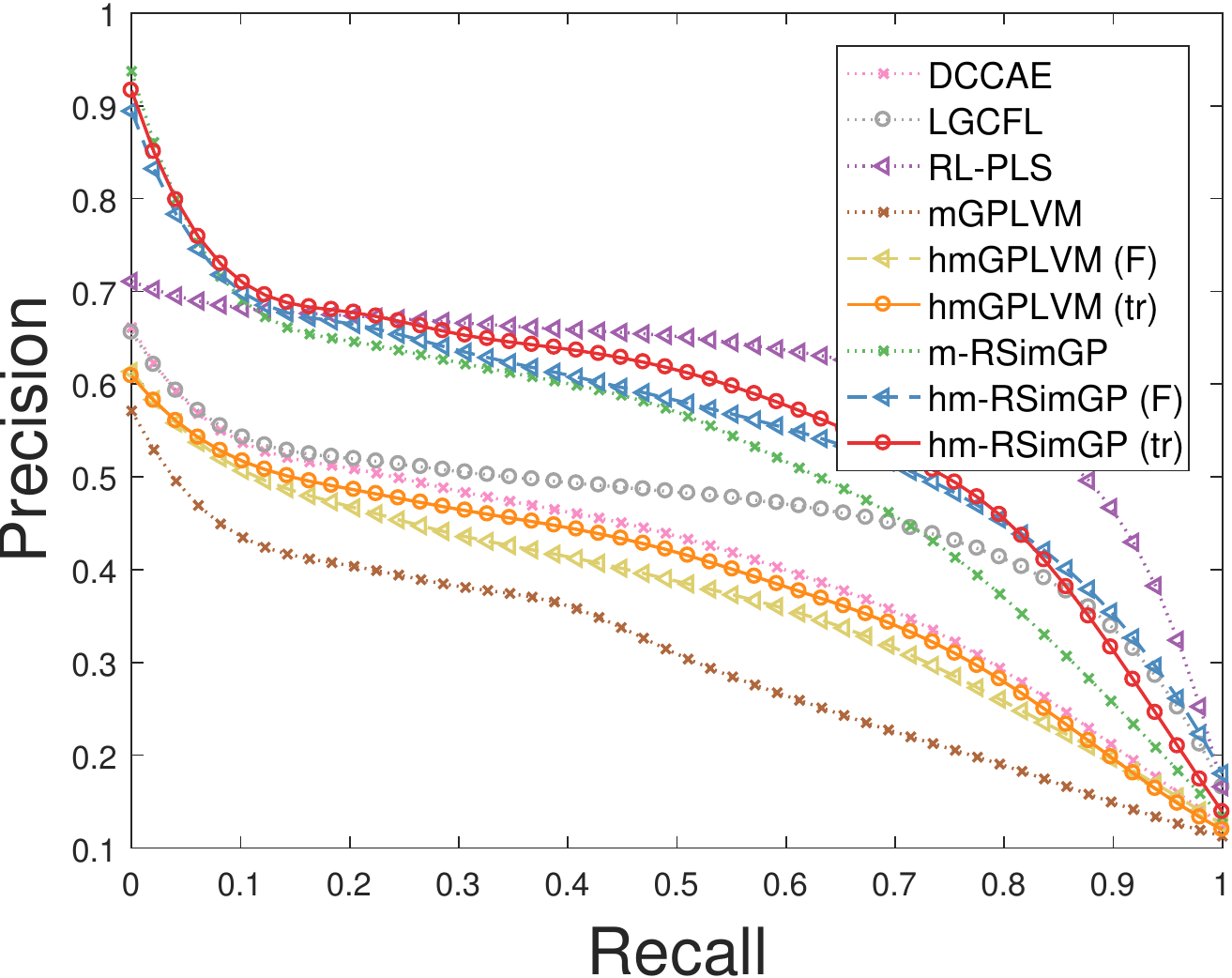}\label{fig:prcurve_im2txt_tvgraz}}
    \hfill
    \subfloat[TVGraz: Text query]{\includegraphics[width=0.22\textwidth]{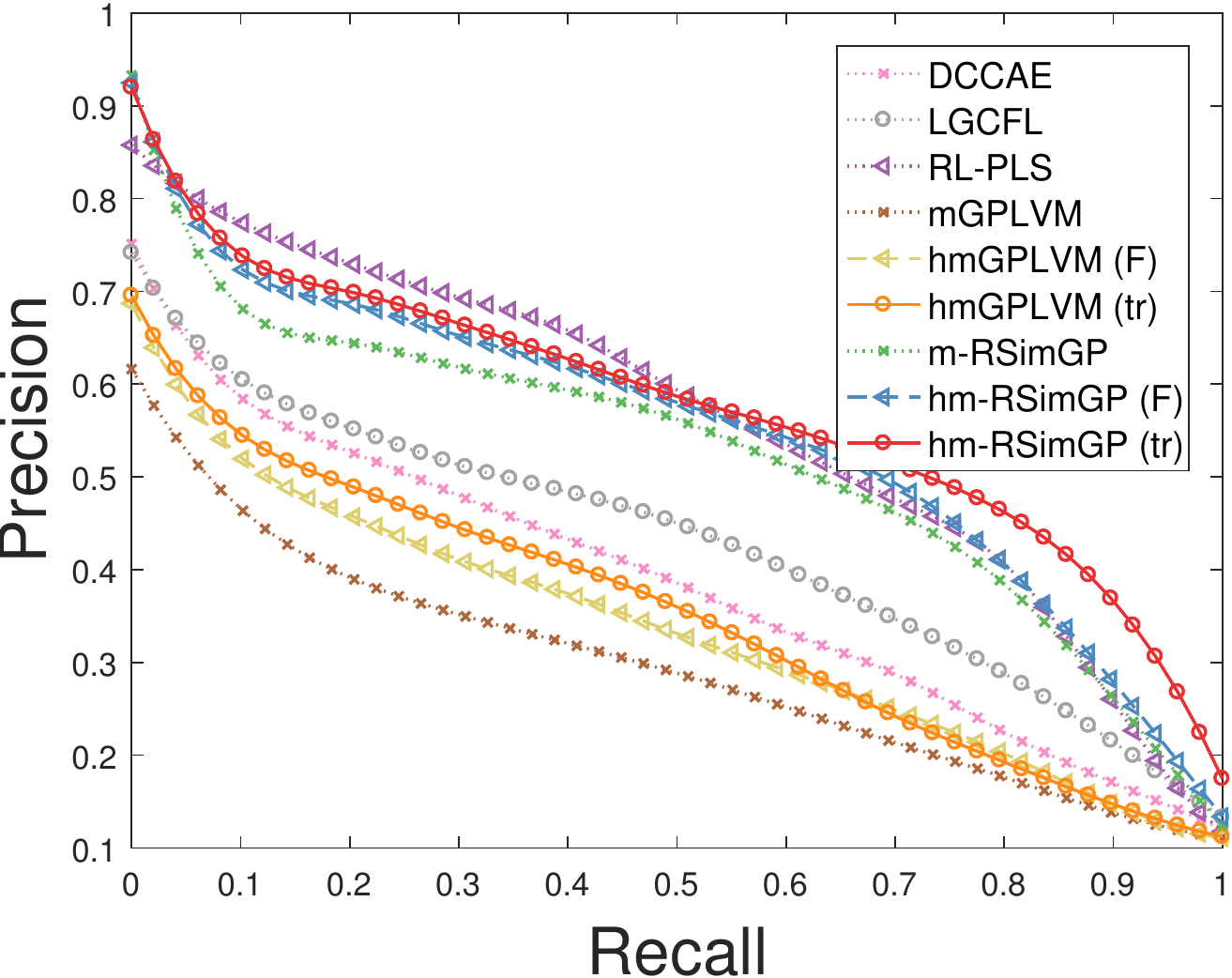}\label{fig:prcurve_txt2im_tvgraz}}
    \hfill
    \subfloat[MSCOCO: Image query]{\includegraphics[width=0.22\textwidth]{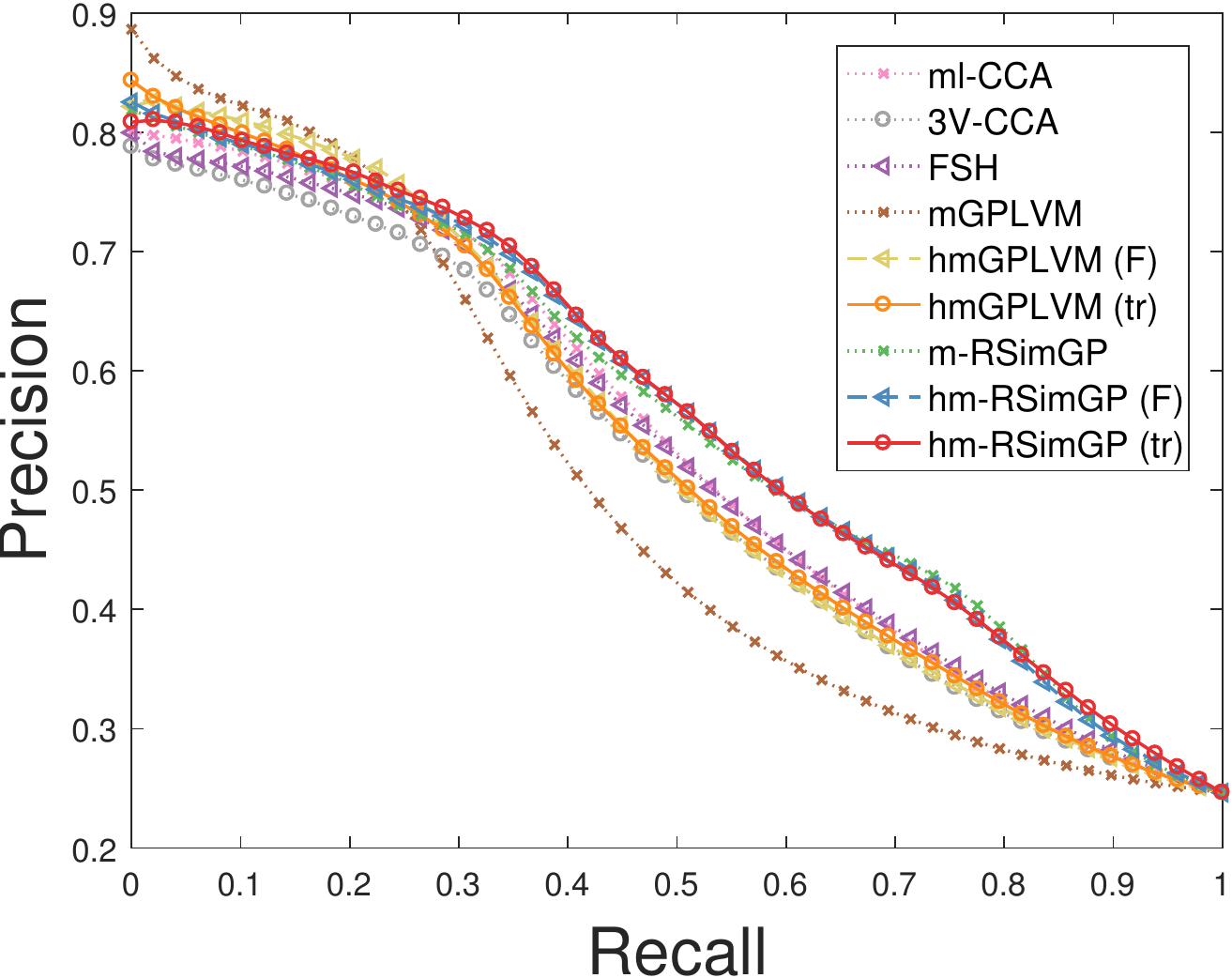}\label{fig:prcurve_im2txt_coco}}
    \hfill
    \subfloat[MSCOCO: Text query]{\includegraphics[width=0.22\textwidth]{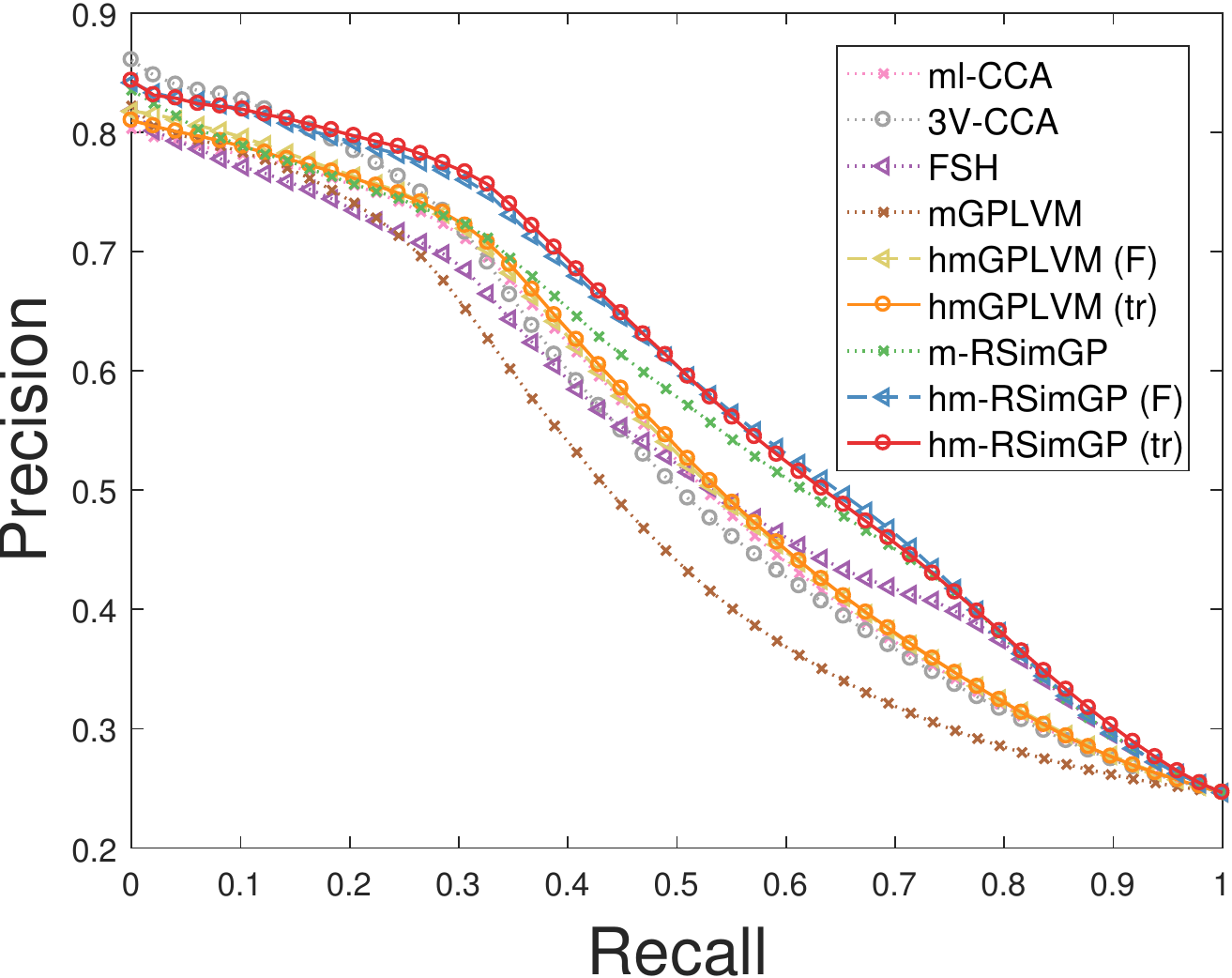}\label{fig:prcurve_txt2im_coco}}
\caption{The performance comparison of different methods for cross-modal retrieval based on precision-recall curve.} \label{fig:prcurves}
\end{figure*}

\begin{figure*}[!b]
\centering
\includegraphics[width=0.9\textwidth]{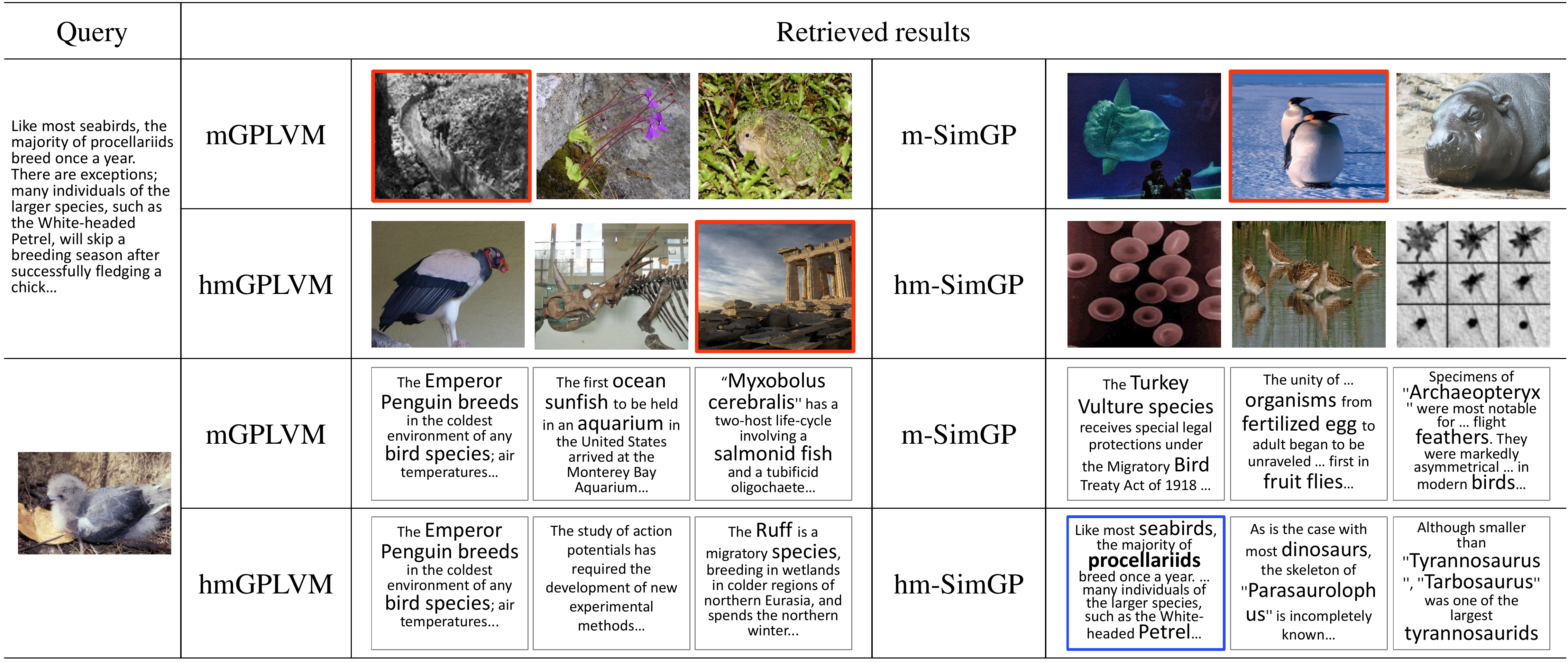}
\caption{Examples of cross-modal retrieval on Wiki dataset for the proposed models with the trace harmonization. The top three retrieved results are presented. We highlight the relevant keywords in the retrieved texts for comparison. Red rectangle indicates a false positive example, and blue rectangle indicates the ground truth instance.}\label{fig:retrievedresults}
\end{figure*}

Table~\ref{tab:pascal_results} shows the mAP scores of our proposed methods and the compared methods on the PASCAL dataset. It clearly shows that the harmonization mechanism is well adapted to multimodal Gaussian process learning framework. Specifically, each of the three harmonization constraints shows a great boost over the previous GPLVM methods, {\it i.e.}, mGPLVM, m-SimGP and m-RSimGP. From Table~\ref{tab:pascal_results}, we can draw the following observations:
\begin{enumerate}
\item All the variants of hmGPLVM with different constraints achieve significant improvements over the baseline mGPLVM for both forms of cross-modal retrieval tasks. In particular, the average retrieval performance achieved by hmGPLVM with the F-norm constraint outperforms mGPLVM by $14.2\%$ higher mAP, and the best performance achieved by hmGPLVM with the trace constraint outperforms mGPLVM by $22.6\%$ higher mAP score.
\item The proposed parameter harmonization prior contributes to the performance improvement of the original m-SimGP. For all the cross-modal retrieval tasks, the three hm-SimGP variants produce higher mAP scores than m-SimGP, and with the trace constraint, hm-SimGP outperforms the other two distance-based harmonization variants. Besides, the resulting methods significantly outperform the state-of-the-art methods, {\it e.g.}, the linear JFSSL, the supervised LGCFL, the discrete CCQ, the probabilistic MLBE, the non-linear DCCAE, and also outperform two non-parametric compared methods, \emph{i.e.}, DS-GPLVM with a discriminative shared-space prior and m-DSimGP with a global structure consistent prior.
\item The proposed three variants of hm-RSimGP outperform the baseline m-RSimGP which is learned by adding a cross-modal semantic prior over the latent space to m-SimGP. And still hm-RSimGP with the trace constraint gains the best mAP scores for both forms of the cross-modal retrieval tasks. Therefore, our proposed harmonization prior can be combined with a different kind of prior over the latent space.
\end{enumerate}

The experimental results on the Wiki dataset and the TVGraz dataset are shown in Table~\ref{tab:wiki_tvgraz_results}. We can see that the proposed models with any of the three harmonization constraints can consistently gain higher mAP scores over the respective GPLVM baselines. For most of the retrieval tasks on Wiki and TVGraz, the trace constraint still makes the harmonization models achieve better performance compared to the other two constraints. On the Wiki dataset, however, hm-SimGP with the $l_{2,1}$-norm constraint achieves higher average mAP scores than its counterparts. On the TVGraz dataset, we find that RL-PLS can achieve the best mAP performance by using real value class labels as an assistant to less stylistic text modality. The harmonization prior makes m-RSimGP with a simple binary class prior achieve a comparable performance.

\begin{figure*}[!t]
\centering
    \subfloat[PASCAL]{\includegraphics[width=0.22\textwidth]{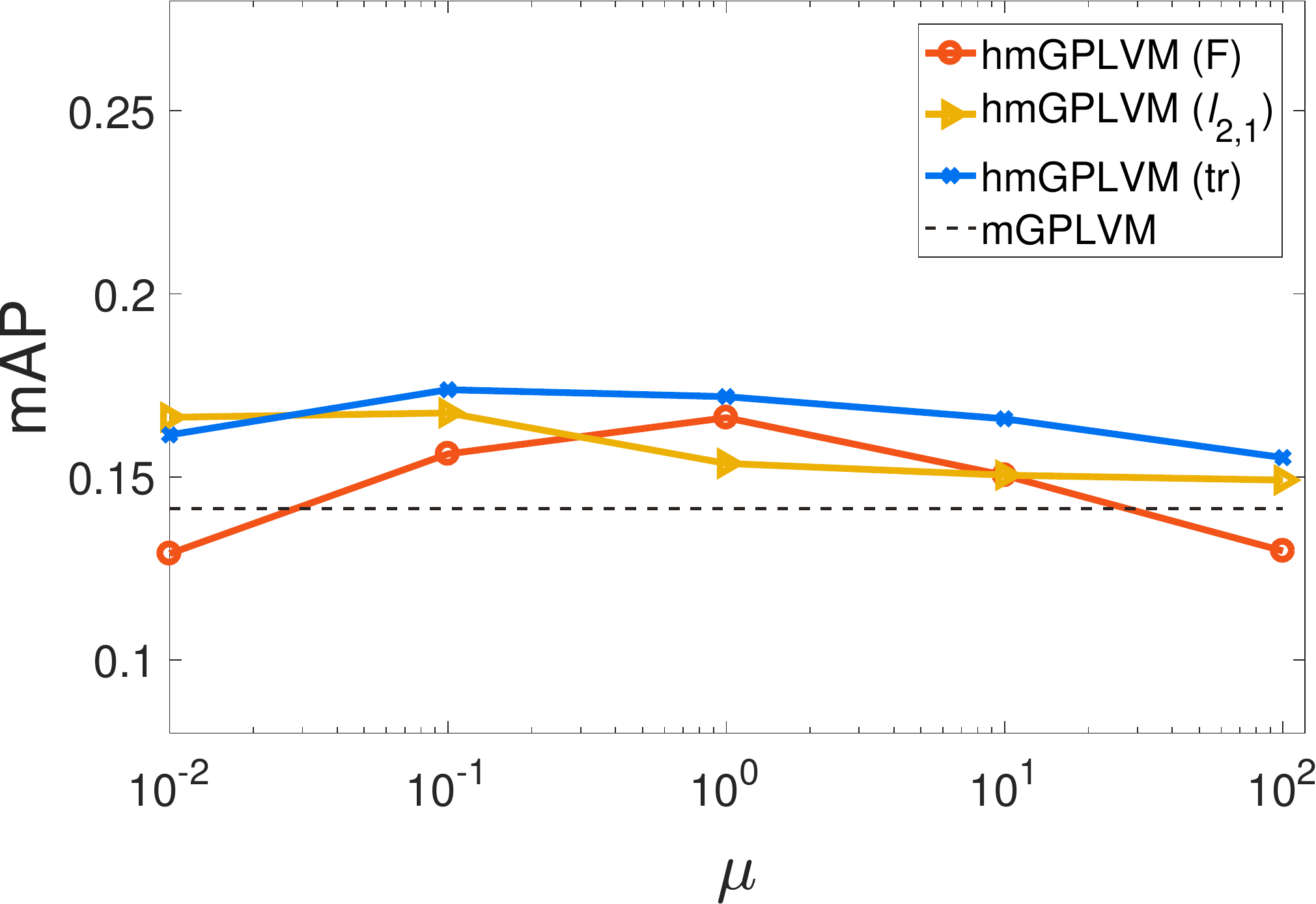}\label{fig:pascal-param-hmgplvm}}
    \hfill
    \subfloat[PASCAL]{\includegraphics[width=0.22\textwidth]{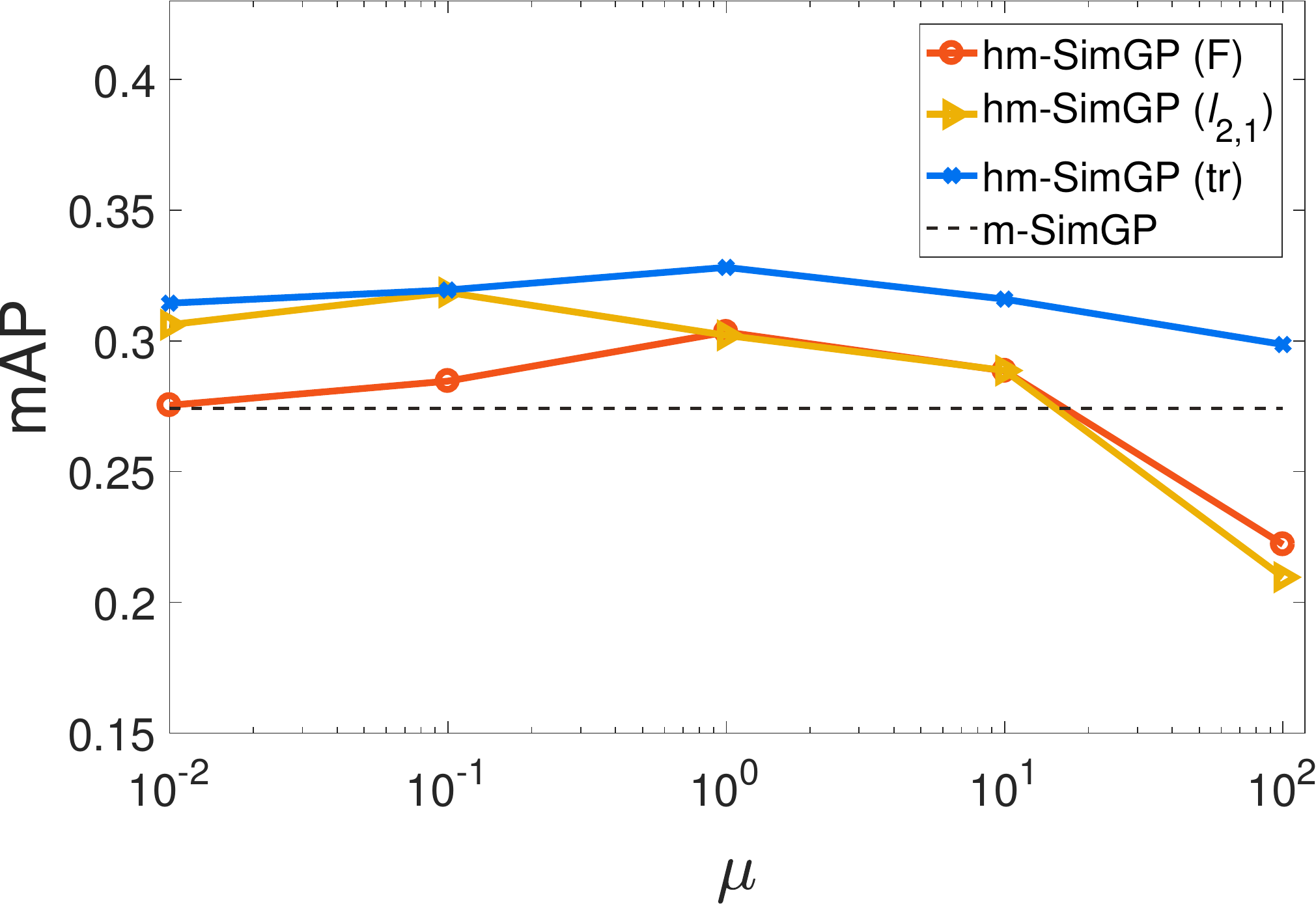}\label{fig:pascal-param-hmsimgp}}
    \hfill
    \subfloat[Wiki]{\includegraphics[width=0.22\textwidth]{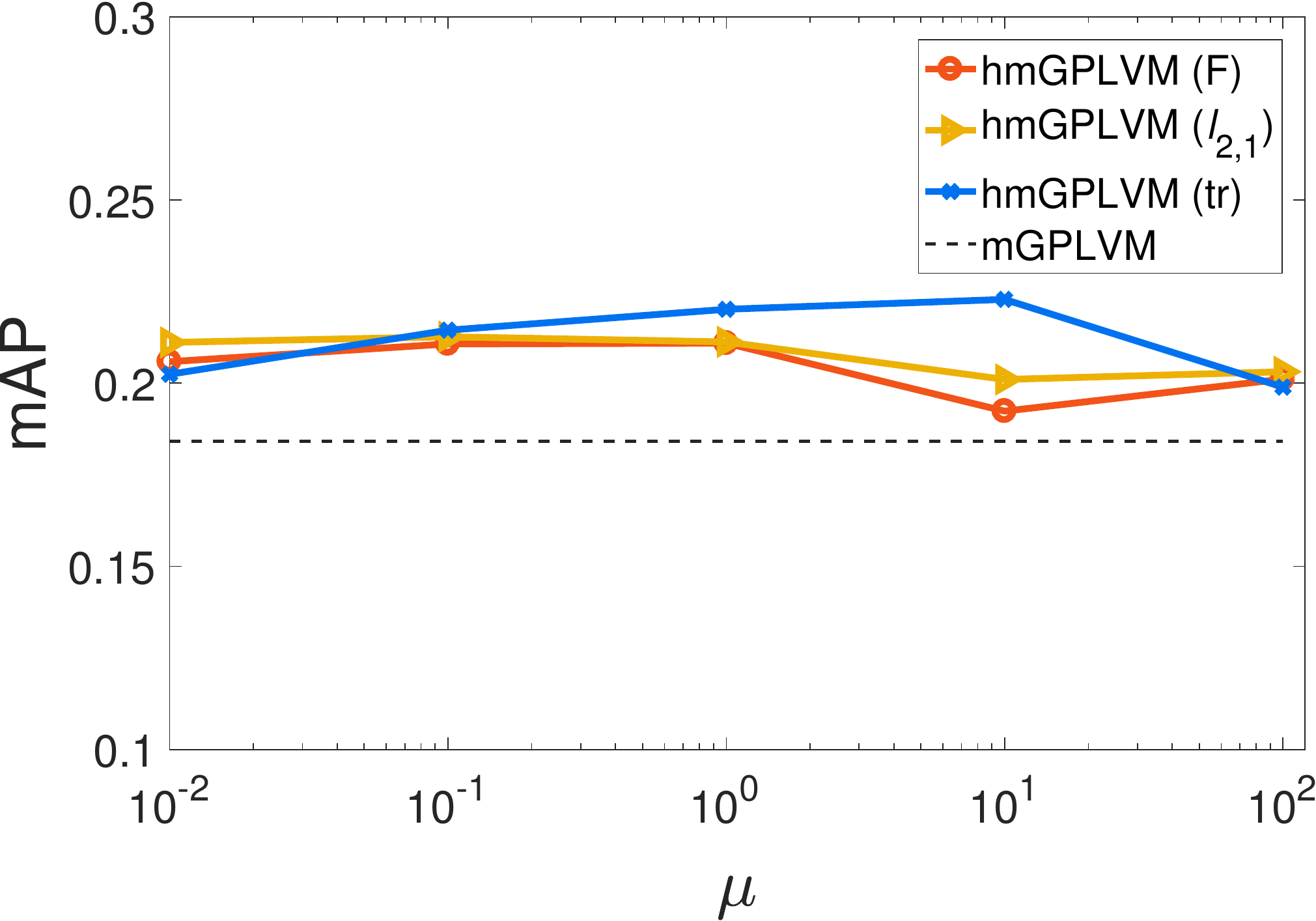}\label{fig:wiki-param-hmgplvm}}
    \hfill
    \subfloat[Wiki]{\includegraphics[width=0.22\textwidth]{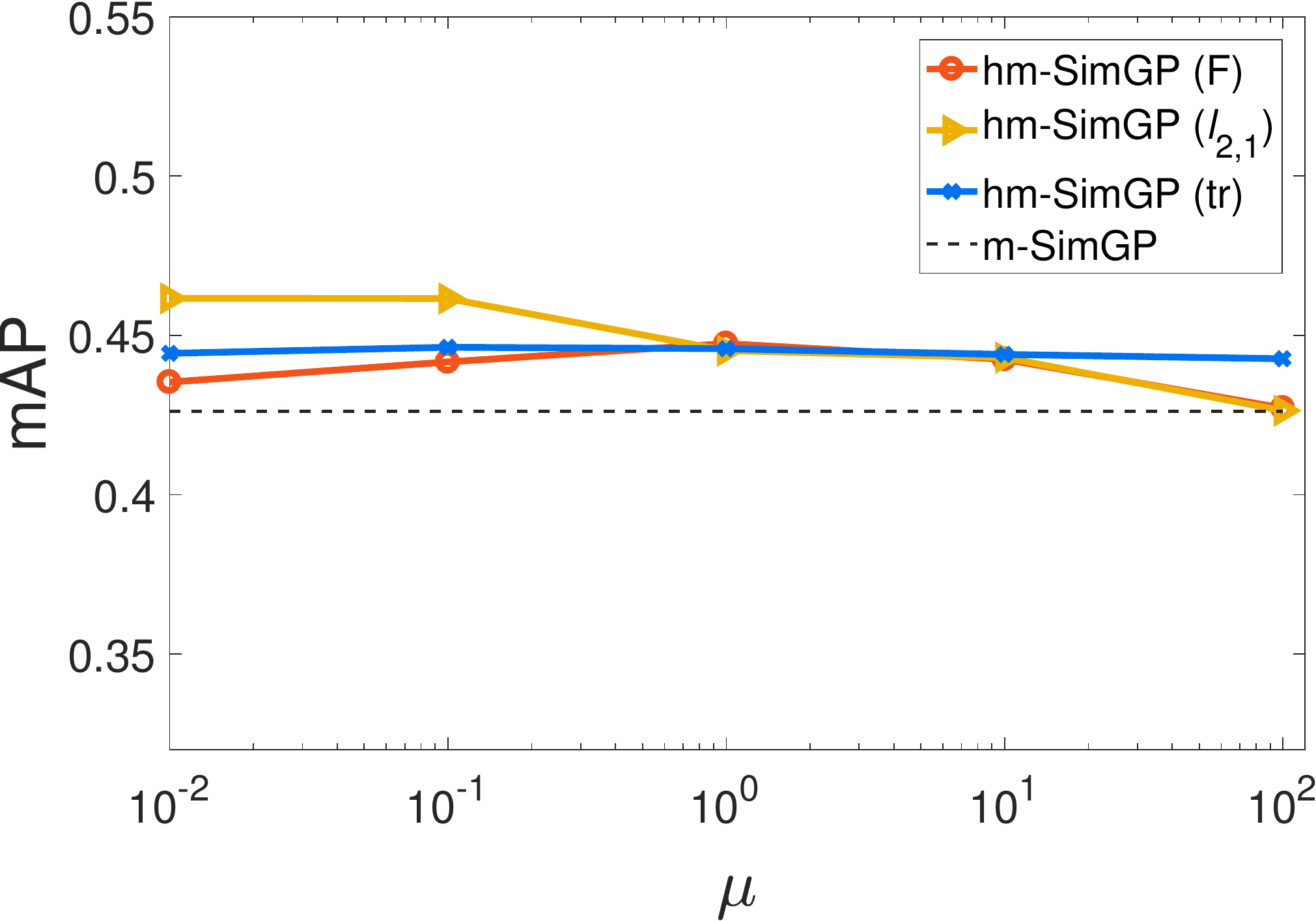}\label{fig:wiki-param-hmsimgp}}
    \\
    \subfloat[TVGraz]{\includegraphics[width=0.22\textwidth]{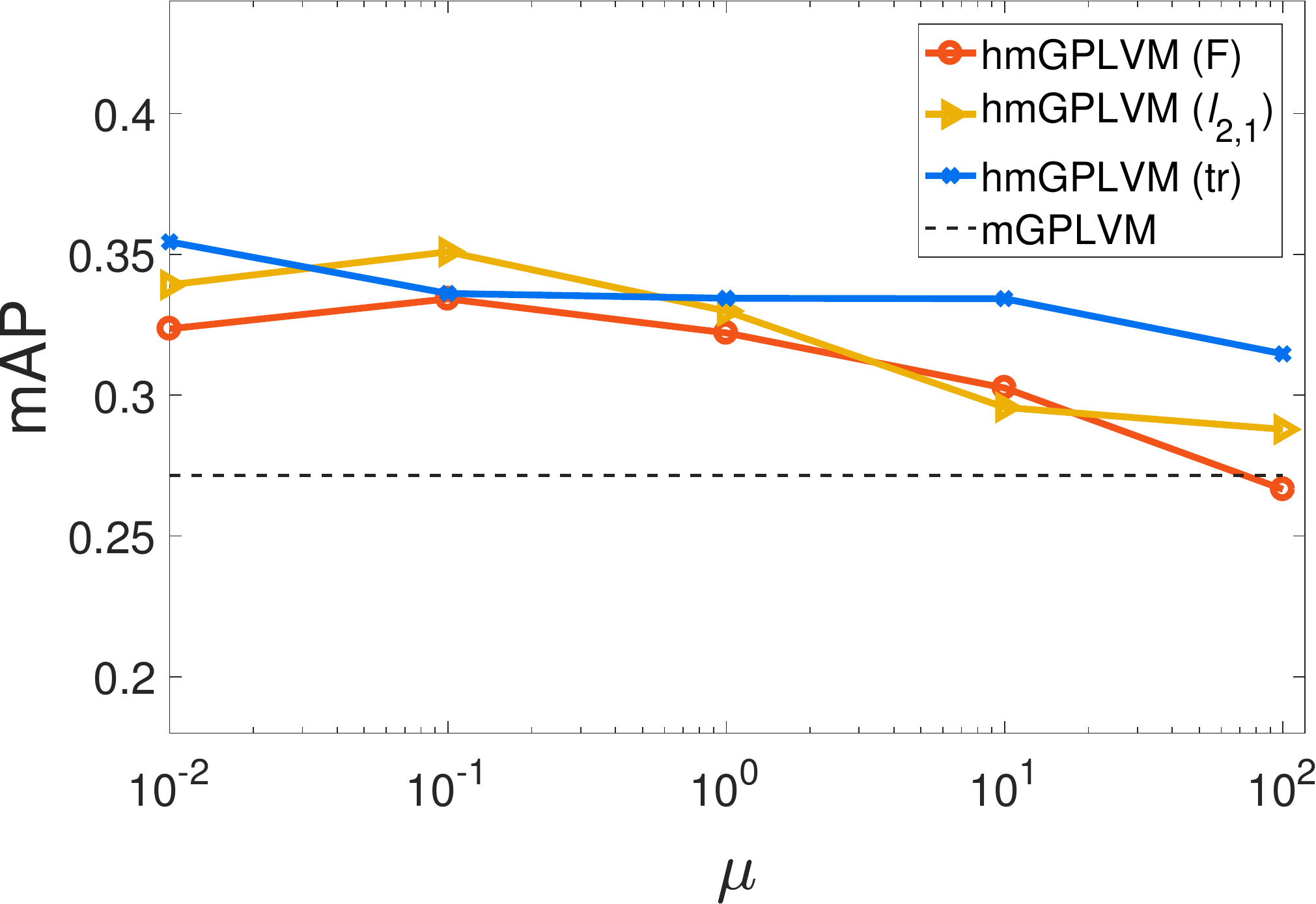}\label{fig:tvgraz-param-hmgplvm}}
    \hfill
    \subfloat[TVGraz]{\includegraphics[width=0.22\textwidth]{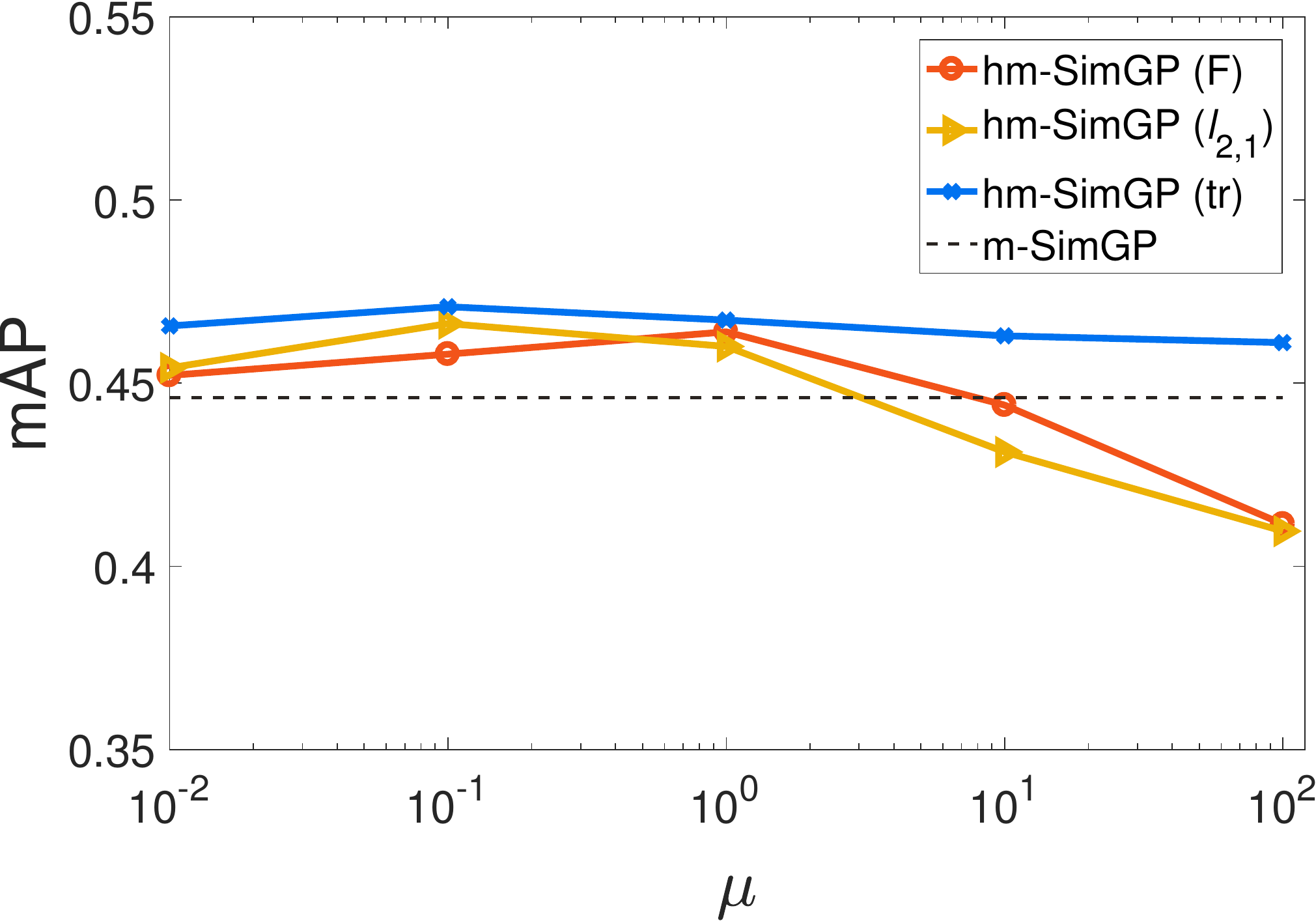}\label{fig:tvgraz-param-hmsimgp}}
    \hfill
    \subfloat[MSCOCO]{\includegraphics[width=0.22\textwidth]{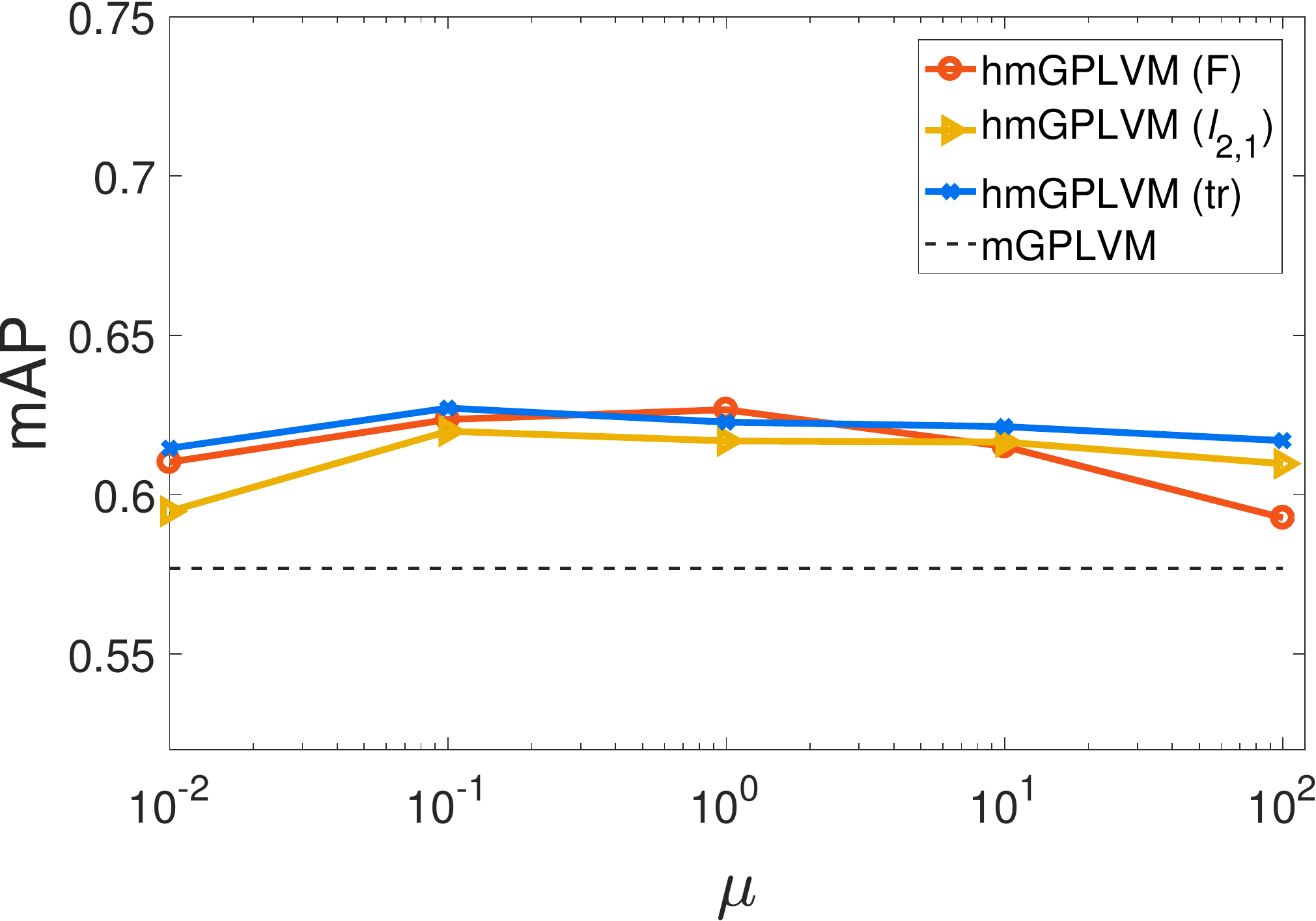}\label{fig:coco-param-hmgplvm}}
    \hfill
    \subfloat[MSCOCO]{\includegraphics[width=0.22\textwidth]{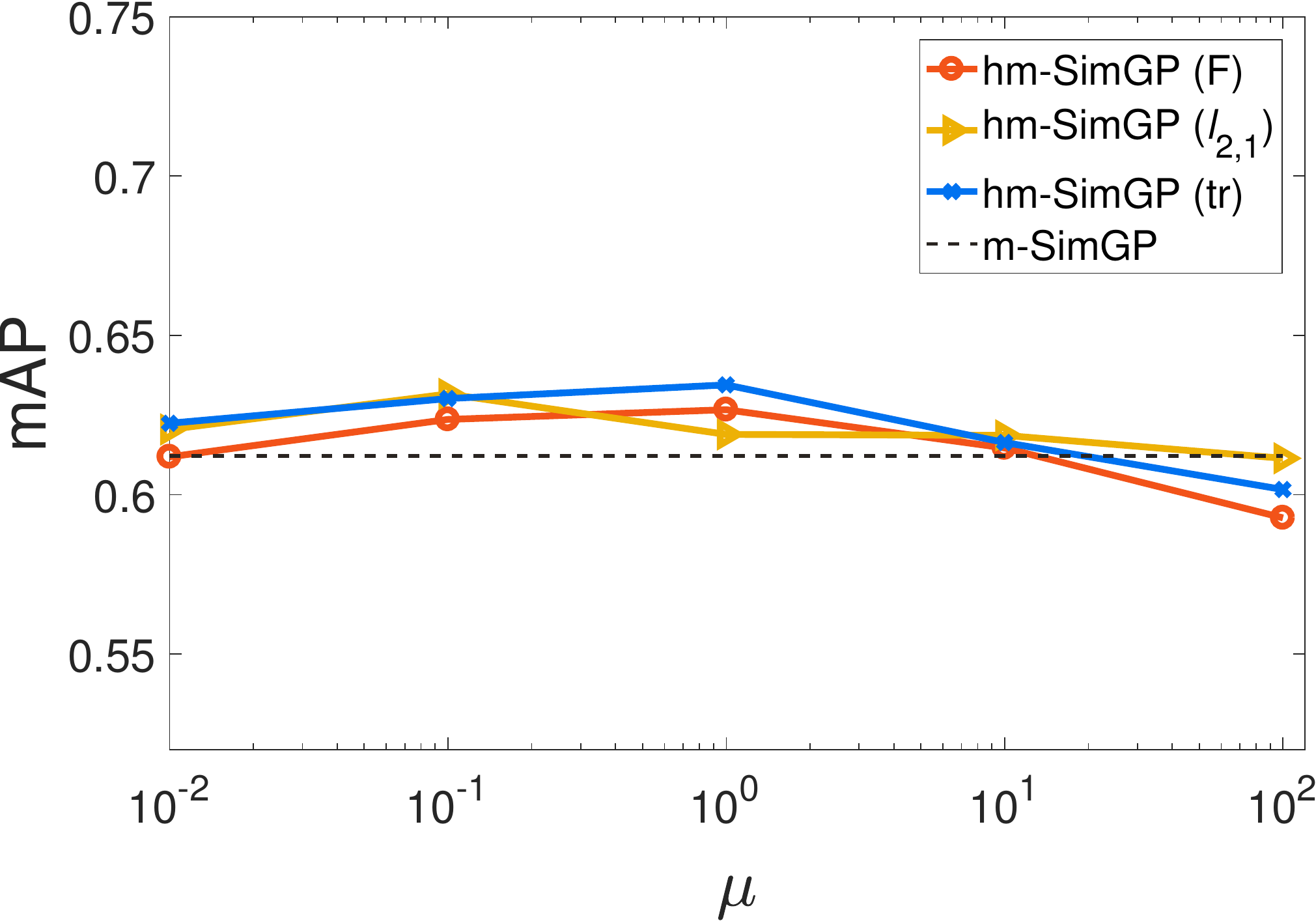}\label{fig:coco-param-hmsimgp}}
\caption{Variants of hmGPLVM and hm-SimGP: average mAP score of two cross-modal retrieval tasks as a function of the harmonization parameters ($\mu = \mu _1 = \mu _2$) for PASCAL, Wiki, TVGraz, and MSCOCO datasets respectively.}\label{fig:param-hmgplvm-hmsimgp}
\end{figure*}

The MSCOCO dataset with multiple labels has recently been used for cross-modal retrieval. We use it to compare our methods with CCA~\cite{neco/HardoonSS04},  FSH~\cite{cvpr/LiuJWHZ17}, 3V-CCA~\cite{gong2014multi}, ml-CCA~\cite{iccv/RanjanRJ15}, and DCCAE~\cite{icml/WangALB15} for both image-to-text (I$\rightarrow$T) and text-to-image (T$\rightarrow$I) retrieval tasks. Table~\ref{tab:coco_results} shows the mAP scores of these methods on MSCOCO. We observe that the experimental results are similar to those on the previous three datasets, and also show that our harmonization methods outperform their counterparts.

Fig.~\ref{fig:prcurves} shows the performance in terms of precision-recall curve for cross-modal retrieval tasks. For clarity, we plot the corresponding precision-recall curves of several representative methods, including our hmGPLVM (F), hmGPLVM (tr), hm-RSimGP (F), and hm-RSimGP (tr).
We observe that the experimental results based on precision-recall curve are consistent with the mAP results. Our models with the proposed harmonization constraints outperform their baselines, and the hm-RSimGP with the trace constraint achieves overall the best results compared to other methods.

Some retrieval results on the Wiki dataset are shown in Fig.~\ref{fig:retrievedresults}. We take the proposed models with the trace harmonization prior as an example. An image/text pair from the ``biology" category is used as the query.
A retrieved result is considered correct if it belongs to the same category as the given query~\cite{mm/RasiwasiaPCDLLV10}.
Through a comprehensive comparison, we can see that our harmonized multimodal GPLVMs achieve better cross-modal retrieval performance than the original models without harmonization prior.

We can conclude from the experiments on the four datasets that the proposed harmonization mechanism has strong generalization ability in modeling a joint prior over the latent model parameters for different multimodal GPLVMs. And varied harmonization functions can be capable of playing an active role in capturing the correlation structure of multimodal data. Taken together, for multimodal GPLVM learning, the ratio-based harmonization function, {\it i.e.}, the trace function in Eq.~\eqref{eq:3-R}, performs slightly better than the distance-based harmonization functions, {\it i.e.}, the F-norm function in Eq.~\eqref{eq:3-F} and the $l_{2,1}$-norm function in Eq.~\eqref{eq:3-L21}.

\subsection{Parameter sensitivity analysis }
\label{ssec:Parameter}
In the proposed models, the harmonization parameters ${\mu_1}$ and ${\mu_2}$ can control the consistency between the GP kernels and the similarity matrix in the latent space, and they are assigned with the same value, {\it i.e.}, $\mu _1 = \mu _2 = \mu$, indicating equal importance of different modalities. In this experiment, we perform sensitivity analysis to understand how these parameters influence the cross-modal retrieval performance for different harmonized GP models, including F-norm, $l_{2,1}$-norm, and trace variants. Fig.~\ref{fig:param-hmgplvm-hmsimgp} shows the curves of average mAP scores of two cross-modal retrieval tasks with different settings on the parameter ${\mu}$.

From the subfigures in Fig.~\ref{fig:param-hmgplvm-hmsimgp}, we see that generally the average mAP performance is improved first as the value of ${\mu}$ increases, but degrades quickly as ${\mu}$ further is increased above a certain point. Specifically, the F-norm harmonized models achieve the best mAP score around $1$ for most of cases. For the $l_{2,1}$-norm based models, the mAP performance decreases as ${\mu}$ increases to $10^{-1}$ on all four datasets. The models with the trace constraint perform better when the value of ${\mu}$ is limited to $[10^{-1}, 10^1]$. Fig.~\ref{fig:param-hmgplvm-hmsimgp} also shows that a large $\mu$ tends to make the harmonization models perform worse than the baseline models (as shown in the dashed lines).
We attribute this phenomenon to the fact that a very large value of ${\mu}$ will increase the risk of co-adaptation and cause the model to get stuck in a local optima.

\begin{figure}[!t]
\centering
    \subfloat[hm-RSimGP (F-norm)]{\includegraphics[width=0.2\textwidth]{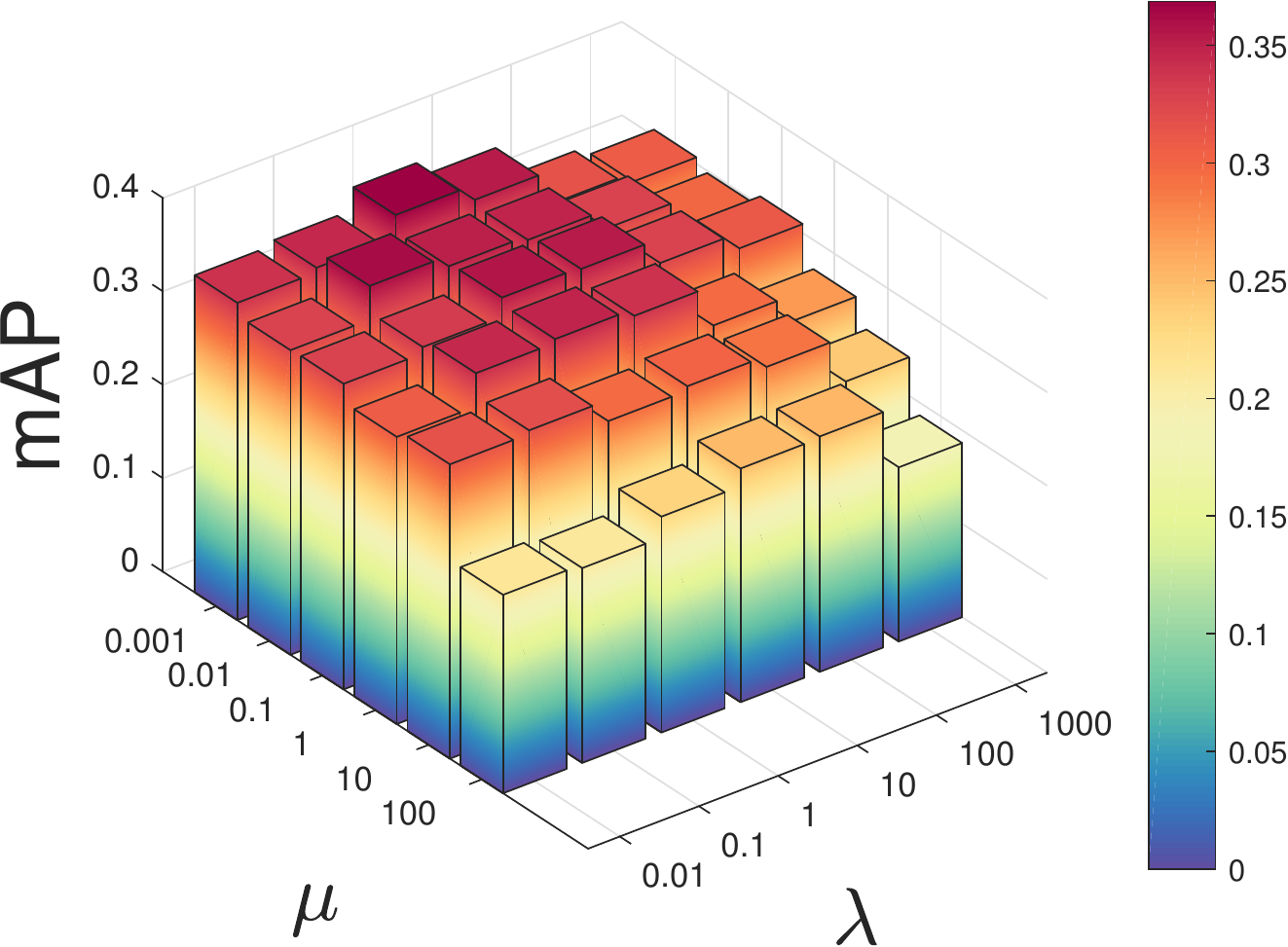}\label{fig:drsimparam_F}}
    \hspace{0.02\textwidth}
    \subfloat[hm-RSimGP ($l_{2,1}$-norm)]{\includegraphics[width=0.2\textwidth]{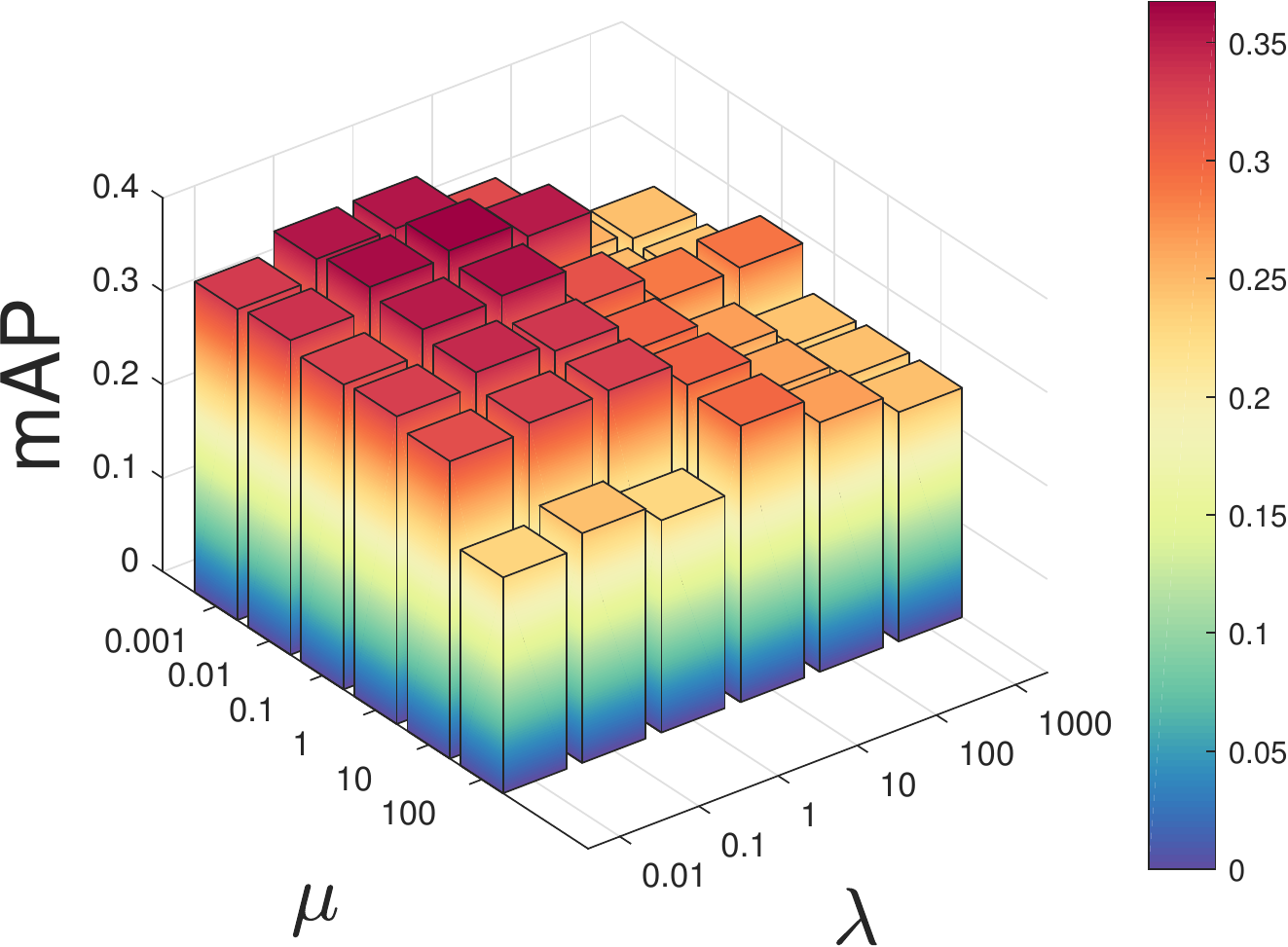}\label{fig:drsimparam_l21}} \\
    \subfloat[hm-RSimGP (trace)]{\includegraphics[width=0.2\textwidth]{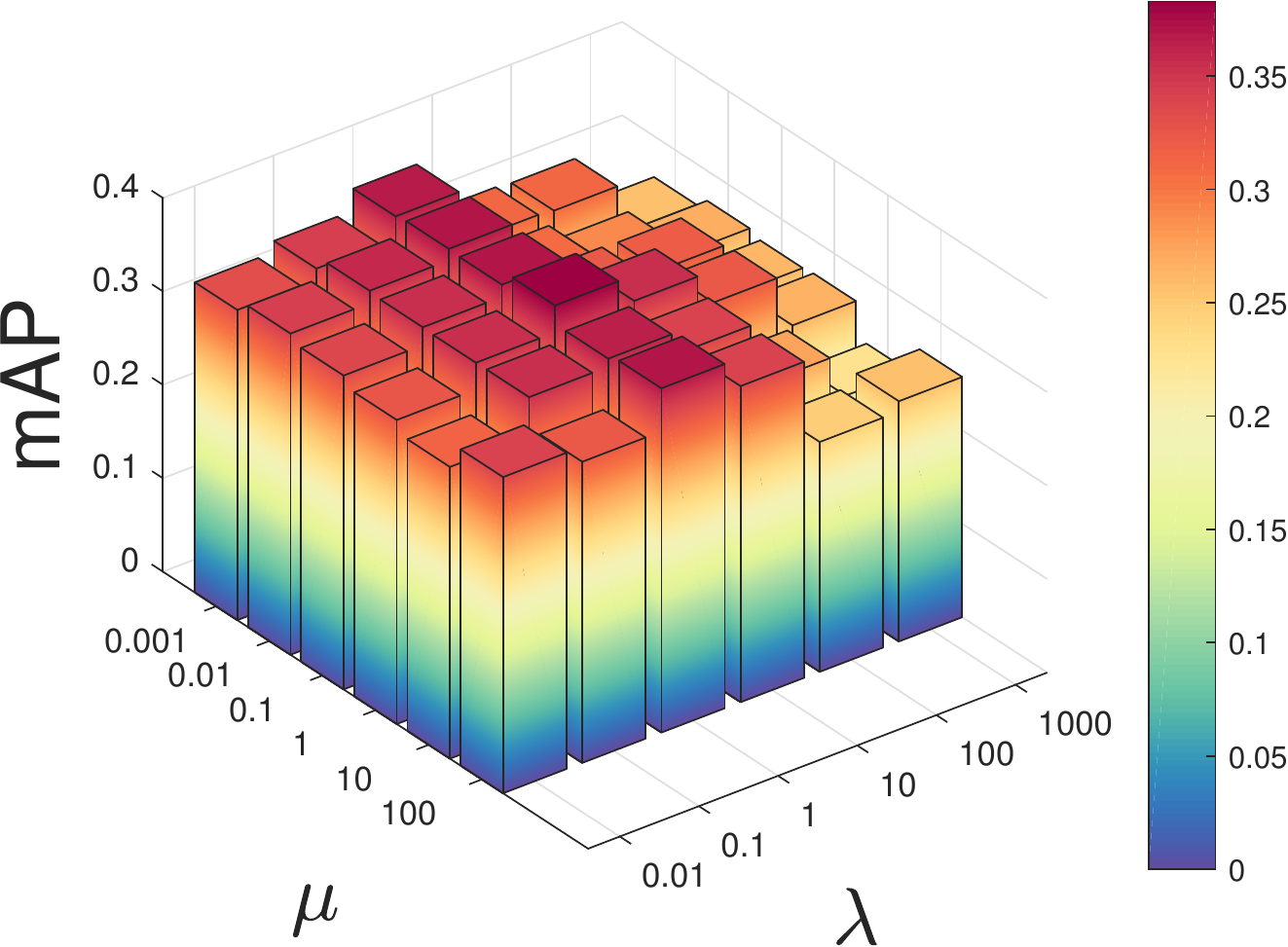}\label{fig:drsimparam_tr}}
\caption{Variants of hm-RSimGP: average mAP score of two cross-modal retrieval tasks as a function of parameters $\mu$ and $\lambda$ for the PASCAL dataset, where $\mu = \mu _1 = \mu _2$, $\lambda = \lambda _1 = \lambda _2$.} \label{fig:parameterhmrsim}
\end{figure}

For hm-RSimGP in Eq.~\eqref{eq:3-23} with different harmonization constraints, we also conduct sensitivity analysis on the trade-off parameters, including ${\mu _1}$ and ${\mu _2}$ for the harmonization terms and ${\lambda_1}$ and ${\lambda_2}$ for the cross-modal semantic terms. As above, ${\mu _1}$ and ${\mu _2}$ are assigned with the same value, {\it i.e.}, $\mu _1 = \mu _2 = \mu$. Further, $\lambda_1$ and $\lambda_2$ are also assigned with the same value, {\it i.e.}, $\lambda _1 = \lambda _2 = \lambda$, indicating that both similar and dissimilar semantic relation are used in the hm-RSimGP model. In this experiment, we perform cross-modal retrieval on the PASCAL dataset, and the average mAP results with different settings on the parameters $\mu$ and $\lambda$ are shown in Fig.~\ref{fig:parameterhmrsim}. We can see that our hm-RSimGP with any one of the three harmonization constraints can achieve relatively better performance as long as the value of $\mu$ or $\lambda$ is not too large. Taking the hm-RSimGP model with the F-norm constraint as an example, the mAP performance is much better when the values of $\mu$ and $\lambda$ are limited to $[0.001, 0.1]$ and $[0.1, 10]$ respectively.
In particular, for all the given $\lambda$s, the performance is significantly reduced when $\mu$ is set to a larger value than 10. It again shows that a very large value of the tradeoff parameter $\mu$ will lead to the degradation on the cross-modal correlation learning ability of the harmonization models. On the whole, a harmonization prior with an appropriate parameter setting can improve the cross-modal retrieval performance of the m-RSimGP model with the semantic prior on the latent points.

\subsection{Analysis on the latent space}
\label{ssec:latentspace}
\begin{figure}[!t]
\centering
    \subfloat[mGPLVM]{\includegraphics[width=0.15\textwidth]{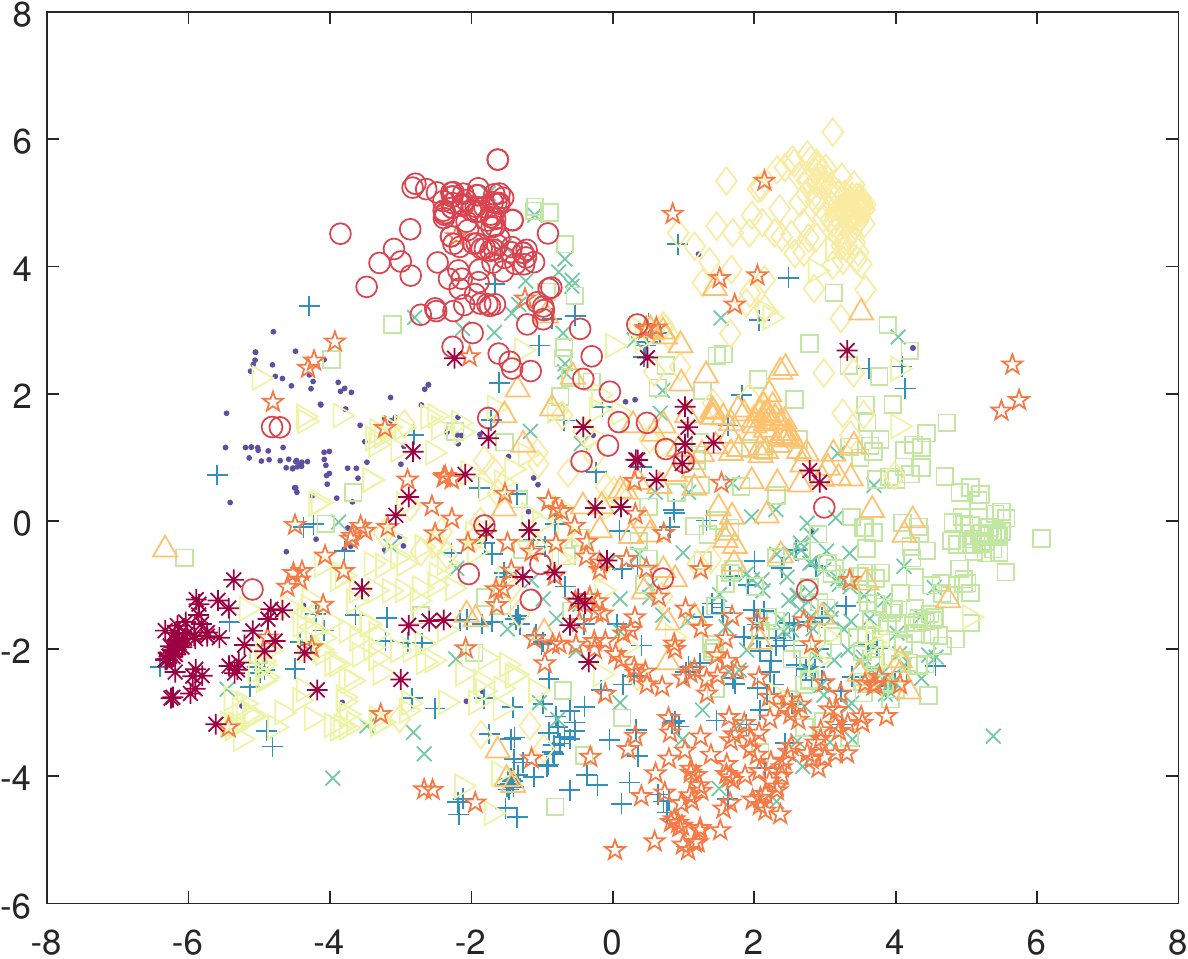}\label{fig:vismGPLVM}}
    \hfill
    \subfloat[m-SimGP]{\includegraphics[width=0.15\textwidth]{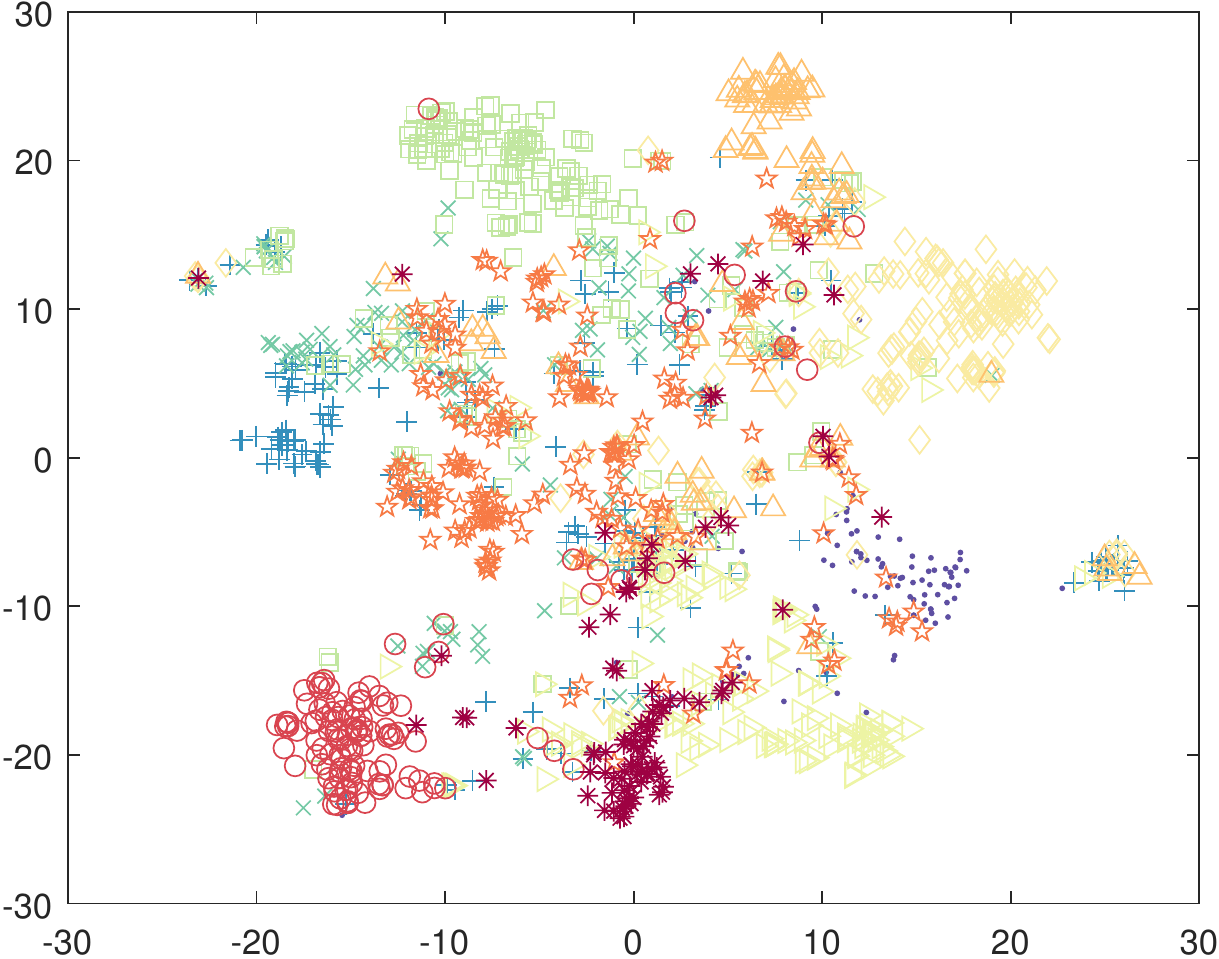}\label{fig:vissimgp}}
    \hfill
    \subfloat[m-RSimGP]{\includegraphics[width=0.15\textwidth]{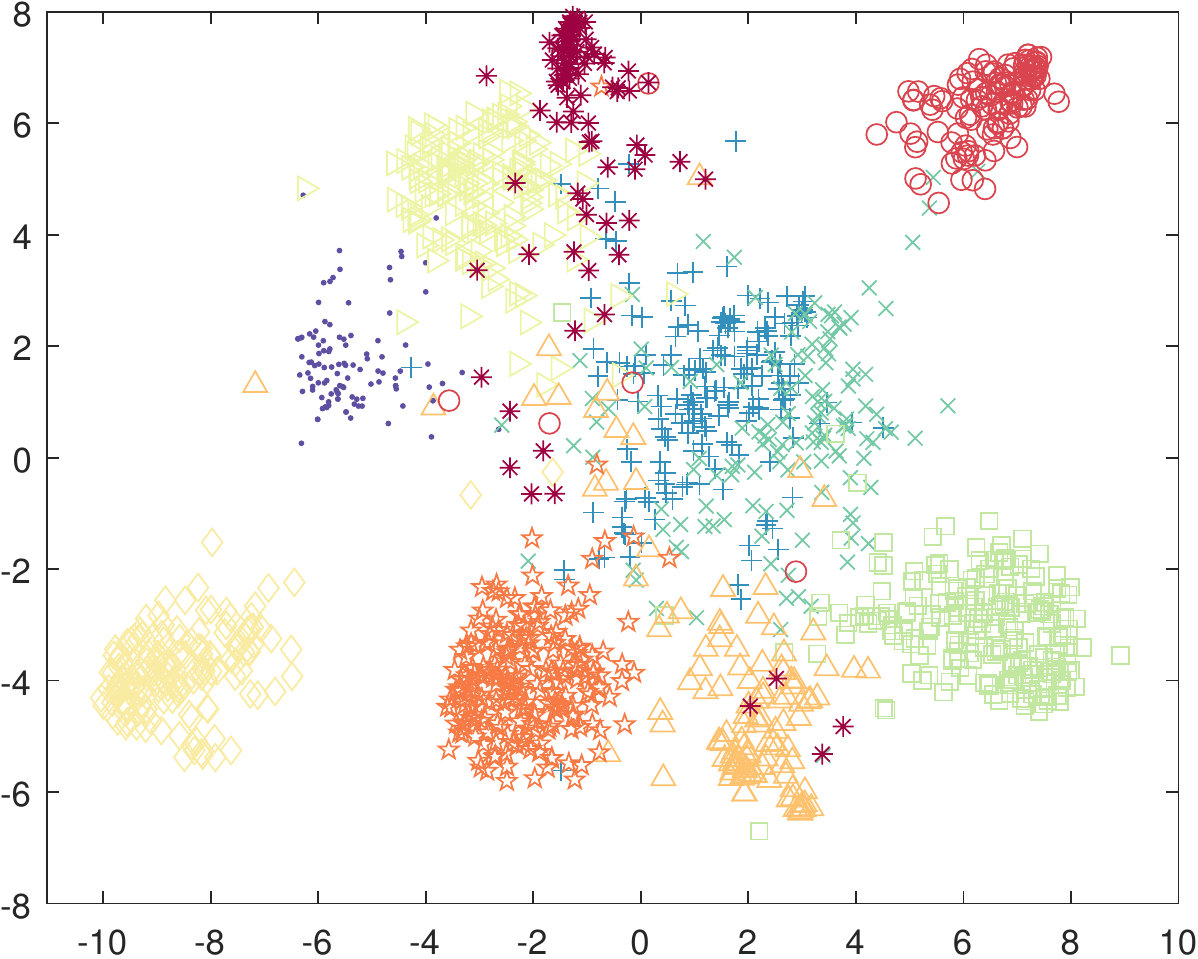}\label{fig:visrsimgp}}
    \\
    \subfloat[hmGPLVM]{\includegraphics[width=0.15\textwidth]{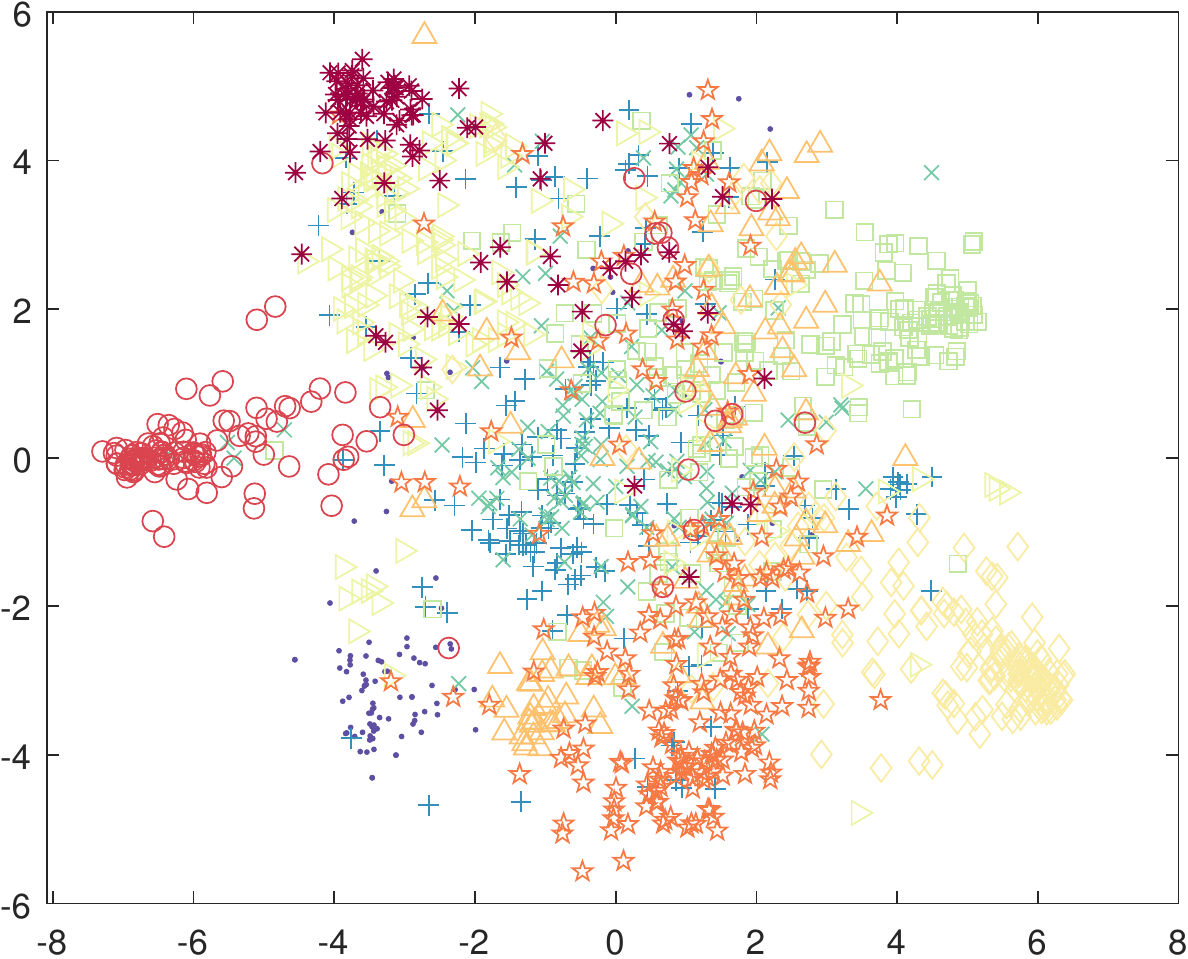}\label{fig:vishmgplvm}}
    \hfill
    \subfloat[hm-SimGP]{\includegraphics[width=0.15\textwidth]{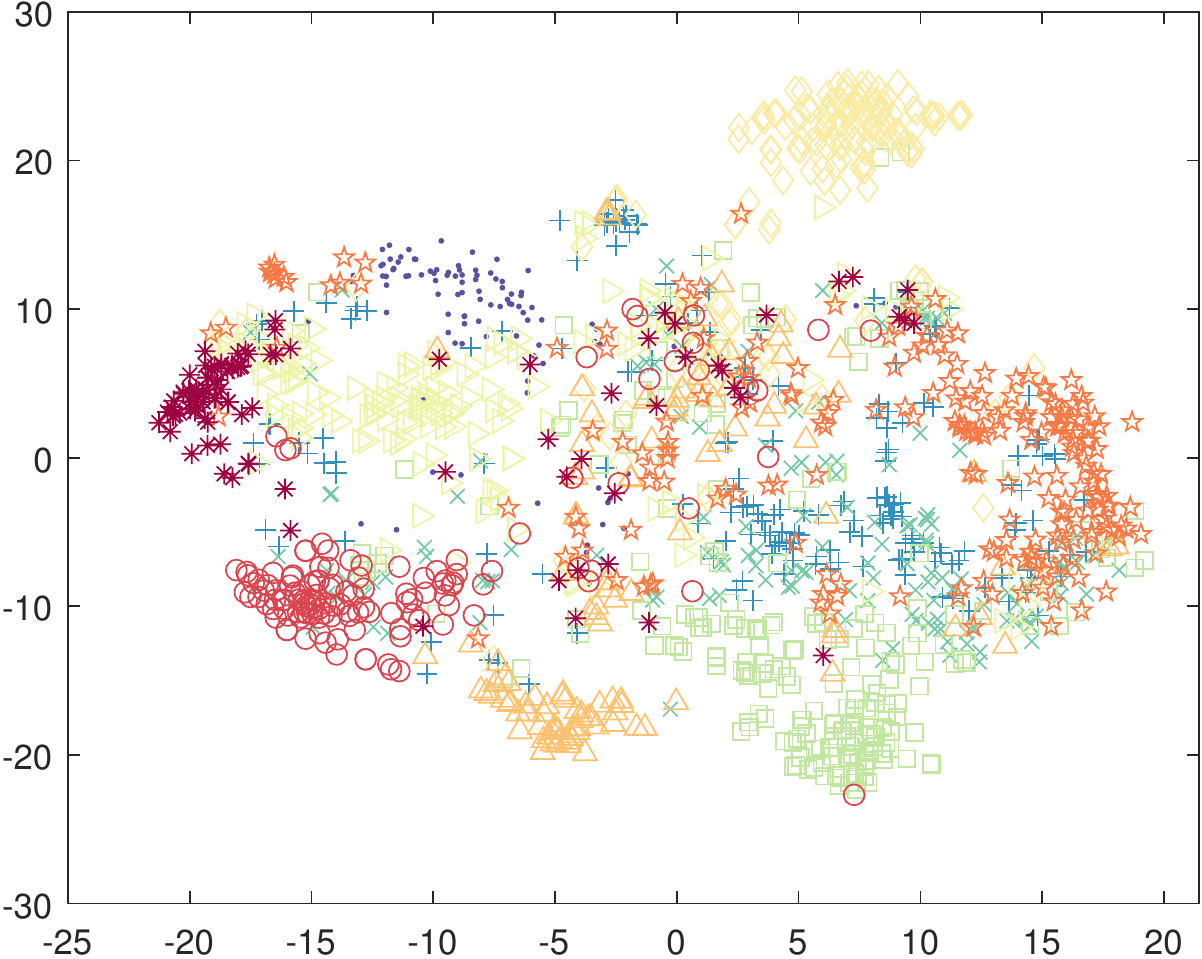}\label{fig:vishmsimgp}}
    \hfill
    \subfloat[hm-RSimGP]{\includegraphics[width=0.15\textwidth]{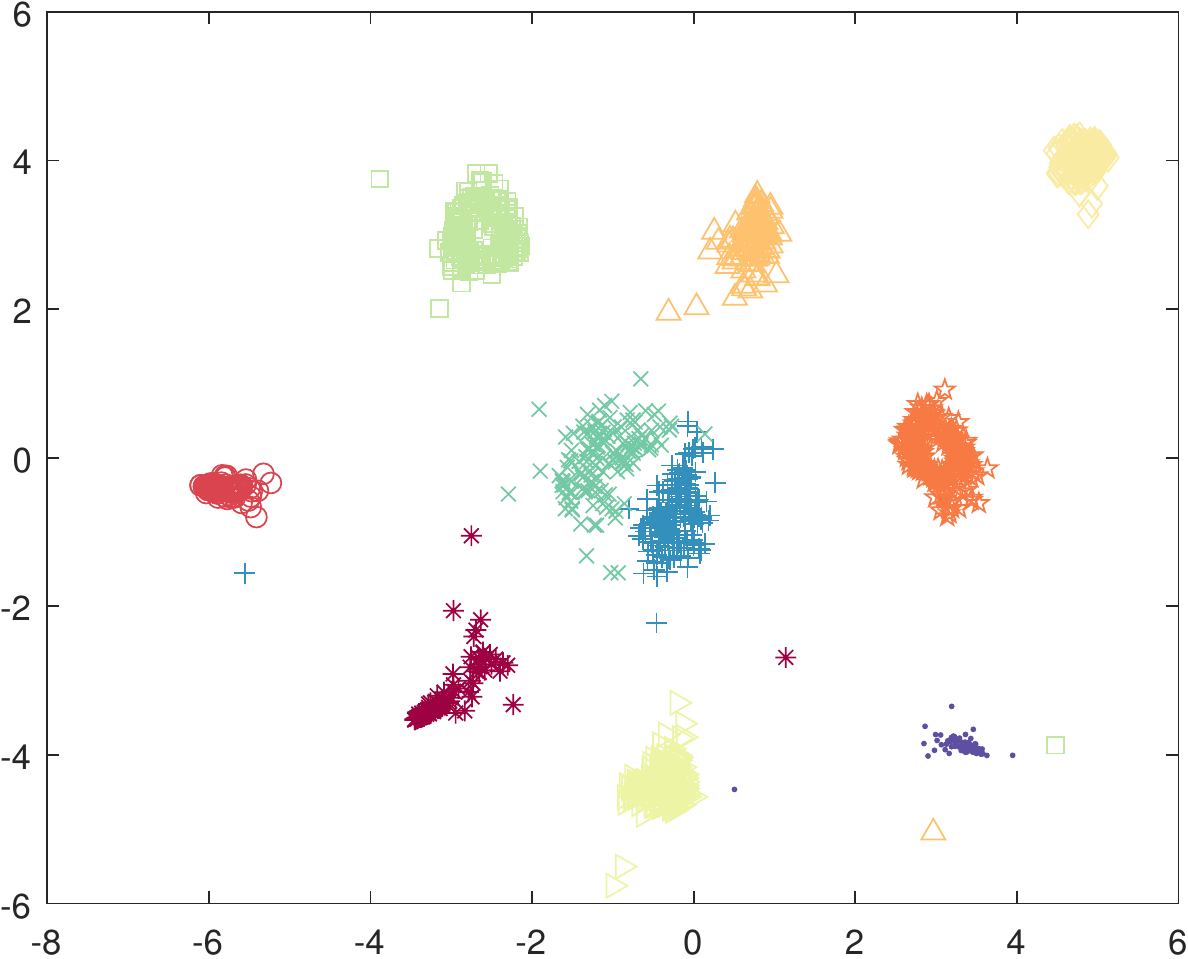}\label{fig:vishmrsimgp}}
\caption{Visualization of the discovered latent representations for the proposed models with the trace harmonization prior on the TVGraz dataset (Better viewed in color PDF). }\label{fig:visualization}
\end{figure}

Here we take the proposed models with the trace harmonization prior for example, and evaluate the discovered latent space of multimodal data.

\subsubsection{Visualization of the latent space}\label{ssec:visual}
We perform a visualization experiment on the TVGraz dataset with 10 categories, where the t-SNE algorithm~\cite{van2008visualizing} is used to project the 10-dimensional latent representations into a 2-dimensional space. Fig.~\ref{fig:visualization} shows that our harmonized approaches perform much better in producing a low dimensional representation compared to the GP baselines. Specifically, the latent representations discovered by the standard mGPLVM provide little information on the category structure of the data  objects, whereas the latent representations discovered by the proposed hmGPLVM (tr) exhibit a clearer grouping structure for the data from the same category.
Therefore, our harmonized GPLVMs can discover more discriminative latent representations for multimodal data.

\begin{figure}[!t]
\centering
     \subfloat[PASCAL]{\includegraphics[width=0.2\textwidth,height=2.3cm]{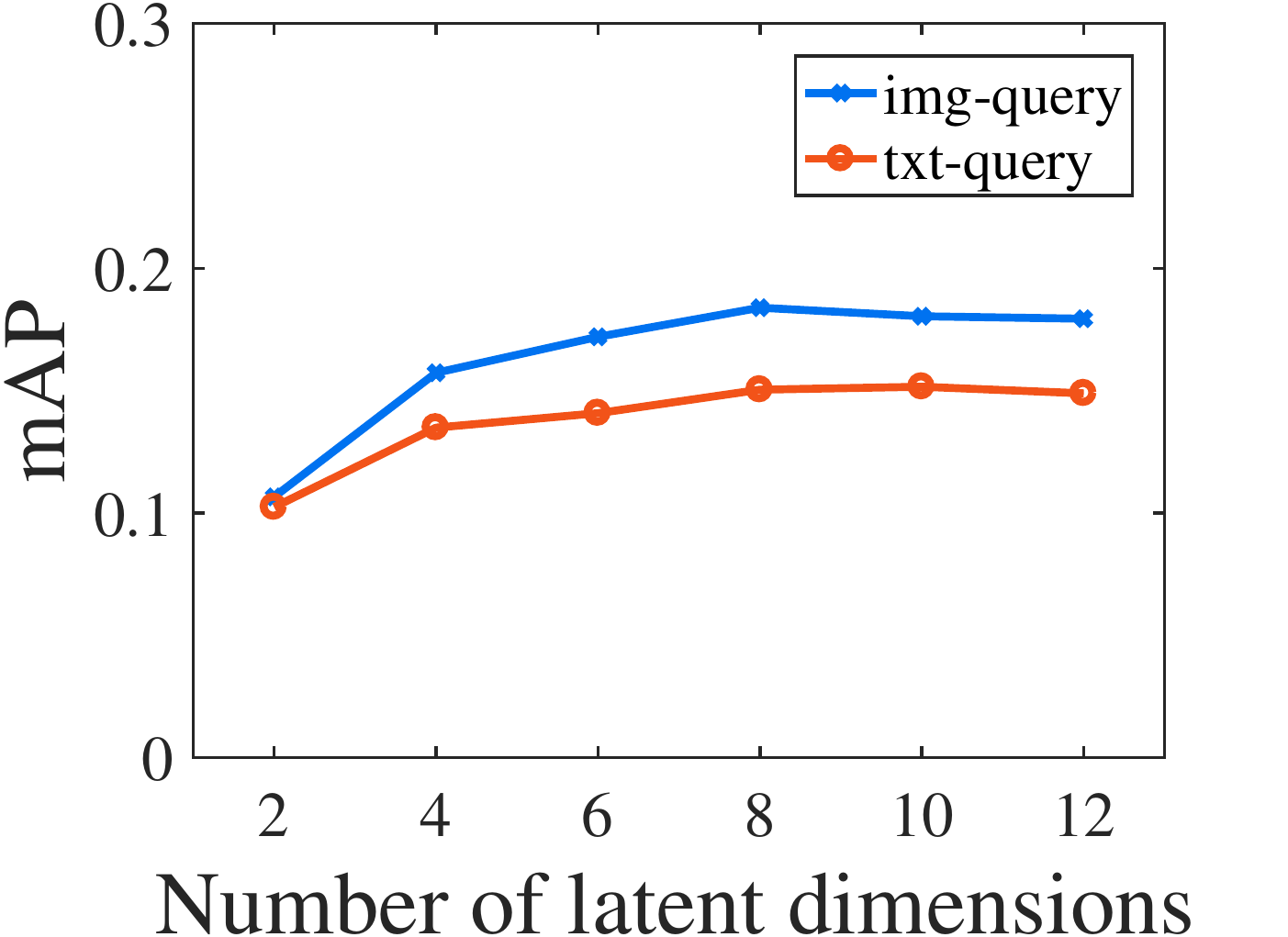}\label{fig:latentpascal}}
     \hspace{0.02\textwidth}
     \subfloat[Wiki]{\includegraphics[width=0.2\textwidth,height=2.3cm]{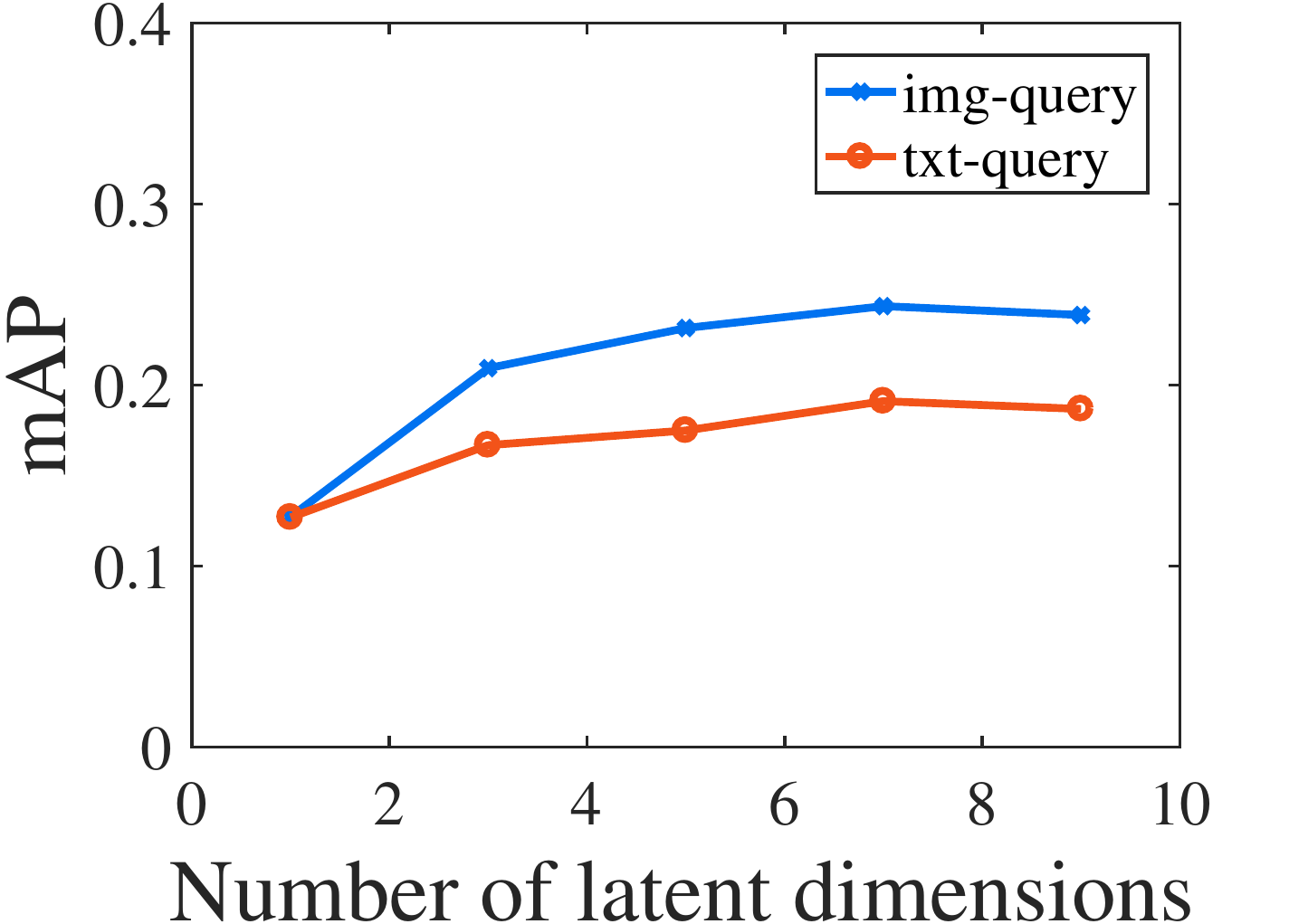}\label{fig:latentwiki}}
     \\
     \subfloat[TVGraz]{\includegraphics[width=0.2\textwidth,height=2.3cm]{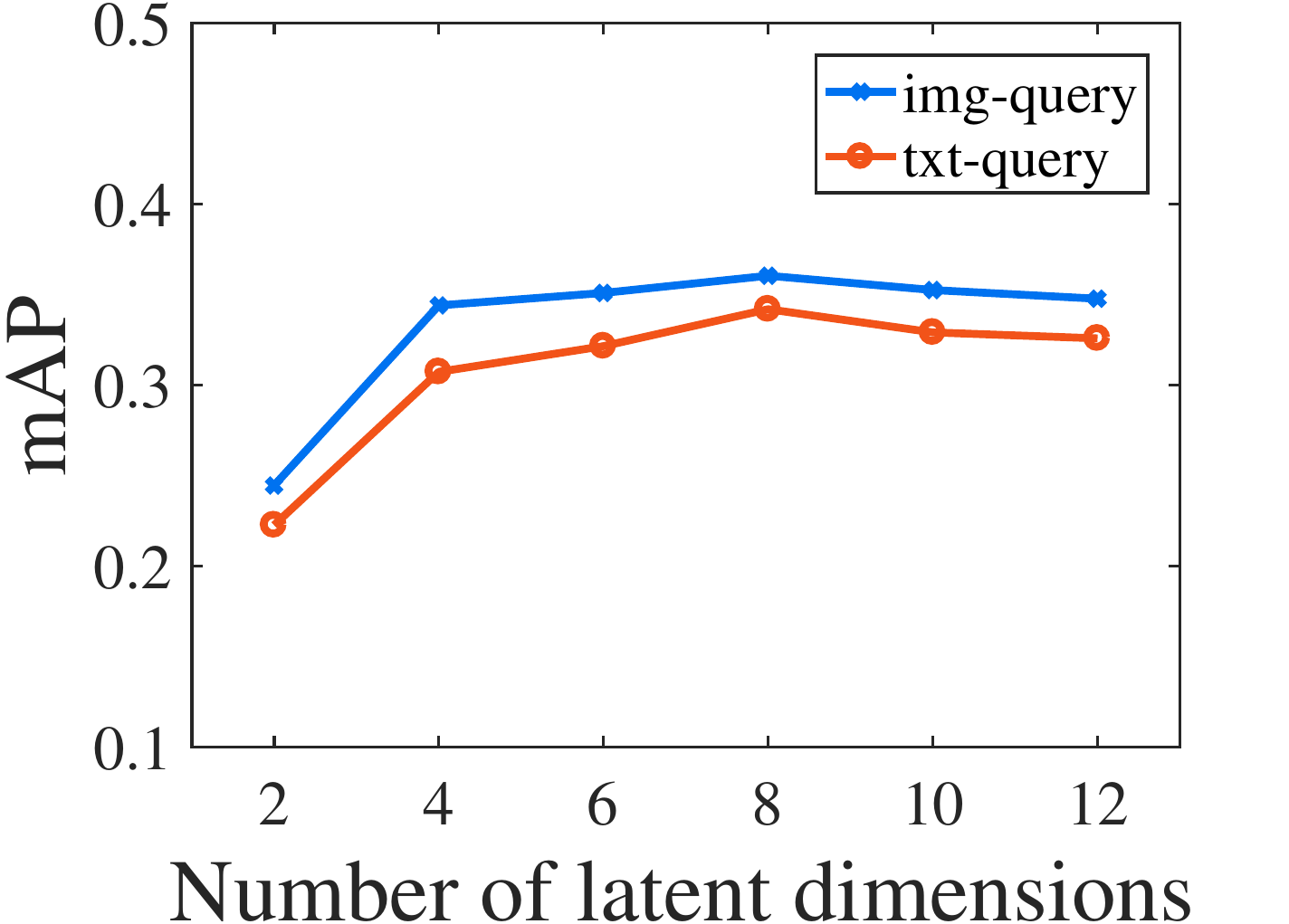}\label{fig:latenttvgraz}}
     \hspace{0.02\textwidth}
     \subfloat[MSCOCO]{\includegraphics[width=0.2\textwidth,height=2.3cm]{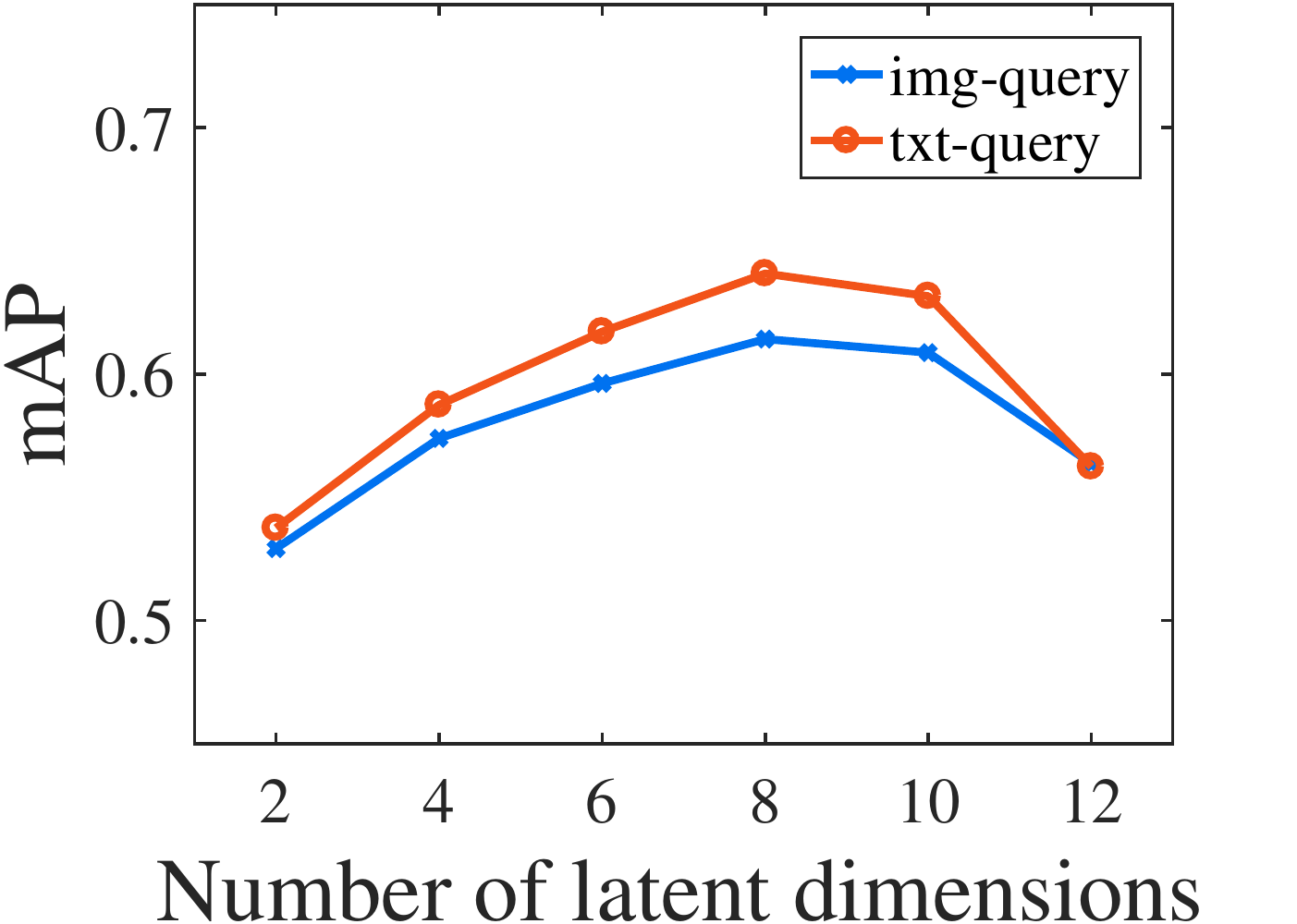}\label{fig:latentcoco}}
    \caption{ The mAP retrieval performance of hmGPLVM (tr) as a function of the dimensionality of the shared latent space. }\label{fig:latentdim}
\end{figure}

\subsubsection{Dimensionality of the latent space}\label{ssec:latentdim}
Taking hmGPLVM with the trace constraint as an example, we evaluate how the dimension of the latent manifold affects the performance for cross-modal correlation learning.
Fig.~\ref{fig:latentdim} shows that there is an optimal setting of the latent dimension on different datasets for cross-modal retrieval. For example, the performance of 8-dimensional latent space achieves the best on PASCAL dataset. Lower dimensional manifolds perform worse due to lack of flexibility in modeling the content divergence and correlation information between modalities. However, the performance may also decrease when the number of latent dimension is large, because additional correlation information may not be provided with extra dimensions. Similar phenomenon can also be observed for other extensions of harmonized GPLVM models.
Compared to existing subspace learning approaches~\cite{icml/WangALB15,iccv/RanjanRJ15} that usually fix the dimensionality of the common space to be 10 as reported in their papers, we obtain a lower dimensional embedding to summarize the high dimensional data, which also shows the remarkable representation learning ability of our non-linear non-parametric model. Thus we can improve the efficiency of our model with latent embeddings of lower dimensionality.
In our experiments, for all the proposed models, the dimensionality of the latent space is selected within $[7,10]$.

\begin{figure*}[!t]
\centering
     \subfloat[mGPLVM \emph{vs.} hmGPLVM]{\includegraphics[width=0.45\textwidth]{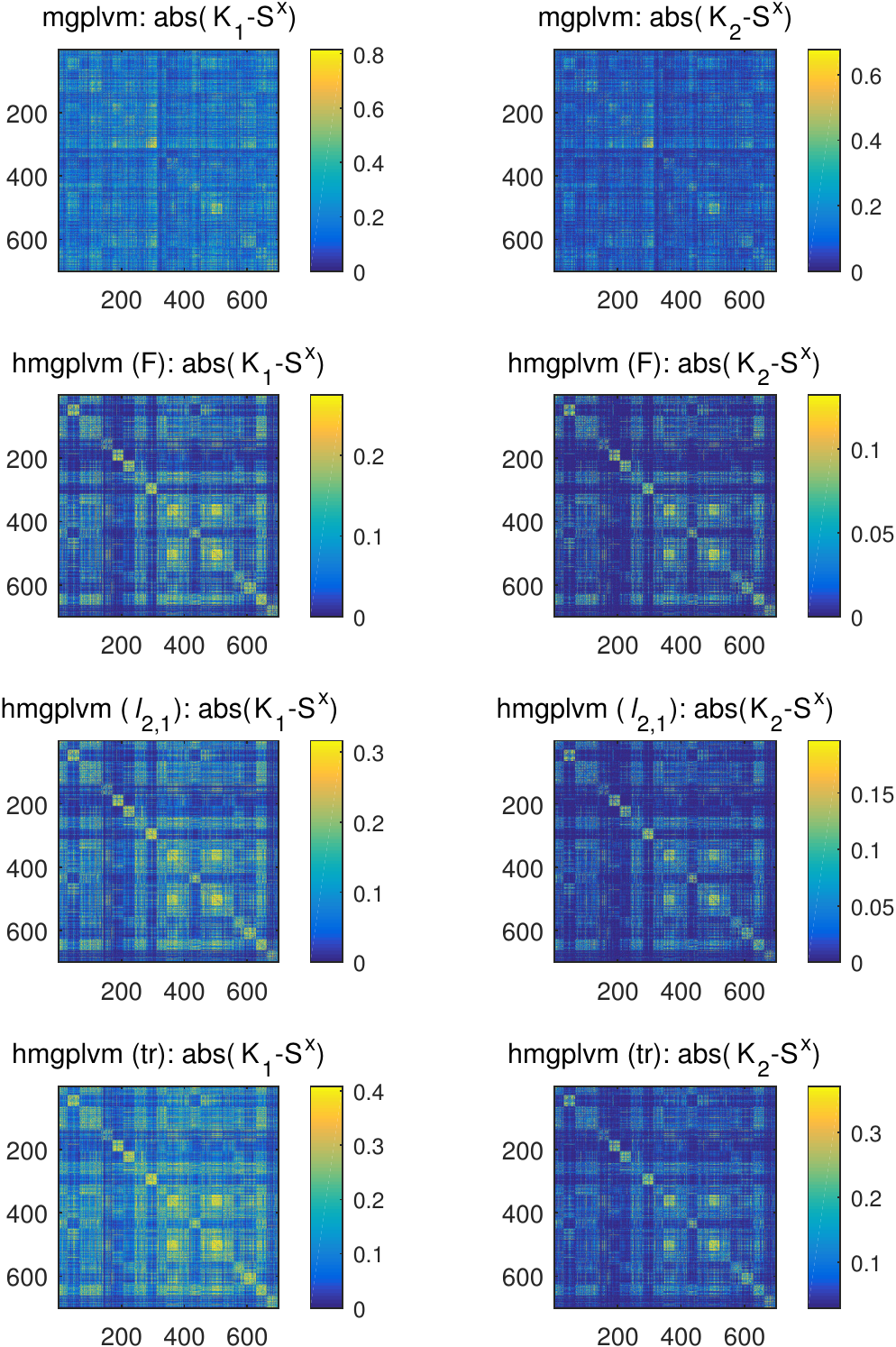}\label{fig:gpnorm}}
      \hfill
     \subfloat[m-SimGP \emph{vs.} hm-SimGP]{\includegraphics[width=0.45\textwidth]{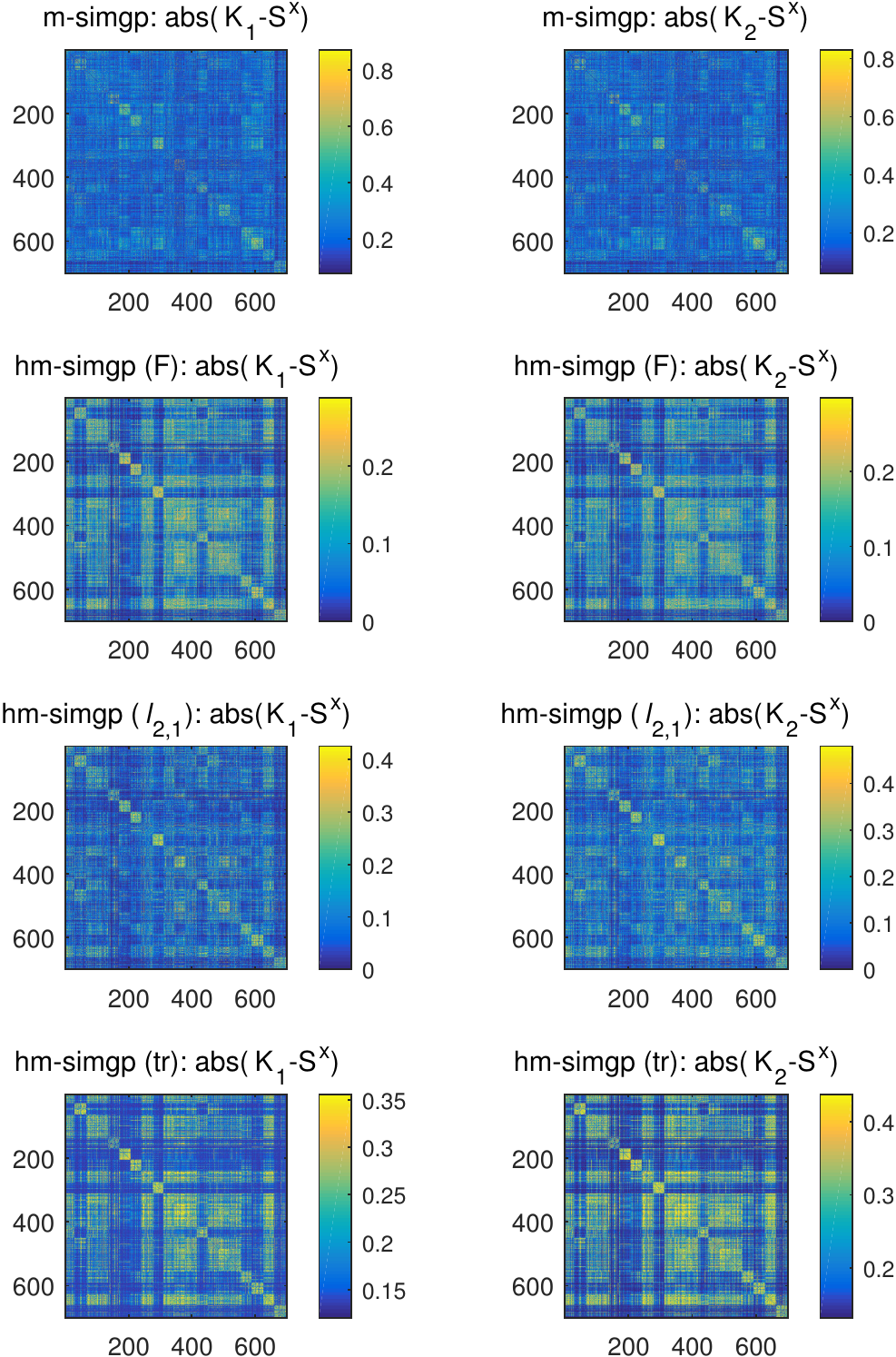}\label{fig:simgpnorm}}
    \caption{ Visualization of the absolute element-wise difference between modality-specific GP kernels and the similarity matrix in the latent space on PASCAL (Better viewed in color PDF). }\label{fig:norm}
\end{figure*}

\subsection{Analysis of the harmonization mechanism}
 \label{ssec:hmvisual}
In order to guarantee that the divergence between similarity structures in observed data space and the shared latent space to be small, we propose the harmonization constraints in Eq.~\eqref{eq:3-2} to preserve the structure consistency among GP kernels  ($K_1, K_2$) and the latent similarity ($S^x$).
Specifically, with the distance-based harmonization constraint, the minimization of the loss function forces them to be element-wise close to each other; with the ratio-based harmonization constraint, the minimization of the loss forces the ``distributions'' of different similarities to be close to each other. Thus, the structure consistency is achieved by a gradual ``resonance'' between the similarities during the model learning process. This experiment provides qualitative and quantitative comparisons between two sets of multimodal GP models on the PASCAL dataset.

Fig.~\ref{fig:norm} shows the absolute element-wise difference between modality-specific GP kernels and the similarity matrix in the latent space.
It is clear that the divergence between the two components of the proposed harmonized models is much smaller than those of the previous models. However, we also observe that the F-norm constraint makes the divergence even smaller than the $l_{2,1}$-norm constraint and the ratio-based trace constraint for both hmGPLVM and hm-SimGP. It indicates that the element-wise closeness between similarities is not the only favorable condition for the complex multimodal correlation learning problem.

\begin{table}[t]
\renewcommand\arraystretch{1.1} 
\caption{The Riemannian distance between modality-specific GP kernels and the similarity matrix of the latent representations on PASCAL.}
\label{tab:mmgpmflickr}
\centering
\begin{tabular}{l c c c }
\toprule
Method &$d\left( {{K_1},{S^x}} \right)$ & $d\left( {{K_2},{S^x}} \right)$ & Total \\

\midrule
mGPLVM & 66.9311 & 49.2587 & 116.1898\\
hmGPLVM (F) & 19.9302 & 14.3379 & 34.2681 \\
hmGPLVM ($l_{2,1}$) & 21.7559 & 15.2247 & 36.9806\\
hmGPLVM (tr) & 19.2857 & 19.4534 & 38.7391 \\

\midrule
m-SimGP & 66.6635 & 65.592 & 132.2555 \\
hm-SimGP (F & 19.4452 & 22.0663 & 41.5115 \\
 hm-SimGP ($l_{2,1}$) & 20.6055 & 20.9638 & 41.5693 \\
 hm-SimGP (tr) & 13.7584 & 17.2906 & 31.049 \\
\bottomrule
\end{tabular}
\end{table}

Table~\ref{tab:mmgpmflickr} provides quantitative results of the divergence between two similarity matrices measured by the Riemannian distance~\cite{forstner2003metric}, which is defined by using the sum of the squared logarithms of the eigenvalues, \emph{i.e.}, $d\left( {{K_c},{S^x}} \right) = \sqrt {\sum\limits_{i = 1}^N {{{\ln }^2}{\lambda _i}\left( {{K_c},{S^x}} \right)} }$. This distance metric can be used for evaluating positive definite covariance matrices.
We can see that for hmGPLVM the F-norm constraint achieves a smaller distance than the other two constraints, while for the similarity based model hm-SimGP the trace constraint achieves the smallest Riemannian distance between different covariance matrices. Still the quantitative results demonstrate that the distance between GP kernels and the latent similarity is reduced in our models with the harmonization mechanism.

\section{Conclusion}
\label{sec:conclusion}
We have presented a novel scheme for multimodal learning which enforces information sharing among different GPLVM components and across data modalities. We develop a harmonized learning method which is performed by imposing a joint prior on GP kernel parameters and the latent representations.
By incorporating the joint prior into a variety of multimodal GPLVMs, we propose several harmonized extensions for multimodal correlation learning, \emph{i.e.}, hmGPLVM, hm-SimGP and hm-RSimGP.
Compared to previous approaches, our method allows the intra-modal structure information to be transferred on the latent model parameter space. In return, the acquired latent representation is endowed with more ability to preserve the modality-specific topologies. Furthermore, the additional information transfer between kernel hyperparameters can help to avoid inappropriate solutions caused by noise and correlation sparsity in observed data, and thus can produce more semantically consistent latent representations.

Currently, the harmonization mechanism is built on GPLVM-based models. In future work, we intend to introduce it into other kernel-based approaches for multimodal learning to make one data modality to influence the training of another. Also we plan to investigate more complicated and flexible harmonization functions to adapt to the complex content of multimodal data. Further, we will apply the proposed method to more complicated multimodal correlation learning tasks, \emph{e.g.}, image-text matching~\cite{iccv/MaLSL15}.
We will study an extension of our method to be able to deal with missing data, since real-world multimodal data usually are unpaired or incomplete. 



\ifCLASSOPTIONcompsoc
  \section*{Acknowledgments}
\else
  \section*{Acknowledgment}
\fi

The authors would like to thank the associate editor and the reviewers for their time and effort provided to review the manuscript. This work was supported in part by National Natural Science Foundation of China: 61672497, 61620106009, U1636214 and 61836002, in part by National Basic Research Program of China (973 Program): 2015CB351802, in part by Key Research Program of Frontier Sciences of CAS: QYZDJ-SSW-SYS013, and in part by Project funded by China Postdoctoral Science Foundation: 119103S291. Shuhui Wang is the corresponding author.

\ifCLASSOPTIONcaptionsoff
  \newpage
\fi

\bibliographystyle{IEEEtran}
\bibliography{IEEEabrv,bib}

\begin{IEEEbiography}[{\includegraphics[width=1in,height=1.25in,clip,keepaspectratio]{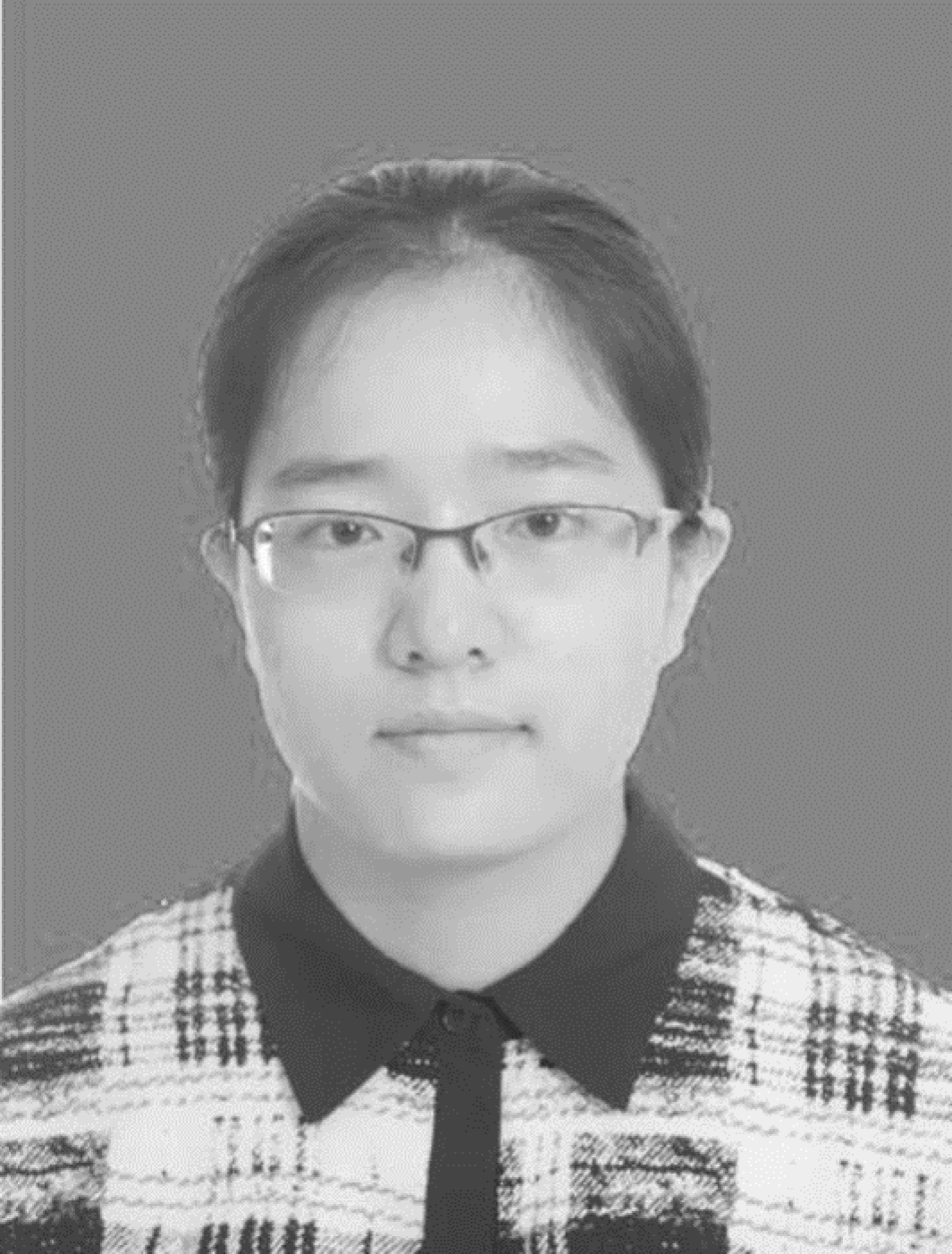}}]{Guoli Song} received the Ph.D. degree in computer engineering from University of Chinese Academy of Sciences in 2018, and received the B.S. degree in mathematics and applied mathematics and the M.S. degree in operational research and cybernetics from Zhengzhou University, in 2009 and 2012, respectively. She is currently a postdoc in the School of Computer Science and Technology, University of Chinese Academy of Sciences. Her  research interests include cross-media content analysis, computer vision, and machine learning.
\end{IEEEbiography}

\begin{IEEEbiography}[{\includegraphics[width=1in,height=1.25in,clip,keepaspectratio]{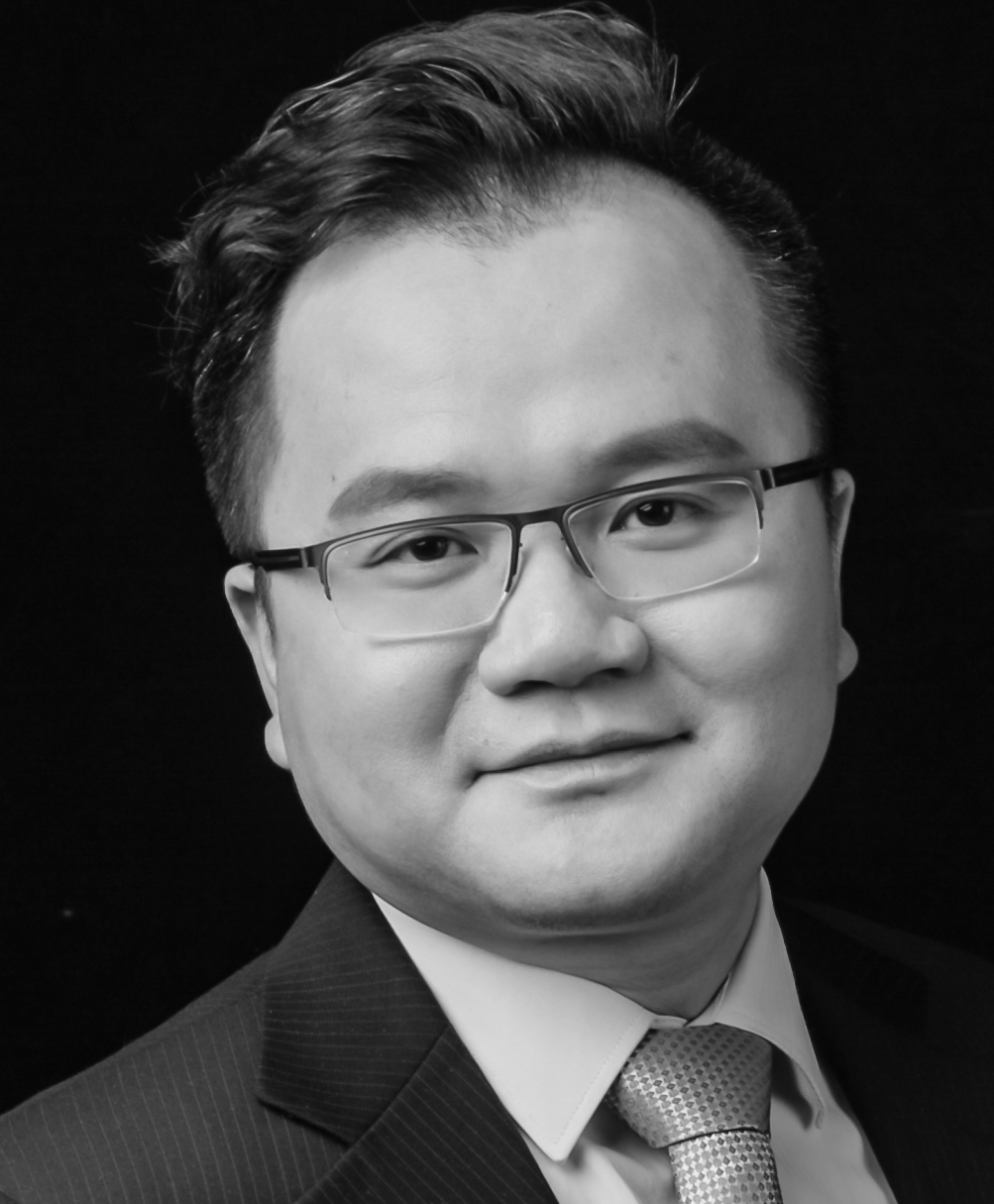}}]{Shuhui Wang} received the B.S. degree in electronics
engineering from Tsinghua University, Beijing, China, in 2006, and the Ph.D. degree from the Institute of Computing Technology, Chinese Academy
of Sciences, Beijing, China, in 2012. He is currently an Associate Professor with the Institute of Computing Technology, Chinese Academy of Sciences.
He is also with the Key Laboratory of Intelligent Information Processing, Chinese Academy of Sciences. His research interests include semantic
image analysis, image and video retrieval and large-scale Web multimedia data mining.
\end{IEEEbiography}

\begin{IEEEbiography}[{\includegraphics[width=1in,height=1.25in,clip,keepaspectratio]{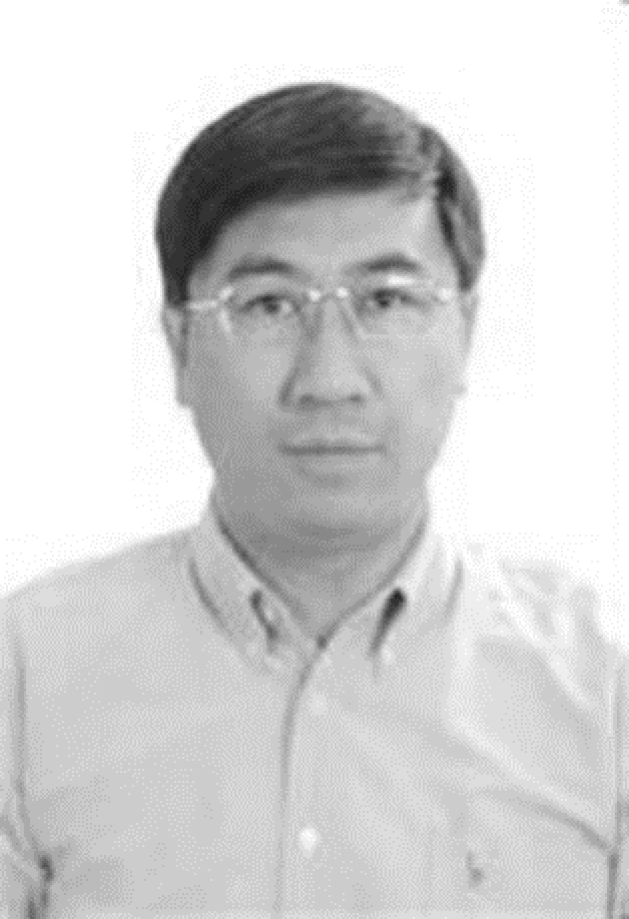}}]{Qingming Huang} received the B.S. degree
in computer science and Ph.D. degree in computer engineering from the Harbin Institute of Technology, Harbin, China, in 1988 and 1994,
respectively.
He is currently a Distinguished Professor with the School of Computer Science and Technology, University of Chinese Academy of Sciences. He has authored over 300 academic papers in international journals, such as IEEE Transactions on Pattern Analysis and Machine Intelligence, IEEE Transactions on Image Processing, IEEE Transactions on Multimedia, IEEE Transactions on Circuits and Systems for Video Technology, and top level international conferences, including the ACM Multimedia,
ICCV, CVPR, ECCV, VLDB, and IJCAI. He is the Associate Editor of IEEE Transactions on Circuits and Systems for Video Technology and the Associate Editor of Acta Automatica Sinica.
His research interests include multimedia computing, image/video processing, pattern recognition, and computer vision.
\end{IEEEbiography}

\begin{IEEEbiography}[{\includegraphics[width=1in,height=1.25in,clip,keepaspectratio]{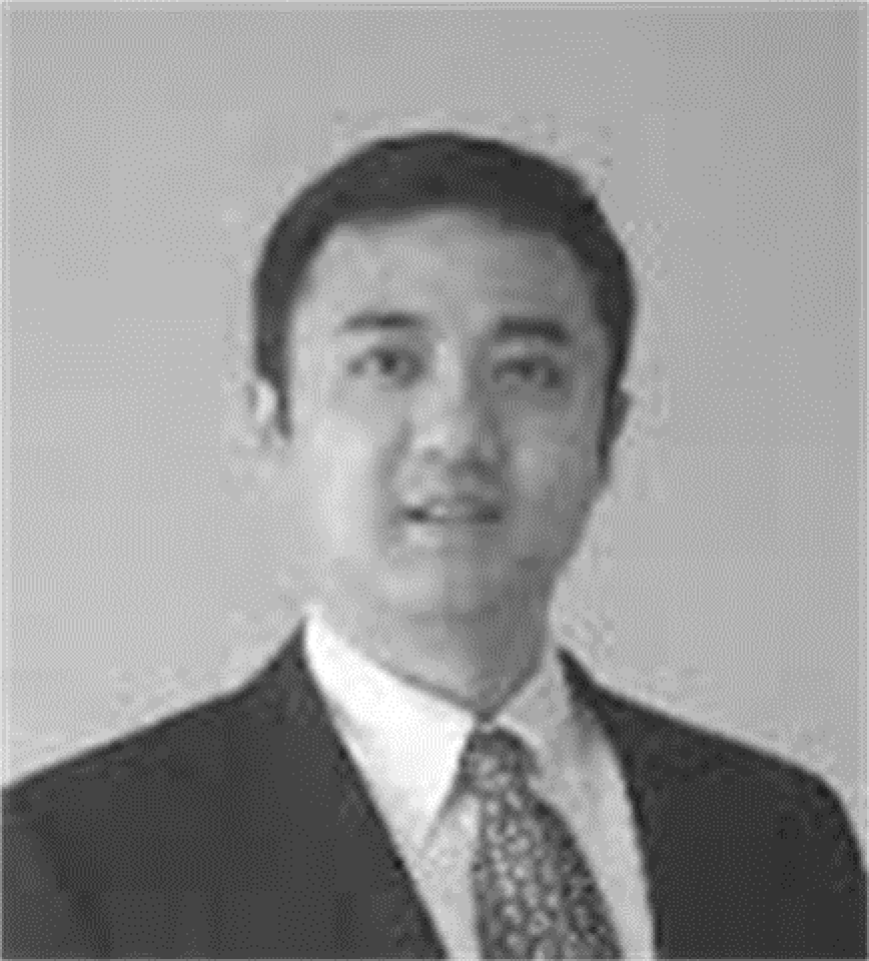}}]{Qi Tian} is currently a Full Professor in the Department of Computer Science, the University of Texas at San Antonio (UTSA).
He received his Ph.D. in ECE from University of Illinois at Urbana-Champaign (UIUC) in 2002 and received his B.E. in Electronic Engineering from Tsinghua University in 1992 and M.S. in ECE from Drexel University in 1996, respectively. His research interests include multimedia information retrieval, computer vision, pattern recognition and bioinformatics and published over 400 refereed journal and conference papers.
He is the associate editor of IEEE Transactions on Multimedia, IEEE Transactions on Circuits and Systems for Video Technology, ACM Transactions on Multimedia Computing, Communications, and Applications, Multimedia System Journal, and in the Editorial Board of Journal of Multimedia and Journal of Machine Vision and Applications.  Dr. Tian is the Guest Editor of IEEE Transactions on Multimedia, Journal of Computer Vision and Image Understanding, etc. Dr. Tian is a Fellow of IEEE.
\end{IEEEbiography}

\vfill


\end{document}